\def\year{2021}\relax
\documentclass[letterpaper]{article} 
\usepackage{aaai21} 
\usepackage{times}  
\usepackage{helvet} 
\usepackage{courier}  
\usepackage[hyphens]{url}  
\usepackage{graphicx} 
\usepackage{natbib}  
\usepackage{caption} 
\usepackage{tikz}
\usepackage{preamble}
\usepackage{caption}
\usepackage{wrapfig}
\usepackage[switch]{lineno}

\usepackage[acronym]{glossaries}

\newacronym[longplural={Gaussian Processes}]{gp}{GP}{Gaussian Process}
\newacronym[longplural={Deep Gaussian processes}]{dgp}{DGP}{Deep Gaussian Process}
\newacronym[longplural={Deep Recursive Residual Gaussian Processes}]{drgp}{DR2GP}{Deep Recursive Residual Gaussian Process}

\newcommand{\paragraf}[1]{\noindent\textbf{#1}\hspace{4pt}}
\newcommand{\dgp}{\acrshort{dgp}}
\newcommand{\dgps}{\acrshortpl{dgp}}

\title{Characterizing Deep Gaussian Processes via Nonlinear Recurrence Systems}
\author{
Anh Tong\textsuperscript{\rm 1},
Jaesik Choi\textsuperscript{\rm 2, 3}\\
}
\affiliations{
    \textsuperscript{\rm 1} Ulsan National Institute of Science and Technology\\
    \textsuperscript{\rm 2} Korea Advanced Institute of Science and Technology \\
    \textsuperscript{\rm 3} INEEJI\\
    anhth@unist.ac.kr, jaesik.choi@kaist.ac.kr
}

\begin{document}
\maketitle

\begin{abstract}
  Recent advances in Deep Gaussian Processes (DGPs) show the potential to have more expressive representation than that of traditional Gaussian Processes (GPs). However, there exists a pathology of deep Gaussian processes that their learning capacities reduce significantly when the number of layers increases. In this paper, we present a new analysis in DGPs by studying its corresponding nonlinear dynamic systems to explain the issue. Existing work reports the pathology for the squared exponential kernel function. We extend our investigation to four types of common stationary kernel functions. The recurrence relations between layers are analytically derived, providing a tighter bound and the rate of convergence of the dynamic systems. We demonstrate our finding with a number of experimental results.
\end{abstract}

\section{Introduction}
\acrfull{dgp}~\cite{deep_gp_2013} is a new promising class of models which are constructed by a hierarchical composition of Gaussian processes. The strength of this model lies in its capacity to have richer representation power from the hierarchical construction and its robustness to overfitting from the probabilistic modeling. Therefore, there have been extensive studies~\cite{nested_deep_gp, auto_encode_deep_gp, expect_propagation_deep_gp, rff_deep_gp, doubly_deep_gp, hamiltonian_deep_gp, deep_gp_importance_weight, dgp1, dgp2} contributing to this research area.

There exists a pathology, stating that the increase in the number of layers degrades the learning power of~\acrshort{dgp}~\cite{pathology_deep_gp}. That is, the functions produced by~\dgp{} priors become flat and cannot fit data. It is important to develop theoretical understanding of this behavior, and therefore to have proper tactics in designing model architectures and parameter regularization to prevent the issue.
Existing work~\cite{pathology_deep_gp} investigates the Jacobian matrix of a given model which can be analytically interpreted as the product of those in each layer. Based on the connection between the manifold of a function and the spectrum of its Jacobian, the authors show the degree of freedom is reduced significantly at deep layers. Another work~\cite{how_deep} studies the ergodicity of the Markov chain to explain the pathology.

To explain such phenomena, we study a quantity which measures the distance of any two layer outputs. We present a new approach that makes use of the statistical properties of the quantity passing from one layer to another layer. Therefore, our approach accurately captures the relations of the distance quantity between layers. By considering kernel hyperparameters, our method recursively computes the relations of two consecutive layers. Interestingly, the recurrence relations provide a tighter bound than that of~\cite{how_deep} and reveal the rate of convergence to fixed points. Under this unified approach, we further extend our analysis to five popular kernels which are not analyzed yet before. For example, the spectral mixture kernels do not suffer the pathology. We further provide a case study in~\dgp{}, showing the connection between our recurrence relations and learning~\dgps{}.

Our contributions in this paper are: (1) we provide a new perspective of the pathology in~\dgp{} under the lens of chaos theory; (2) we show that the recurrence relation between layers gives us the rate of convergence to a fixed point; (3) we give a unified approach to form the recurrence relation for several kernel functions including the squared exponential kernel function, the cosine kernel function, the periodic kernel function, the rational quadratic kernel function and the spectral mixture kernel; (4) we justify our findings with numerical experiments. We use the recurrence relations in debugging~\dgps{} and explore a new regularization on kernel hyperparameters to learn zero-mean~\dgps{}.
\begin{figure*}
    \centering
    \includegraphics[width=0.85\textwidth]{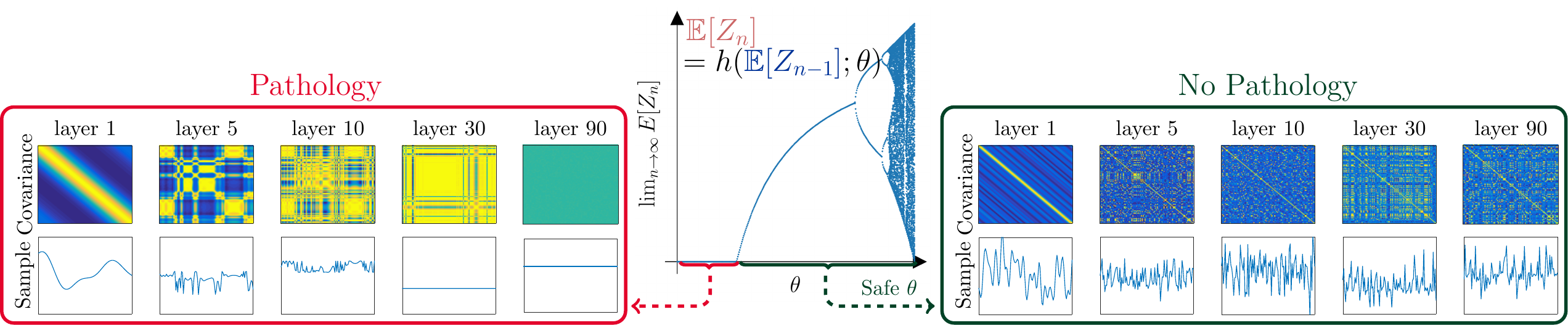}
    \caption{\small Studying the squared distance, $Z_n$, between outputs of two consecutive layers. The asymptotic property (middle plot) of the recurrence relation of this quantity between two consecutive layers decides the existence of pathology for a very deep model. Here, $\theta$ indicates kernel hyperparameters. The middle plot is the bifurcation plot providing the state of DGP at very deep layer. The pathology is identified by the zero-value region where $\expect[Z_n] \to 0$. Note that this bifurcation plot is for illustration purpose only.}
    \label{fig:main_figure_2}
\end{figure*}

\section{Background}
\textbf{Notation\hspace{0.2cm}} Throughout this paper, we use the boldface as vector or vector-value function. The superscript i.e. $f^{(d)}(\vx)$ is the $d$-th dimension of vector-valued function $\vf(\vx)$. 

\subsection{Deep Gaussian Processes}
We study~\dgps{} in composition formulation where~\acrshort{gp} layers are stacked hierarchically. An $N$-layer~\dgp{} is defined as
\begin{equation*}
    \vf_N \circ \vf_{N-1} \circ \dots \circ \vf_1(\vx), 
\end{equation*}
where, at layer $n$, for dimension $d$, $f^{(d)}_n| \vf_{n-1} \sim \mathcal{GP}(0, k_n(\cdot, \cdot))$ independently. Note that the GP priors have the mean functions set to zero. The nonzero-mean case is discussed later (Section~\ref{sec:nonpathological}). We shorthand $\vf_n \circ \vf_{n-1} \circ \dots \circ \vf_1(\vx)$ as $\vf_n(\vx)$ and write $k_n(\vf_{n-1}(\vx), \vf_{n-1}(\vx'))$ as $k_n(\vx, \vx')$. Let $m$ be the number of output of $\vf_n$. All layers have the same hyperparameters.

\begin{thm}[\cite{how_deep}]
\label{thm:how_deep}
	Assume that $k(\vx,\vx')$ is given by the squared exponential kernel function with variance $\sigma^2$ and lengthscale $\ell^2$ and that the input $\vx$ is bounded. Then if $\sigma^2< \ell^2/m$,
	$$\mathbb{P}(\normx{\vf_n(\vx) - \vf_n(\vx')}{2} \xrightarrow[n \to \infty]{} 0 \quad \text{  for all } \vx, \vx' \in \mathcal{D}) = 1$$
	where $\mathbb{P}$ denotes the law of process $\{f_n\}$.
\end{thm}

This theorem tells us the criterion that the event of vanishing in output magnitude happens infinitely often with probability $1$. 


\subsection{Analyzing dynamic systems with chaos theory}
\begin{figure}
    \centering
        {\includegraphics[width=0.22\textwidth]{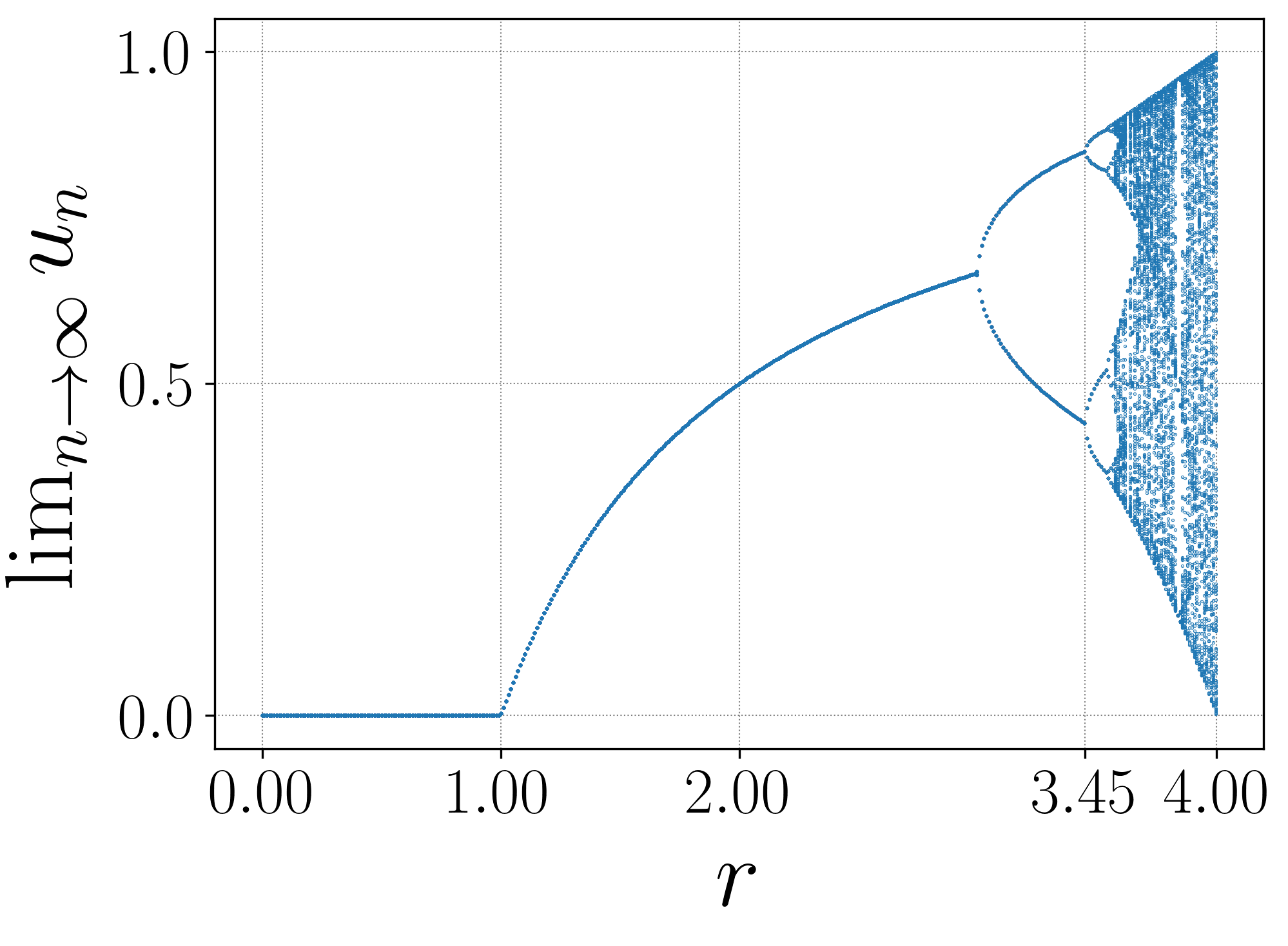}}
        \caption{Bifurcation plot of the logistic function $u_n = ru_{n-1}(1-u_{n-1})$.}
        \label{fig:chaotic}
\end{figure}
Recurrence maps representing dynamic transitions between~\dgp{} layers are nonlinear. Studying the dynamic states and convergence properties for nonlinear recurrences is not as well-established as those of linear recurrences. As an example, given a simple nonlinear model like the logistic map: $u_n = ru_{n-1}(1 - u_{n-1})$, its dynamic behaviors can be complicated~\cite{nature_dynamic}.

Recurrent plots or {bifurcation} plots have been used to analyze the behavior of chaotic systems. The plots are produced by simulating and recording the dynamic states up to very large time points. This tool allows us to monitor the qualitative changes in a system, illustrating fixed points asymptotically, or possible visited values. Other techniques, e.g. transient chaos~\cite{transient_chaos_dnn}, recurrence relations~\cite{deep_information_propagation} have been used to study deep neural networks.

We take the logistic map as an example to understand a recurrence relation. Figure~\ref{fig:chaotic} is the bifurcation plot of the logistic map. This logistic map is used to describe the characteristics of a system which models a population function. We can see that the plot reveals the state of the system, showing whether the population becomes extinct ($0<r<1$), stable ($1<r<3$), or fluctuating ($r>3.4$) by seeing the parameter $r$.


\section{Moment-generating function of distance quantity}

\begin{figure}
        \centering
        \includegraphics[width=0.43\textwidth]{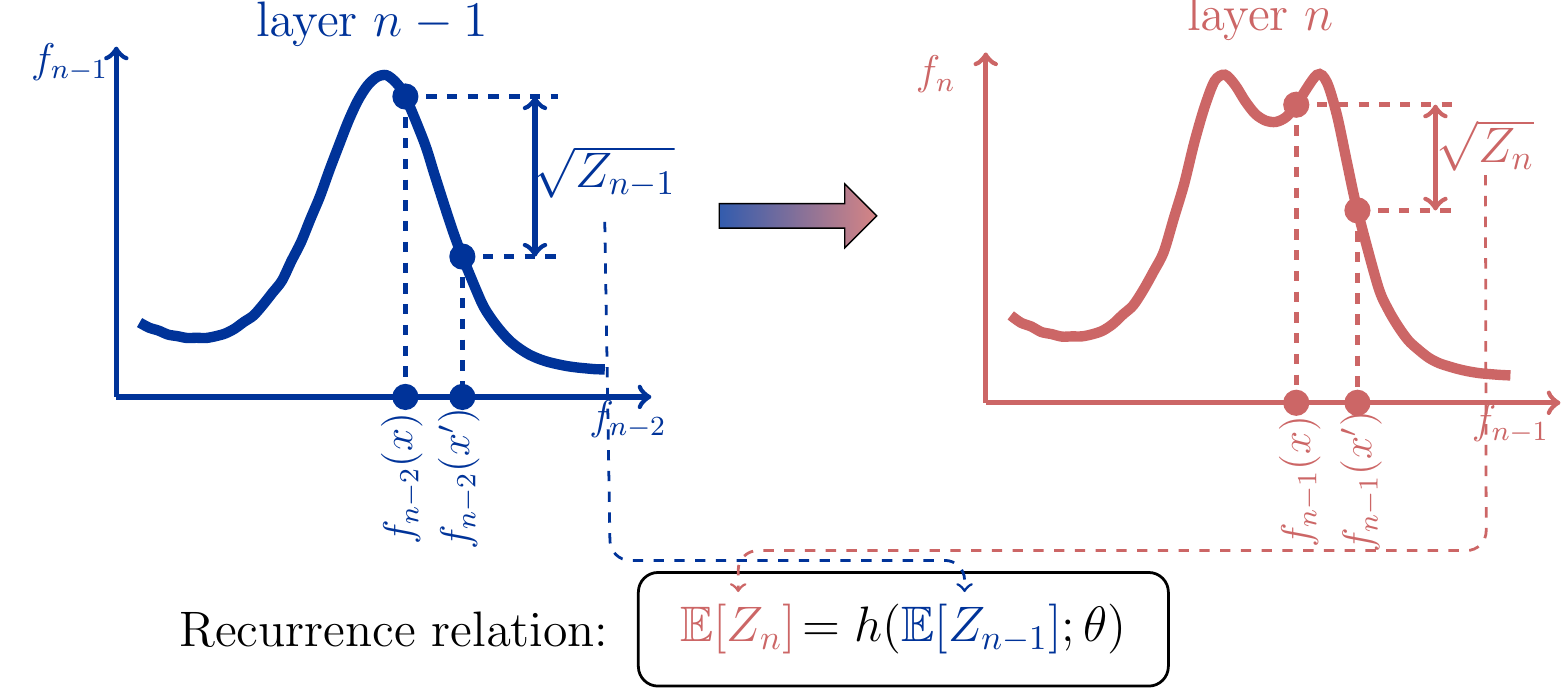}
        \caption{Finding the recurrence relation of the quantity $\expect[(f_n(x)-f_n(x'))^2]$ between two consecutive layers.}
        \label{fig:main_figure_1}
\end{figure}
Throughout this paper, we are interested in quantifying the expectation of the squared Euclidean distance between any two outputs of a layer and thereby study the dynamics of this quantity from a layer to the next layer. Figure~\ref{fig:main_figure_2} shows that we can make use of the found recurrence relations to study the pathology of~\dgps{}.

For any input pair $\vx$ and $\vx'$, we define such quantity at layer $n$ as $Z_n = \normx{\vf_n(\vx) - \vf_n(\vx')}{2}^2 = 
\sum_{d=1}^m \left(f^{(d)}_n(\vx) - f^{(d)}_n(\vx')\right)^2.$
When the previous layer $\vf_{n-1}$ is given, the difference between any $f^{(d)}_n(\vx)$ and $f^{(d)}_n(\vx')$ is Gaussian,
\begin{equation*}
    \left(f^{(d)}_n(\vx) - f_n^{(d)}(\vx')\right) | \vf_{n-1} \sim \Normal(0, s_n).
\end{equation*}
Here $s_n = k_n(\vx, \vx) + k_n(\vx', \vx') - 2k_n(\vx, \vx')$ which is obtained from subtracting two dependent Gaussians. We can normalize the difference between $f^{(d)}_n(\vx)$ and $f^{(d)}_n(\vx')$ by a factor $\sqrt{s_n}$ to obtain  the form of standard normal distribution as
\begin{equation*}
    \frac{(f^{(d)}_n(\vx) - f^{(d)}_n(\vx'))}{\sqrt{s_n}} | \vf_{n-1} \sim \Normal(0, 1).
\end{equation*}
Since all dimensions $d$ in a layer are independent, we can say that $\frac{Z_n}{s_n}|\vf_{n-1} \sim \chi^2_m,$ is distributed according to the Chi-squared distribution with $m$ degrees of freedom. 

One useful property of the Chi-squared distribution is that the moment-generating function of $\frac{Z_n}{s_n}|\vf_{n-1}$ can be written in an analytical form, with $t \leq 1/2$, 
\begin{equation}
    M_{\frac{Z_n}{s_n}|\vf_{n-1}} (t) = \expect\left[\exp\left(t\frac{Z_n}{s_n}\right)| \vf_{n-1}\right] = (1 -  2t)^{-m/2}. 
    \label{eq:moment_generating_chi_squared}
\end{equation}

We shall see that the expectation of the distance quantity $Z_n$ is computed via a kernel function which, in most cases, involves exponentiations. Given that the input of this kernel is governed by a distribution, i.e., $\chi^2$, the moment-generating function becomes convenient to obtain our desired expectations.

Figure~\ref{fig:main_figure_1} depicts our approach to extract a function $h(\cdot)$ which models the recurrence relation between $\expect[Z_n]$ and $\expect[Z_{n-1}]$. This is also the main theme of this paper. 
\section{Finding recurrence relations}
This section presents the formalization of the recurrence relation of $\expect[Z_n]$ for each kernel function. We start off with the squared exponential kernel function.
\subsection{Squared exponential kernel function}
The squared exponential kernel (\kSE) is defined in the form of
\begin{equation}
    \kSE(\vx,\vx') = \sigma^2 \exp\left(-{\normx{\vx - \vx'}{}^2}/{2\ell^2}\right). 
\end{equation}

\begin{thm}[{\dgp} with $\kSE$]
\label{thm:dgp_se}
Given a triplet $(m, \sigma^2, \ell^2),$ $m \ge 1$ such that the following sequence converges to $0$:
\begin{equation}
    u_n = 2m\sigma^2\left(1-(1 + {u_{n-1}}/{m\ell^2})^{-m/2}\right),
    \label{eq:recur_se}
\end{equation}
Then, 
$\mathbb{P}(\normx{\vf_n(\vx) - \vf_n(\vx')}{2} \xrightarrow[n \to \infty]{} 0 \text{ for all } \vx, \vx' \in \mathcal{D}) = 1.$
\end{thm}{}
\begin{proof}[Proof] 
Note that we do not directly have access to $\expect[Z_n]$ but $\expect[Z_n|\vf_{n-1}]$ because of the Markov structure of the~{\dgp} construction. Getting $\expect[Z_n]$ is done via $\expect[Z_n|\vf_{n-1}]$ where we use the law of total expectation $\expect[Z_n] = \expect_{\vf_{n-1}}[\expect[Z_n|\vf_{n-1}]]$.

Now, we study the term $\expect[Z_n|\vf_{n-1}]$:
\begin{equation}
    \begin{aligned}
    \expect[Z_n|\vf_{n-1}] = & \expect[\sum_{d=1}^m(f_n^{(d)}(x) - f_n^{(d)}(x'))^2|\vf_{n-1}] \\
	= & 2m\sigma^2 - 2m k_n(\vx,\vx'). 
    \end{aligned}
    \label{eq:key_equation}
\end{equation}
The second equality is followed by $\expect[(f^{(d)}_n(\vx))^2]={\expect[(f_n^{(d)}(\vx'))^2]}=\sigma^2$ and $\expect[f_n^{(d)}(\vx)f_n^{(d)}(\vx')] = k_n(\vx,\vx')$. Recall that we write $k_n(\vx,\vx) = k_n(\vf_{n-1}(\vx), \vf_{n-1}(\vx'))$. 
By the definition of $\kSE$ kernel, we have
\begin{align*}
    \expect[Z_n|\vf_{n-1}] &= 2m\sigma^2\left(1 - \exp\left(-\frac{Z_{n-1}}{2\ell^2}\right)\right).
\end{align*}
Applying the law of total expectation, we have
\begin{align*}
    \expect[Z_n] = 2m\sigma^2\left(1 - \expect\left[\exp\left( -\frac{Z_{n-1}}{2\ell^2}\right)\right]\right).
\end{align*}
Again, we can only compute $\expect[\exp(-\frac{Z_{n-1}}{2\ell^2})] = \expect_{\vf_{n-2}}[\expect[\exp(-\frac{Z_{n-1}}{2\ell^2})| \vf_{n-2}]$. 
The expectation will be computed by the formula of the moment-generating function with respect to $\frac{Z_{n-1}}{s_{n-1}}| \vf_{n-2}$ where $t = -\frac{s_{n-1}}{2\ell^2}$ in Equation~(\ref{eq:moment_generating_chi_squared}). Choosing this value also satisfies the condition $t\leq 1/2$. Now, we have
\begin{equation}
    \begin{aligned}
    \expect[Z_n] &= 2m\sigma^2\left(1 - \expect\left[\left(1 + {s_{n-1}}/{\ell^2}\right)^{-m/2}\right]\right) \\
    &\leq 2m\sigma^2\left(1 - \left(1 + {\expect\left[s_{n-1}\right]}/{\ell^2}\right)^{-m/2}\right).
    \end{aligned}
    \label{eq:some_eq}
\end{equation}
Here, Jensen's inequality is used as $(1 + x)^{-a}$ is convex for any $x >0$. 
By Equation~(\ref{eq:key_equation}), we have
\begin{equation*}  
    \frac{\expect[Z_{n-1}|\vf_{n-2}]}{m} =  2\sigma^2 - 2k_{n-1}(\vx, \vx') = s_{n-1}.
\end{equation*}
Replacing $s_{n-1}$ in Equation~(\ref{eq:some_eq}) and applying the law of total expectation for the case of $Z_{n-1}$, we obtain recurrence relation between layer $n-1$ and layer $n$ is
$$\expect[Z_n] \leq 2m\sigma^2\left(1 - \left(1 + {\expect[Z_{n-1}]}/{m\ell^2}\right)^{-m/2}\right).$$

Using the Markov inequality, for any $\epsilon$, we can bound
$    \mathbb{P}(Z_n \geq \epsilon) \leq \frac{\expect[Z_n]}{\epsilon^2}.
$

At this point, $u_n$ defined in Equation~(\ref{eq:recur_se}) is considered as the upper bound of $\expect[Z_n]$. We condition that $\{u_n\}$ converges to $0$, then $\{\expect[Z_n]\}$ converges to $0$ as well. By the first Borel-Cantelli lemma, we have
$    \mathbb{P}(\limsup_{n \to \infty}{Z_n} \geq \epsilon) = 0,$
which leads to the conclusion in the same manners as~\cite{how_deep}.
\end{proof}
\begin{figure}[t]
\centering
\begin{tikzpicture}
    \node[] at (0, 0) {
        \includegraphics[width=0.16\textwidth]{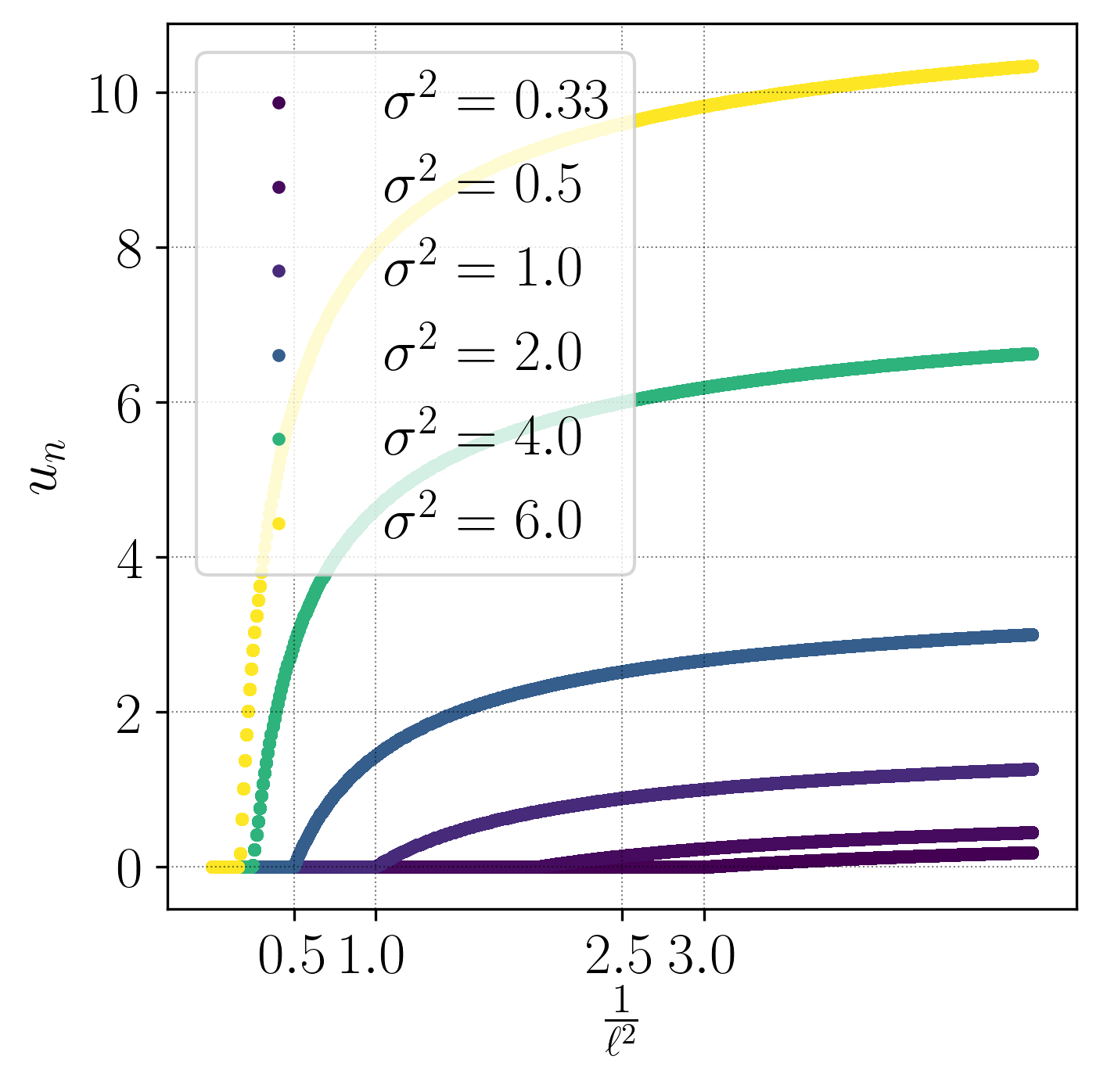}
    };
    \node[] at (3.,0) {
        \includegraphics[width=0.165\textwidth]{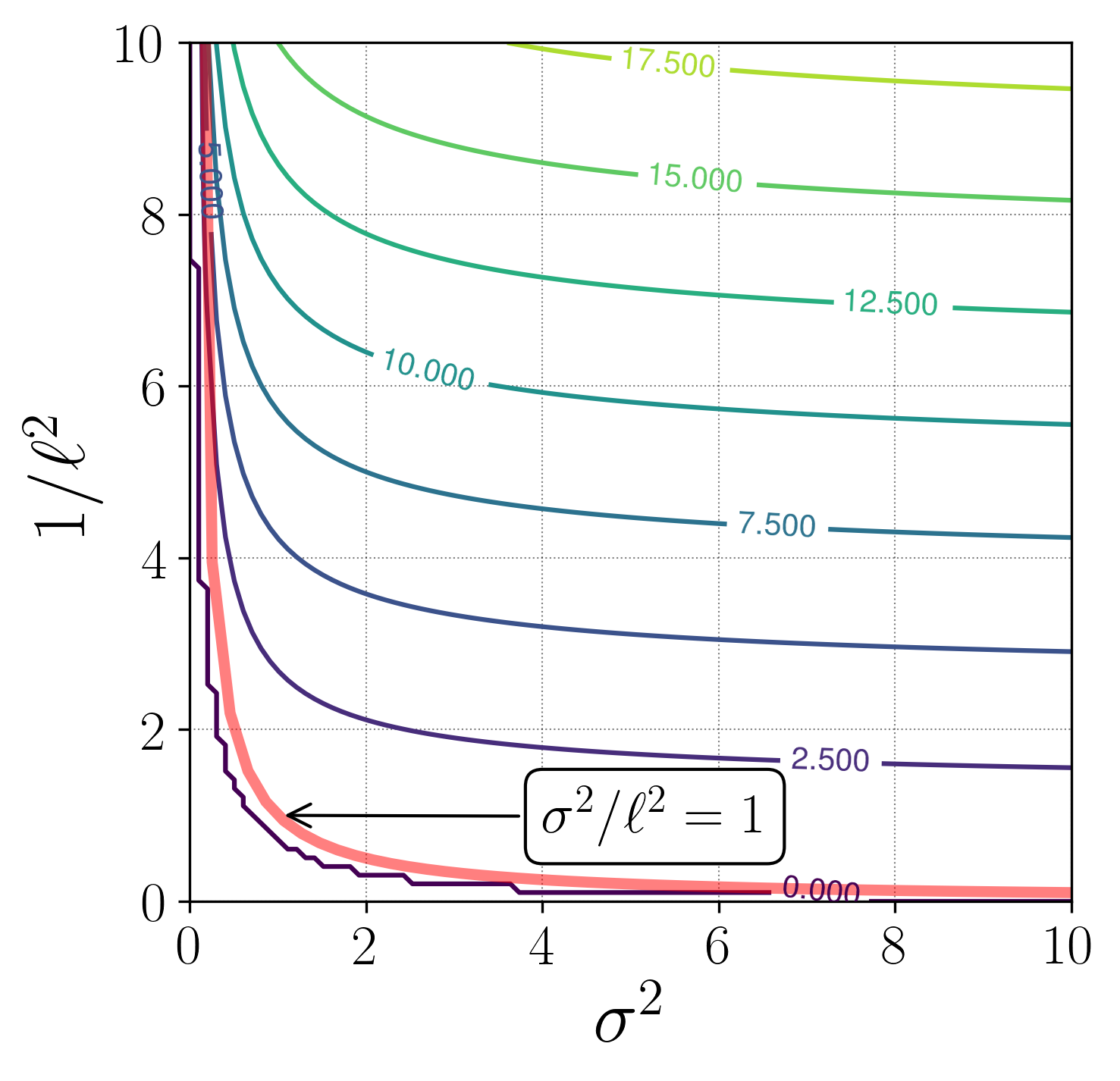}
    };
    
    \node[] at (6.4,0) {
        \includegraphics[width=0.22\textwidth]{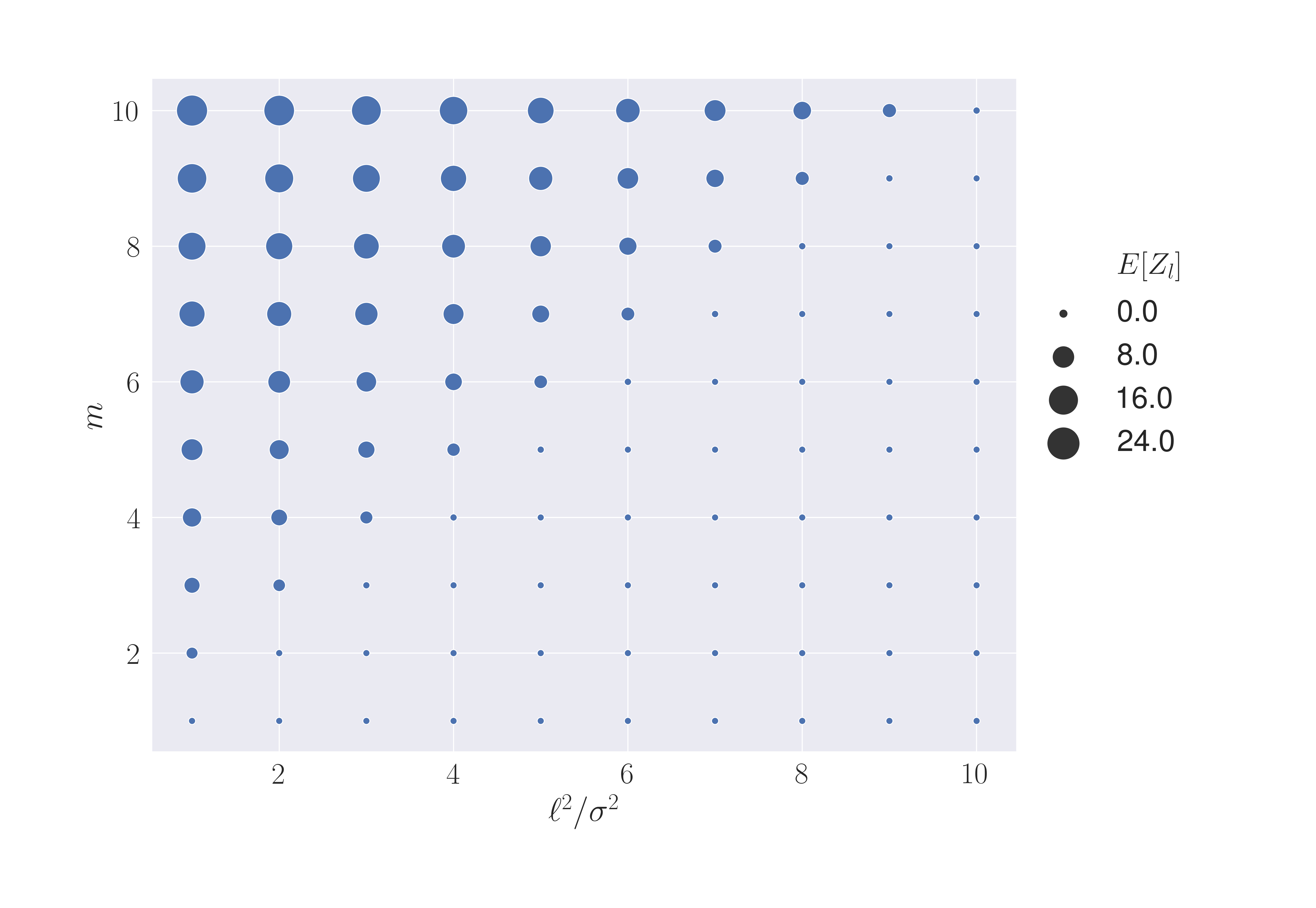}
    };
    
    \node[] at (0.2, 1.6) {\small (a)};
    \node[] at (3.2, 1.6) {\small (b)};
    \node[] at (6.6, 1.6) {\small (c)};
    
\end{tikzpicture}
    \caption{(a): Bifurcation plot of the recurrence relation of $\kSE$ kernel for $m=1$. (b): Contour plot of $u_n$ at layer $n=300$ and $m=1$. The misalignment between the \textcolor{red}{red line} ($\nicefrac{\sigma^2}{\ell^2} = 1$) and the zero-level contour is due to numerical errors. (c): Increase $m > \sigma^2 / \ell^2$ to avoid pathology.}
    \label{fig:se_bifurcation}
\end{figure}
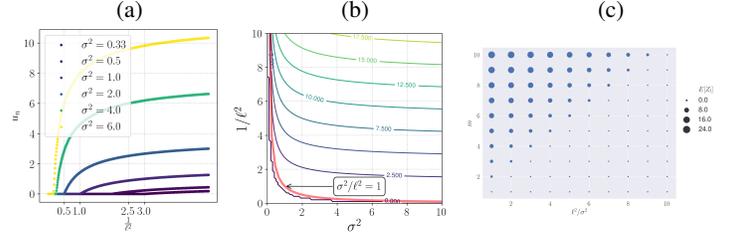

\paragraf{Analyzing the recurrence} 
Figure~\ref{fig:se_bifurcation}a illustrates the bifurcation plot of Equation~(\ref{eq:recur_se}) with $m=1$. The non-zero contour region in Figure~\ref{fig:se_bifurcation}b tells us that $\sigma^2/\ell^2$ should be smaller than $1$ to escape the pathology. When $m> 1$, Figure~\ref{fig:se_bifurcation}c shows that if $m> \sigma^2 /\ell^2$, $u_n$ does not approach to $0$, implying the condition to prevent the pathology. This result is consistent with Theorem~\ref{thm:how_deep} in~\cite{how_deep}.

\paragraf{Discussion} 
Note that the relation between $\expect[Z_n]$ and $\expect[Z_{n-1}]$ presents a tighter bound than existing work~\cite{how_deep}. If we construct the recurrence relation based on~\cite{how_deep}, $\expect[Z_n]$ is bounded by
\begin{equation}
    \expect[Z_n] \leq \frac{m\sigma^2}{\ell^2} \expect[Z_{n-1}].
    \label{eq:how_deep_recur}
\end{equation}
 One can show that $(1 + x)^a \geq 1 - ax, a < 0, x > 0$, implying
\begin{equation*}
    2m\sigma^2 (1 - (1 + \expect[Z_{n-1}]/(m\ell^2))^{-m/2}) \leq m\sigma^2\expect[Z_{n-1}]/\ell^2.
\end{equation*}
In fact, a numerical experiment shows that our bound of $\expect[Z_n]$ is found to be close to the true $\expect[Z_n]$ (Section~\ref{sec:correct_recur_relations}). That is, we can see the trajectory of $\expect[Z_n]$ for every layer of a given model of which the depth is not necessary to be infinitely many. 

One can reinterpret the recurrence relation for each dimension $d$ as
\begin{equation*}
    \textstyle {\expect[Z_n^{(d)}]}\leq 2\sigma^2\left(1 - \left(1 + {\expect[Z_{n-1}^{(d)}]}/{\ell^2}\right)^{-m/2}\right),
\end{equation*}
where ${\expect[Z^{(d)}_n]} {=} {\frac{\expect[Z_n]}{m}}$ with ${Z^{(d)}_n} {=} {\left(f^{(d)}_n(\vx) - f^{(d)}_n(\vx')\right)^2}$.


\paragraf{A guideline to obtain a recurrence relation} Given a specific kernel function, one may follow these steps to acquire the corresponding recurrence relation: (1) considering the form of kernel input where it may be distributed according to either the Chi-squared distribution or its variants (presented in the next sections); (2) checking whether there is a way to represent the kernel function under representations such that statistical properties of kernel inputs are known; (3) caring about the convexity of the function  after choosing a proper setting (as we bound the expectation with Jensen's inequality in the proof of Theorem~\ref{thm:dgp_se}).

\subsection{Cosine kernel function}
The cosine kernel (\kCos) function takes inputs as the distance between two points instead of the squared distance like in the case of $\kSE$ kernel. We will mainly work with $\sqrt{Z_n}$ in this subsection. The cosine kernel function $k(\vx,\vx') = \kCos(\vx,\vx')$ which is defined as
 \begin{equation*}
     \kCos(\vx,\vx') = \sigma^2 \cos\left({\pi \normx{\vx - \vx'}{2}}/{p}\right).
 \end{equation*}
Starting with Equation~(\ref{eq:key_equation}) and using the definition of $\kCos$ kernel, we have
\begin{align*}
\textstyle
     \expect [Z_n| \vf_{n-1} ] = & 2m\sigma^2 - 2m\sigma^2 \cos({\pi \sqrt{Z_{n-1}}}/{p})\\
     = & 2m\sigma^2 - m\sigma^2 \exp({i\pi \sqrt{Z_{n-1}}}/{p}) \\
     & \quad - m\sigma^2 \exp(-{i\pi \sqrt{Z_{n-1}}}/{p}).
 \end{align*}
 Here, Euler's formula is used to represent $\cos(\cdot)$ and $i$ is the imaginary unit ($i^2=-1$). To obtain $\expect[Z_n]$, we use the law of total expectation and compute the two following expectations: $\expect\left[\exp({i\pi\sqrt{Z_{n-1}}}/{p}) | \vf_{n-2}\right]$ and $\expect\left[\exp(-{i\pi\sqrt{Z_{n-1}}}/{p}) | \vf_{n-2}\right]$. From $\frac{Z_n}{s_{n}}|\vf_{n-1} \sim \chi^2_m$, we have
$ \sqrt{\frac{Z_n}{s_{n}}}|\vf_{n-1} \sim \chi_m, $
is distributed according to the Chi distribution. This observation follows the first step in the guideline. The characteristic function of the Chi distribution for random variable $\sqrt{\frac{Z_n}{s_n}}|\vf_{n-1}$ is
\begin{align*}
    &\varphi_{\sqrt{{Z_n}/{s_n}}|\vf_{n-1}}(t) =  \expect
    \left[\exp\left( it \sqrt{{Z_n}/{s_n}}\right)\right]\\
    & =   {}_1F_{1}(\frac{m}{2}, \frac{1}{2}, \frac{-t^2}{2}) + 
    it\sqrt{2}\frac{\Gamma((m+1)/2)}{\Gamma(m/2)} {}_1F_{1}(\frac{m+1}{2}, \frac{3}{2}, \frac{-t^2}{2}). 
\end{align*}
where ${}_1F_{1}(a,b,z)$ is Kummer's confluent hypergeometric function (see Definition in Appendix~\ref{appendix:hypergeometric}). This is considered as the second step in the guideline. Back to our process of finding the recurrence function, we consider the case $\sqrt{\frac{Z_{n-1}}{s_{n-1}}}|\vf_{n-2} \sim \chi_m$. By choosing $t = \pm \frac{\pi\sqrt{s_{n-1}}}{p}$ for its characteristic function, we can obtain
 \begin{equation*}
     \expect[Z_n] = 2m\sigma^2 \left(1 - {}_1F_{1}(\frac{m}{2}, \frac{1}{2}, -\frac{\pi^2 }{2p^2} \expect[Z_{n-1}|\vf_{n-2}] )\right). 
 \end{equation*}
 This is because the imaginary parts of $\varphi(t= \frac{\pi\sqrt{s_{n-1}}}{p})$ and $\varphi(-\frac{\pi\sqrt{s_{n-1}}}{p})$ are canceled out. 
 
 As the third step in the guideline, we perform a sanity check about the convexity of ${}_1F_{1}$. Only  with $m=1$, ${}_1F_{1}(\frac{1}{2}, \frac{1}{2}, \frac{-t^2}{2} ) = \exp(-\frac{t^2}{2})$ is convex. Our result in this case is restricted to $m=1$.
 Now, we can state that the recurrence relation is
 \begin{equation}
     u_n = 2\sigma^2\left(1 - \exp(-{\pi^2 } u_{n-1}/{2p^2} )\right). 
 \end{equation}

\subsection{Spectral mixture kernel function}
In this paper, we consider the spectral mixture (SM) kernel~\cite{sm_kernel} in one-dimensional case with one mixture:
\begin{equation*}
    \kSM(r) = \exp(-2\pi^2 \sigma^2 r^2) \cos(2\pi \mu r), 
\end{equation*}
where $r = \normx{x - x'}{2}$, and $\sigma^2, \mu > 0$. We can rewrite this kernel function as
$\frac{1}{2}w^2\{\exp(-v^2(r + {i} u)^2) + \exp(-v^2(r - {i} u)^2)\}.$
Here we simplify the kernel by change in variables as $w^2= \exp(-\frac{\mu^2}{2\sigma^2})$, $v^2 = 2\pi^2\sigma^2$, and $u = \frac{\mu}{2\pi\sigma^2}$.\\
 With a similar approach, we compute the expectation of $\expect[\exp(-v^2(\sqrt{Z_{n-1}} \pm {i} u)^2)].$
We can identify that $\frac{(\sqrt{Z_{n-1}} \pm {i}u)^2}{s_{n-1}}| f_{n-2} \sim \chi_1'^2(\lambda)$ is distributed according to a \emph{non-central} Chi-squared distribution of which the moment-generating function is
\begin{equation*}
    M_{\chi_1'^2}(t; \lambda) = {(1 - 2t)^{-1/2}}{\exp(\lambda t / (1 - 2t))},
\end{equation*}
with the noncentrality parameter is $\lambda = -u^2/s_{n-1}$. By choosing an appropriate $t = -v^2 s_{n-1}$, we obtain the recurrence as
\begin{equation*}
    \expect[Z_n] \leq 2(1 - w^2M_{{\chi'}_1^2}(t =-{v^2 \expect[Z_{n-1}]}; \lambda = -u^2/\expect[Z_{n-1}])).
\end{equation*}
Note that the convexity requirement is satisfied. This recurrence relation of \textsc{SM} kernel has one additional exponent term when comparing to that of $\kSE$. We provide a precise formula and an extension to the high-dimensional case in Appendix~\ref{appendix:sm}. 
\subsection{Extension to non-pathological cases}
\label{sec:nonpathological}
We use our approach to analyze two cases including \emph{nonzero-mean}~\dgps{} and \emph{input-connected}~\dgps{} where there is no pathology occurring.\\
\paragraf{Nonzero-mean \dgps{}}
Let ${f^{(d)}_n(\mathbf{x})} {\sim} {\mathcal{GP}}{(\mu_n(\mathbf{x})},{ k_n(\mathbf{x},\mathbf{x}'))}$ with the mean function $\mu_n(\mathbf{x})$, the difference between two outputs, $(f^{(d)}_n(\mathbf{x}) - f^{(d)}_n(\mathbf{x}')) \sim \mathcal{N}(\nu_n, s_n)$ with ${\nu_n = \mu_n(\mathbf{x}) - \mu_n(\mathbf{x}')}$. This leads to $\frac{Z_n}{s_n}|\boldsymbol{f}_n \sim \chi'^2_m$, the {\textit{non-central} Chi-squared} distribution with the non-central parameter ${\lambda = m\nu_n^2}$.\\
Since we already provide an analysis involving the {non-central Chi-squared} distribution with spectral mixture kernels, no pathology of nonzero-mean DGPs can be shown by our analysis (Section 4.3). That is, there is no pathology as $\lambda > 0$. When $\lambda=0$, this case falls back to zero-mean or constant-mean. Mean functions greatly impact the recurrence relation because $\lambda$ is inside an exponential function. 

\begin{figure}[t]
	\centering
	\scalebox{0.75}{\input{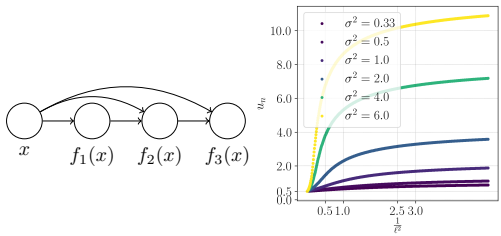}}
	\caption{\emph{Left}: Graphical model of input-connected construction suggested by~\cite{Neal_thesis,pathology_deep_gp}. \emph{Right}: The bifurcation plot of input-connected~\acrshort{dgp}.}
	\label{fig:input_connect}
\end{figure}

To the best of our knowledge, this is the first analytical explanation for the nonexistence of pathology in nonzero-mean~\dgps{}. In practice, there is existing work choosing mean functions~\cite{doubly_deep_gp}.~\cite{how_deep} briefly makes a connection between nonzero-mean~\dgps{} and stochastic differential equations. However, there is no clear answer given for this case, yet.\\
\paragraf{Input-connected~\dgps{}}
Previously, \cite{Neal_thesis,pathology_deep_gp} suggest to make each layer connect to input. The corresponding dynamic system is
\begin{equation*}
    u_n = 2m\sigma^2(1 - (1 + {u_{n-1}}/{m\ell^2})^{-m/2}) + c,
\end{equation*}
with $c$ is computed from the kernel function taking input data $\vx$.
By seeing its bifurcation plot in Figure~\ref{fig:input_connect}, we can reconfirm the solution from~\cite{Neal_thesis,pathology_deep_gp}. That is, $u_n$ converges to the value which is greater than zero, and avoids the pathology. However, the convergence rate of $\expect[Z_n]$ stays the same. 
\section{Analysis of recurrence relations}
\begin{table*}[t]
\centering
\caption{Kernel functions (middle column) and corresponding recurrence relations (right column)}
\label{tab:kernel}
\scalebox{0.8}{
    \begin{tabular}{@{}lll@{}}
    \toprule
     Rational quadratic ($\kRQ$)& $ \sigma^2\left(1 + {\normx{\vx - \vx'}{}^2}/{(2\alpha\ell^2)}\right)^{-\alpha}$ & $u_n = 2m(1 - {}_2F_{0}(\alpha;\frac{m}{2}; \frac{-u_{n-1}}{\alpha \ell^2}))$ \\ \midrule
     Periodic ($\kPer$)& $\sigma^2 \exp\left(-\frac{2\sin^2\left({\pi\normx{\vx - \vx'}{2}}/{p}\right)}{\ell^2}\right)$  & $u_n = \frac{2m\sigma^2}{\ell^2} \left(1 - {}_1F_{1}(\frac{m}{2},\frac{1}{2}, -\frac{2\pi^2}{p^2} u_{n-1})\right)$  \\ \bottomrule
    \end{tabular}
} 
\end{table*}

This section explains the condition of hyperparameters that causes the pathology for each kernel function. Then we discuss the rate of convergence for the recurrence functions. 
\subsection{Identify the pathology}
Table~\ref{tab:kernel} provides the recurrence relations of two more kernel functions: the periodic (\kPer) kernel function and the rational quadratic ($\kRQ$) kernel function. The detailed derivation is in Appendix~\ref{appendix:periodic} and~\ref{appendix:rq}.

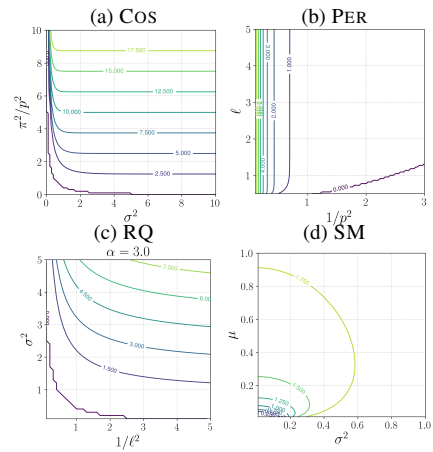
\begin{figure}[t]
    \centering
        \scalebox{0.8}{
        \begin{tikzpicture}
            \node[] at (0,0) {\includegraphics[width=0.2\textwidth]{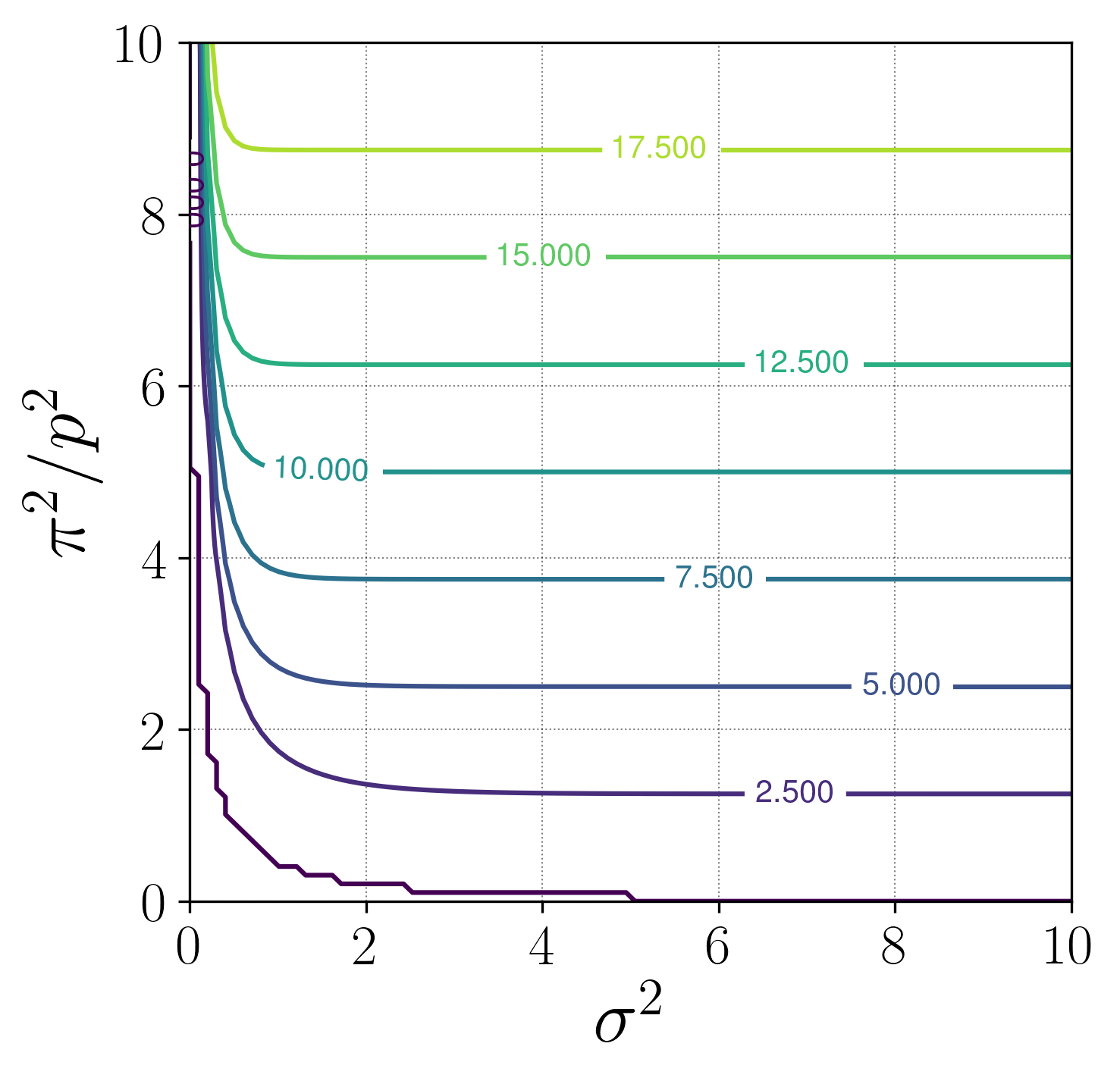}};
            
            \node[] at (3.5,0) {\includegraphics[width=0.19\textwidth]{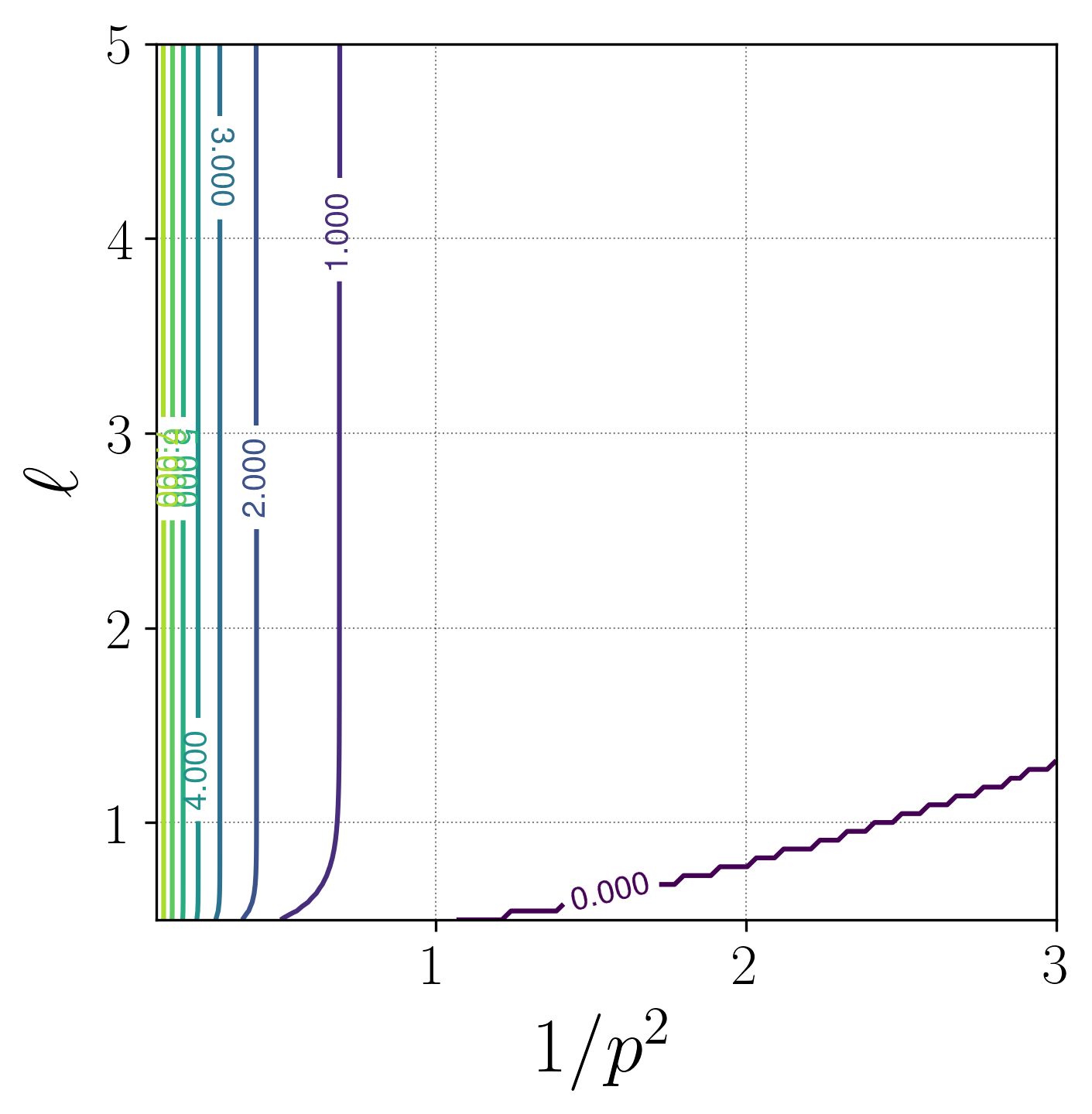}};
            
            \node[] at (0,-3.7) {\includegraphics[width=0.185\textwidth]{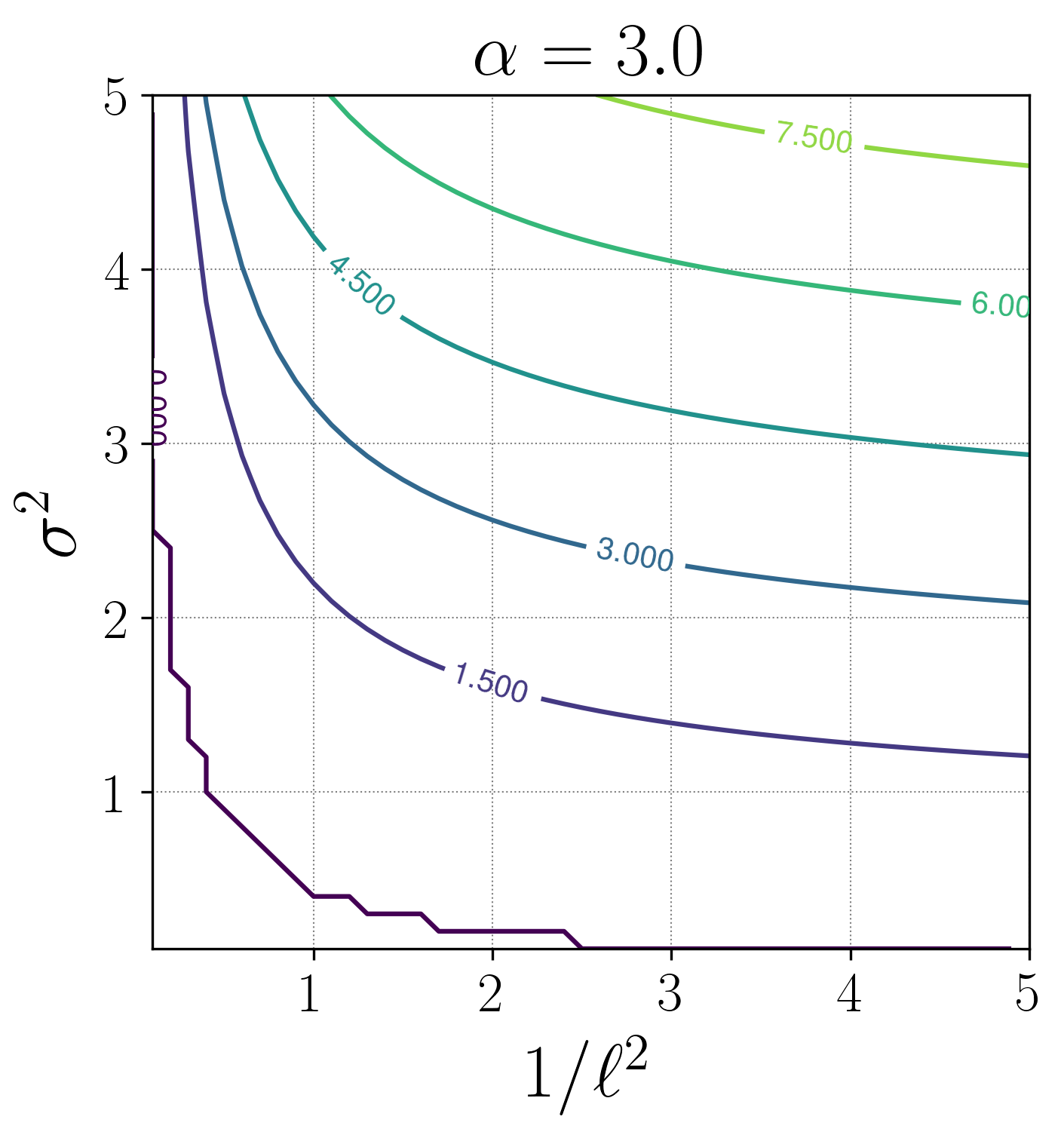}};
            \node[] at (3.5,-3.7) {\includegraphics[width=0.2\textwidth]{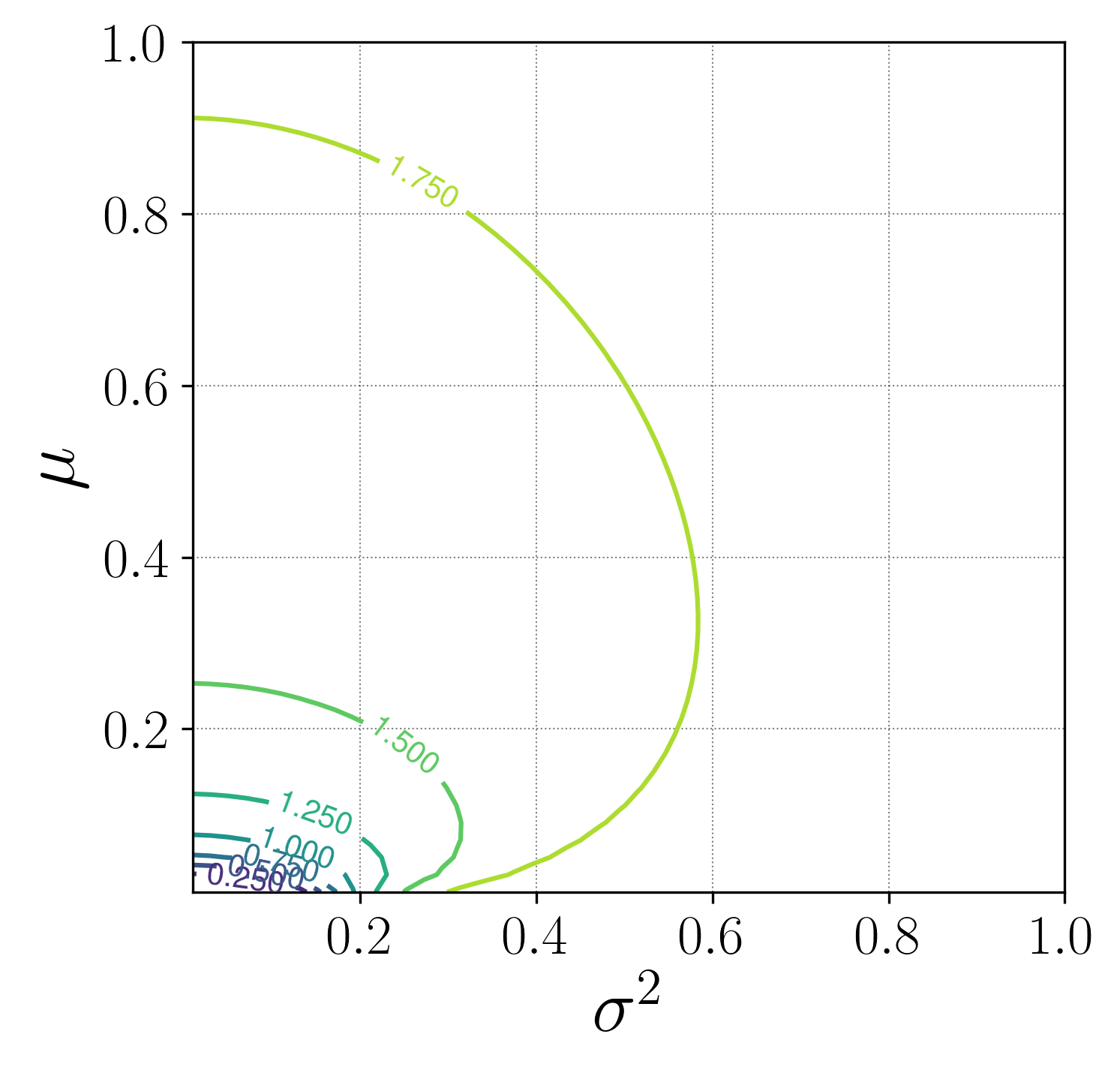}};
            
            \node[] at (0.1, 1.8) {(a) $\kCos$};
            \node[] at (3.6, 1.8) {(b) $\kPer$};
            \node[] at (0.1, -1.8)  {(c) $\kRQ$};
            \node[] at (3.6, -1.8) {(d) $\kSM$};
        \end{tikzpicture}
        } 
        \caption{Contour plots of $\expect[Z_n]$ at $n=300$ with respect to four kernel functions. }
        \label{fig:gathering all}
\end{figure}

\noindent Figure~\ref{fig:gathering all} shows contour plots based on our obtained recurrence relations. This will help us identify the pathology for each case. The corresponding bifurcation plots are in Appendix~\ref{appendix:more_bifurcation}.\\
\textbf{$\kCos$ kernel \hspace{0.2cm}} Similar to $\kSE$, the condition to escape the pathology is $\pi^2\sigma^2 / p^2 > 1$. \\
\textbf{$\kPer$ kernel\hspace{0.2cm}} If we increase $\ell$, then we should decrease the periodic length $p$ to prevent the pathology.\\
\textbf{$\kRQ$ kernel\hspace{0.2cm}} The behavior of this kernel resembles that of~\kSE. We also observe that the change in the hyperparameter $\alpha$ does not affect the condition to avoid the pathology (Appendix~\ref{appendix:more_bifurcation}, Figure~\ref{fig:appendix_rq}).\\
\textbf{$\kSM$ kernel\hspace{0.2cm}} Interestingly, this kernel does \emph{not} suffer the pathology. If $(\sigma^2, \mu)$ goes to $(0, 0)$, $\expect[Z_n]$ approaches to $0$. However, $\expect[Z_n]$ is never equal to $0$ since both $\sigma^2$ and $\mu$ are positive. 

\begin{figure}[t]
    \centering
    \begin{tikzpicture}
        
        \node[inner sep=0pt] at (0.1, 1.6) {(a)};
        \node[inner sep=0pt] at (0, 0) {\includegraphics[width=0.13\textwidth]{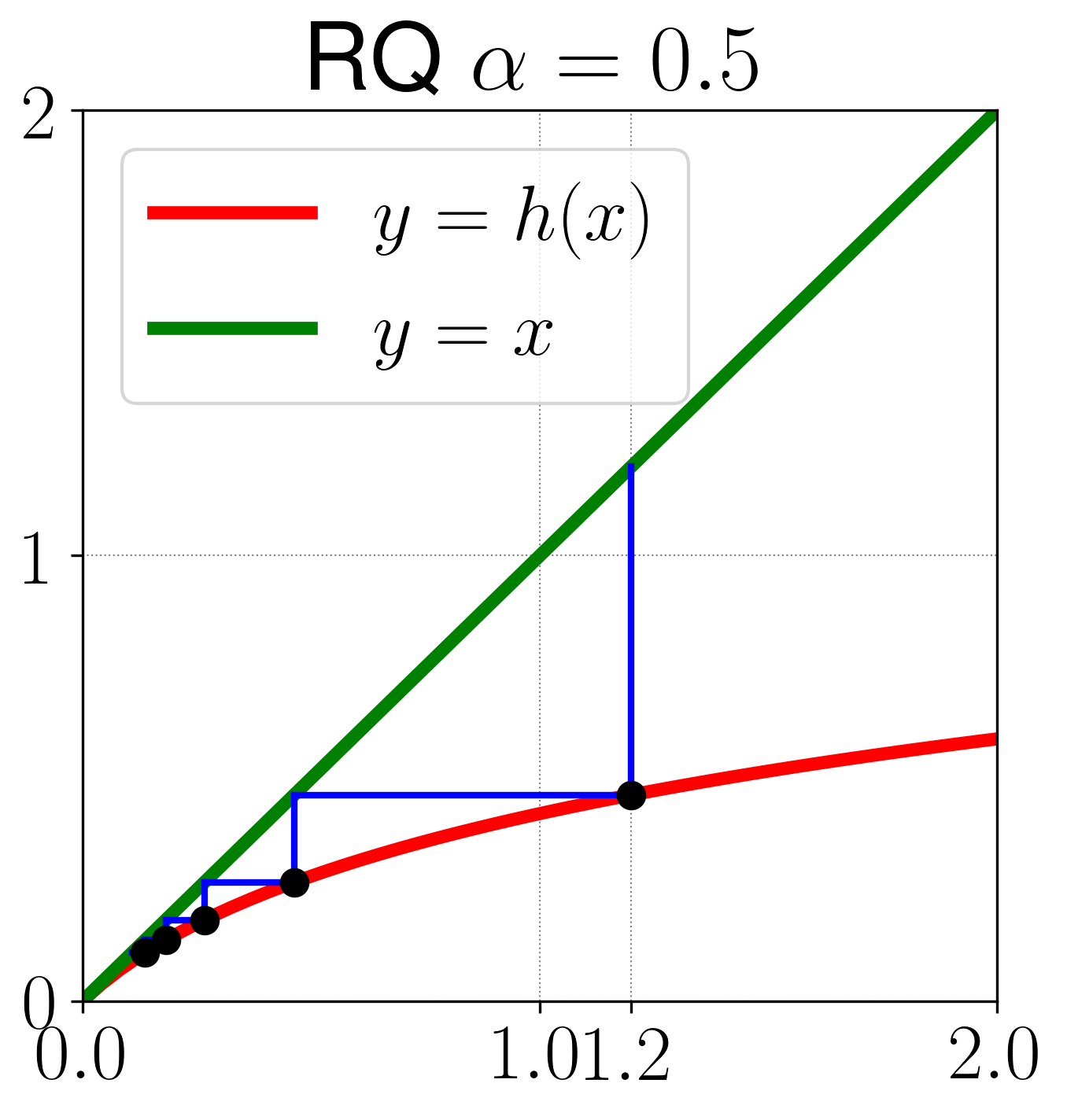}};
        
        \node[inner sep=0pt] at (2.6, 1.6) {(b)};
        \node[inner sep=0pt] at (2.5, 0) {\includegraphics[width=0.13\textwidth]{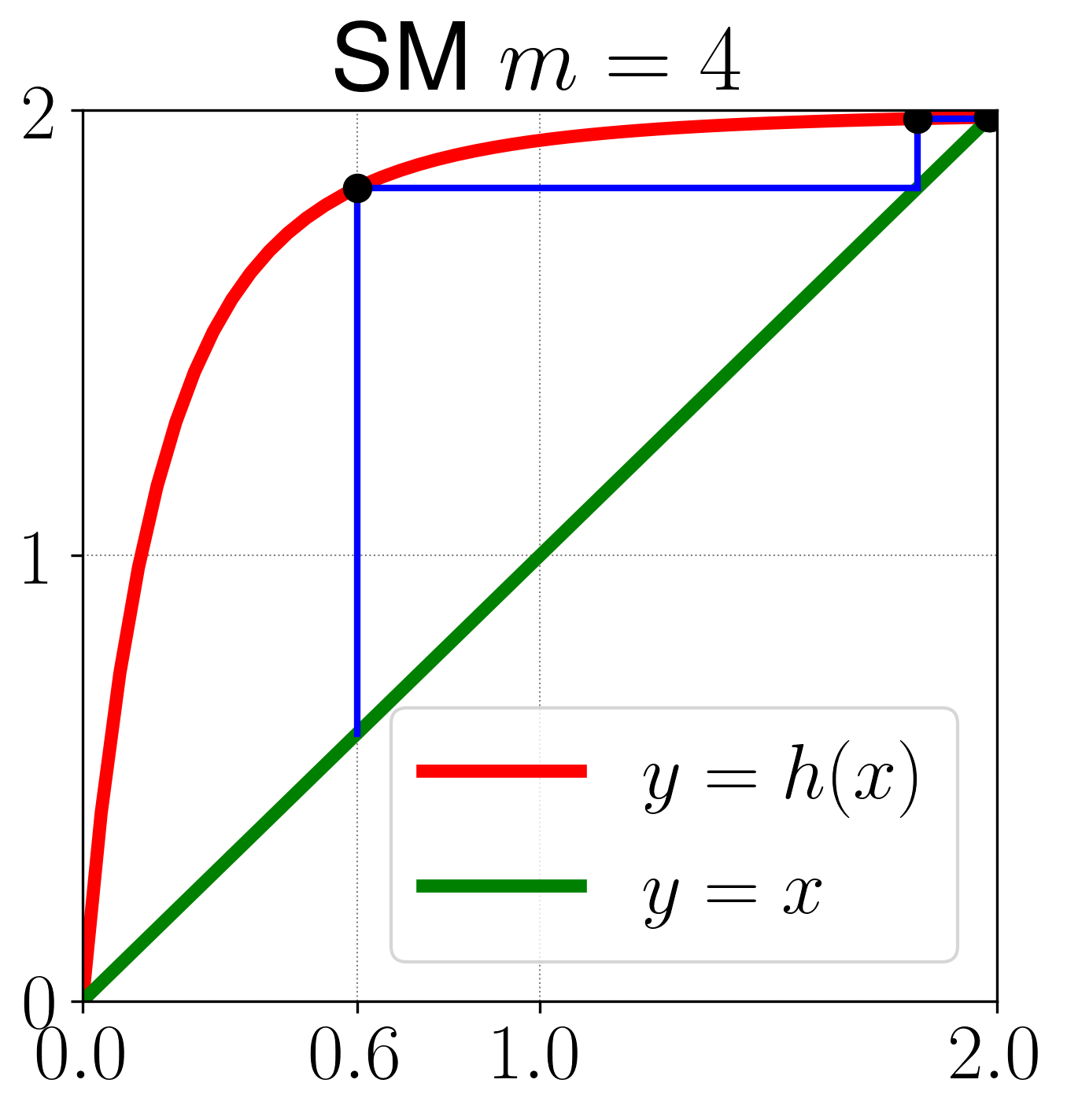}};
        
        \node[inner sep=0pt] at (5.6, 1.6) {(c)};
        \node[inner sep=0pt] at (6, 0) {\includegraphics[width=0.25\textwidth]{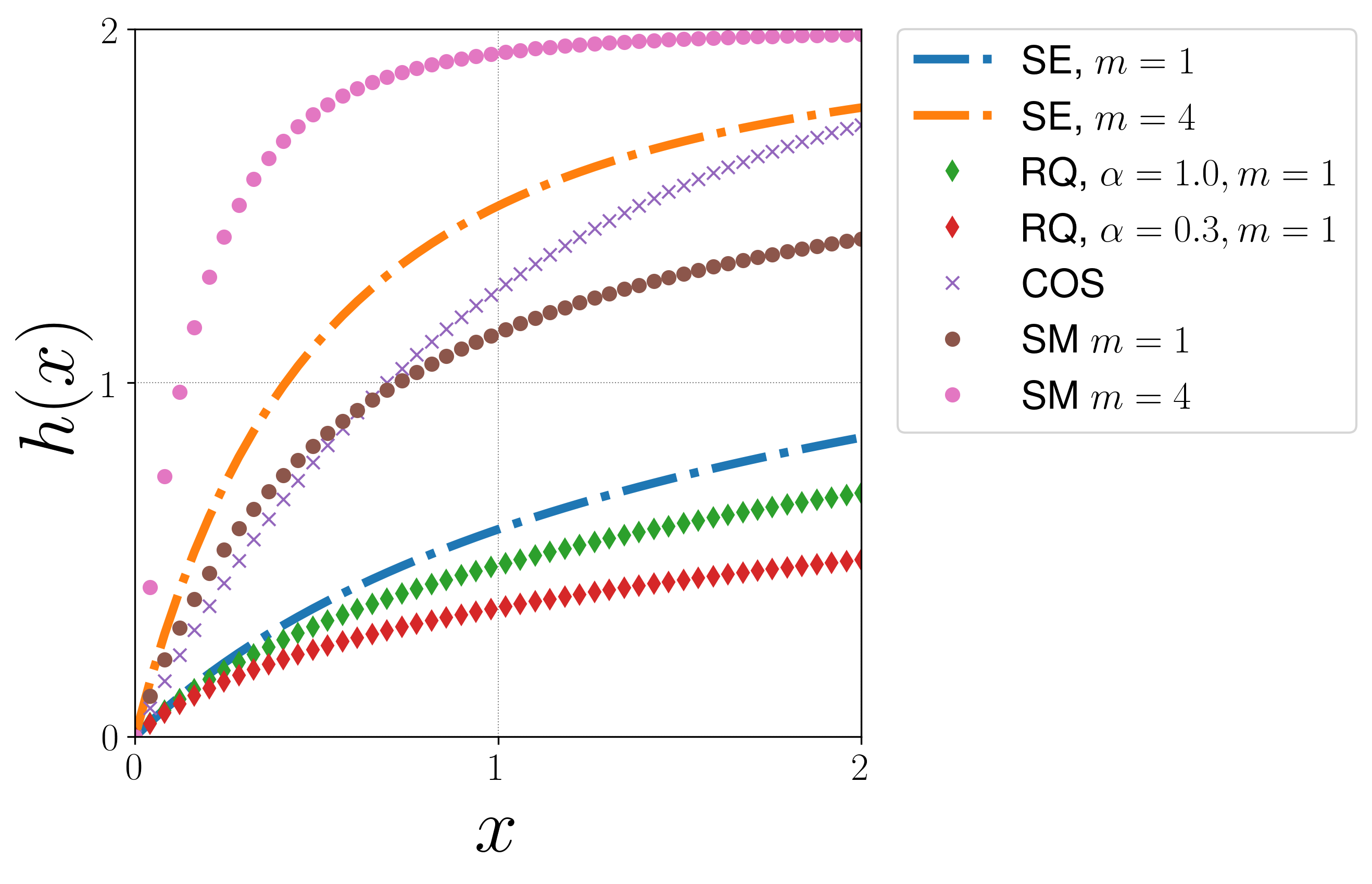}};
    \end{tikzpicture}
    \caption{ \small
    (a-b) Paths to fixed points for two cases: $\kRQ$ and $\kSM$. 
    Iterations of $\kRQ$ start from $x=1.2$ and converge to $0$. Those of $\kSM$ start from $x=0.6$ and converge to a point near $1$. (c) Plot of all recurrence functions $h(x)$. Note that $x$ is not input data but plays the role of $\expect[Z_n]$.
    }
    \label{fig:sm_and_hx}
\end{figure}
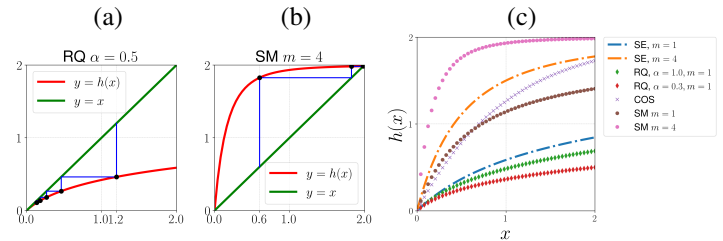

\subsection{Rate of convergence}
Recall that $h(\cdot)$ is the function modeling the recurrence relation between $\expect[Z_{n}]$ and $\expect[Z_{n-1}]$. According to Banach fixed-point theorem~\cite{fixed_point_book}, the rate of convergence is decided by the Lipchitz constant of $h(\cdot)$, $L=\sup h'(\cdot)$. The more curved the functions are, the faster the convergence rates are (see Figure~\ref{fig:sm_and_hx}a and~\ref{fig:sm_and_hx}b). Figure~\ref{fig:sm_and_hx}c compares the recurrence relation under the function $h(x)$. Specifically, for $\kSE$, the rate of convergence to a fixed point depends on the dimension parameter $m$. In general, $\kSM$ has the fastest convergence rate among all. On the other hand, the class of $\kRQ$ kernels has the slowest rate. 
  
 Understanding the convergence rate to a fixed point of recurrence relations can be helpful. For example, if a dynamic system corresponding to a~\dgp{} model quickly reaches its fixed point, it may be not necessary to have a very deep model. This can give an intuition for designing architectures in~\dgp{} given a kernel. 

\section{Experimental results}
This section verifies our theoretical claims empirically. Firstly, we investigate the correctness of recurrence relations. Then, we check the condition avoiding pathology. Furthermore, we provide case studies in real-world data sets. All kernels and models are developed based on {GPyTorch} library~\cite{gpytorch}. 
\begin{figure}[t]
    \centering
    \scalebox{0.8}{
    \begin{tikzpicture}
             
            \node[inner sep=0pt] at (2.5,0) {\includegraphics[width=0.25\textwidth]{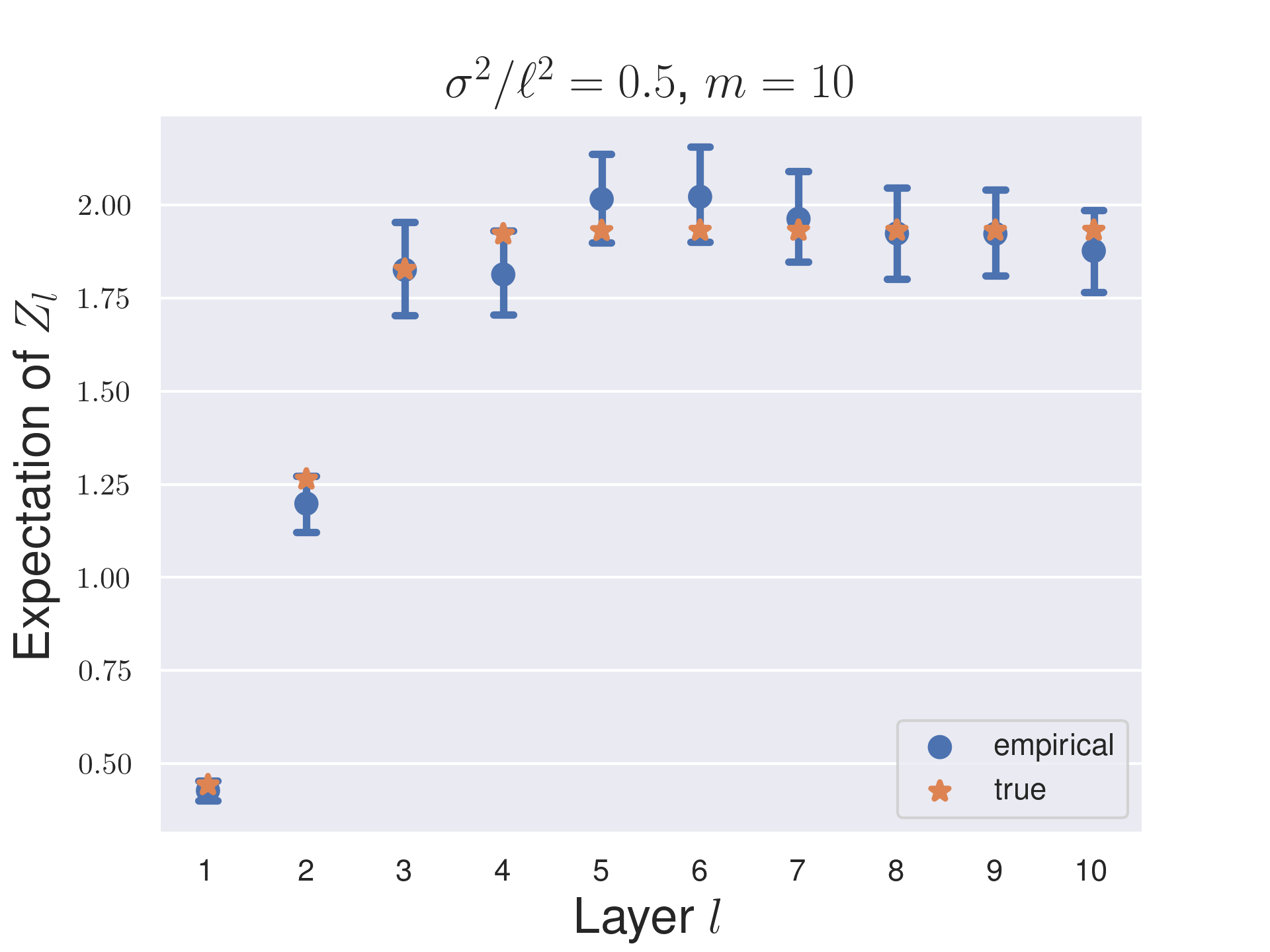}};
            
            \node[inner sep=0pt] at (7,0) {\includegraphics[width=0.22\textwidth]{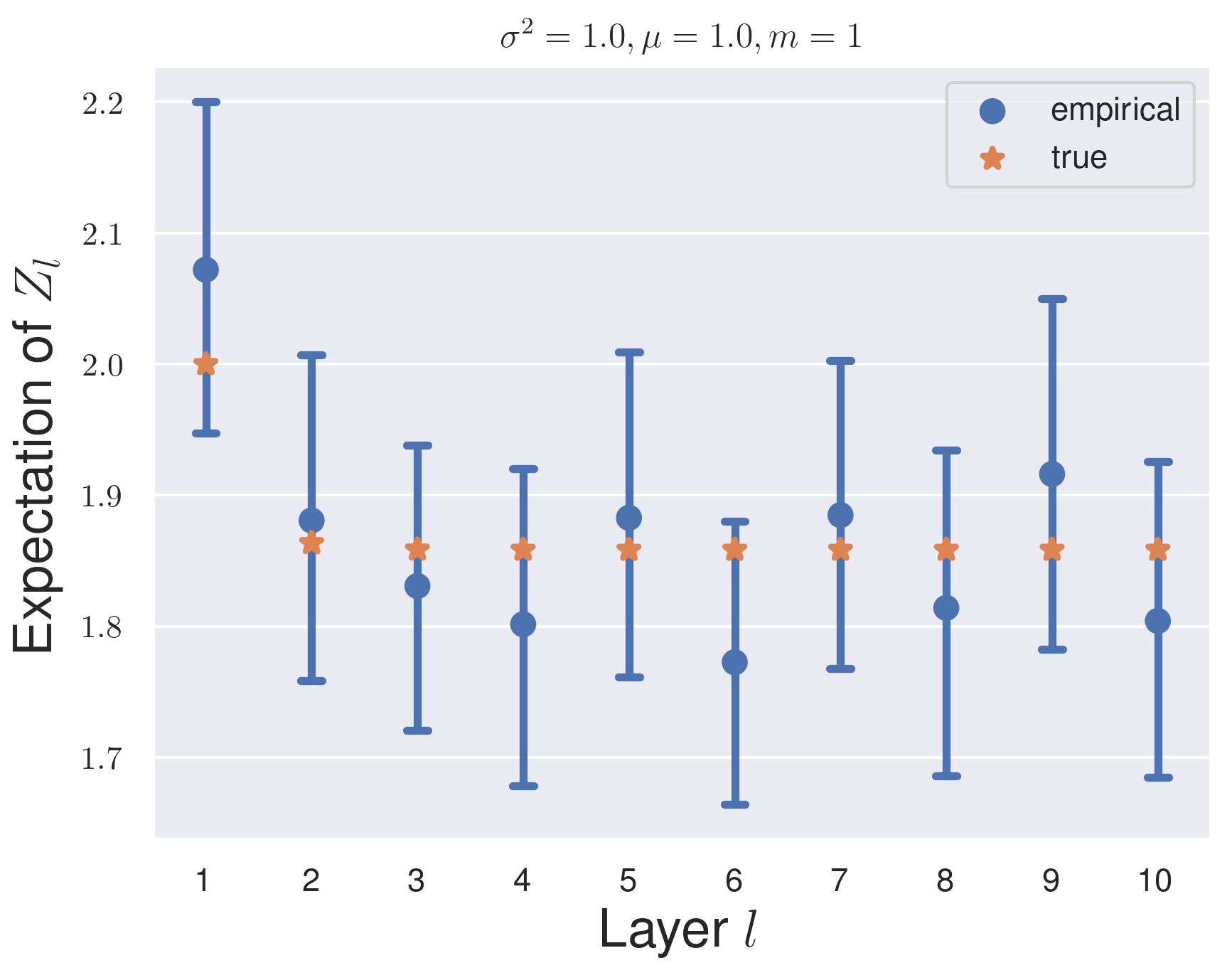}};
            
            \node[] at (2.6,1.6) {\small $\kSE$};
            \node[] at (7.1,1.65) {\small $\textsc{SM}$};
            
        \end{tikzpicture}
        } 
        \caption{\small $\expect[Z_n]$ computed from recurrence vs. empirical estimation of $\expect[Z_n]$ for two kernel functions. }
        \label{fig:tracking_expectation}
\end{figure}

\begin{figure}[t]
    \centering
    \begin{tikzpicture}
            \node[inner sep=0pt] at (0,0) {\includegraphics[width=0.24\textwidth]{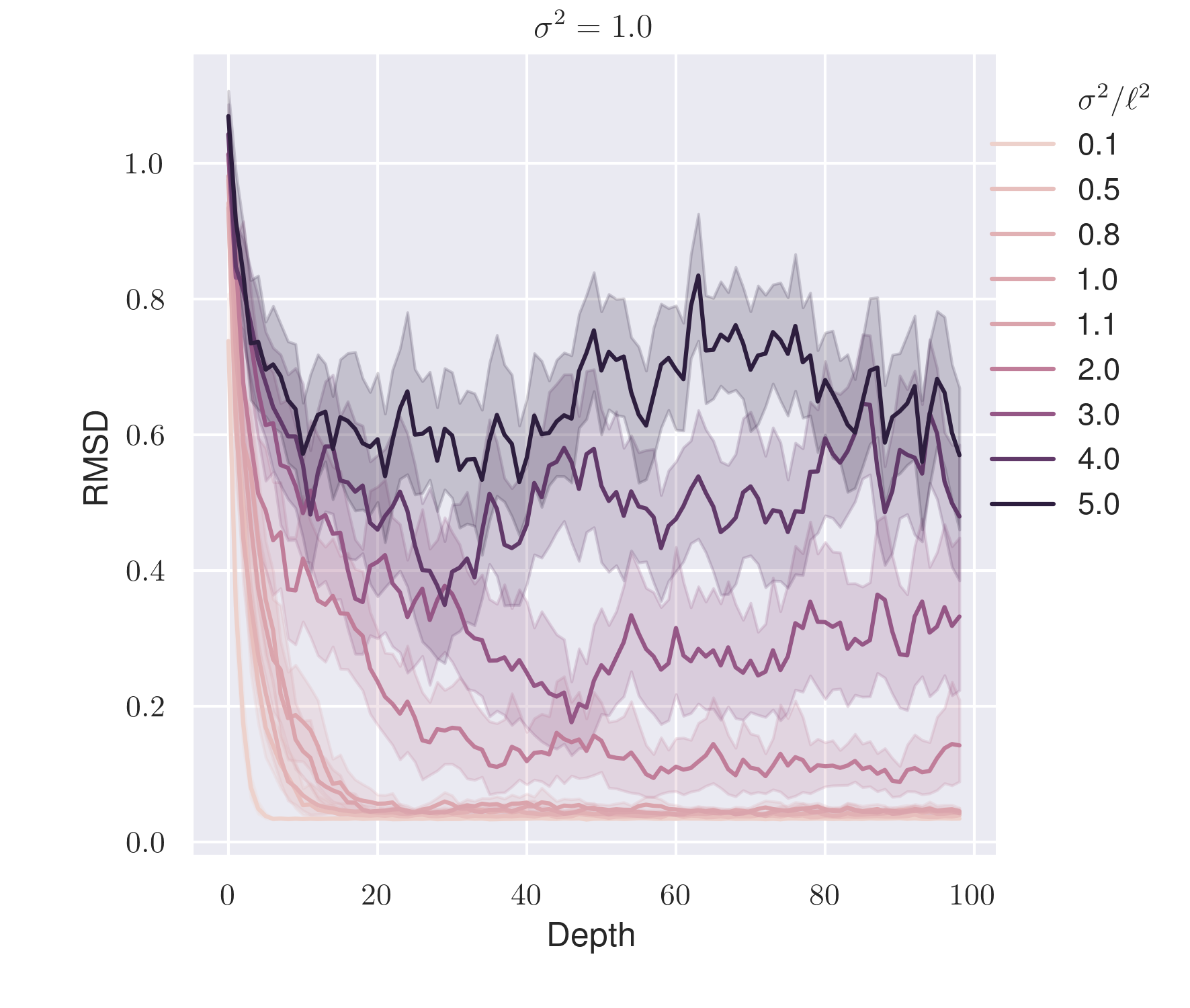}};
        
            \node[inner sep=0pt] at (4.3,0) {\includegraphics[width=0.24\textwidth]{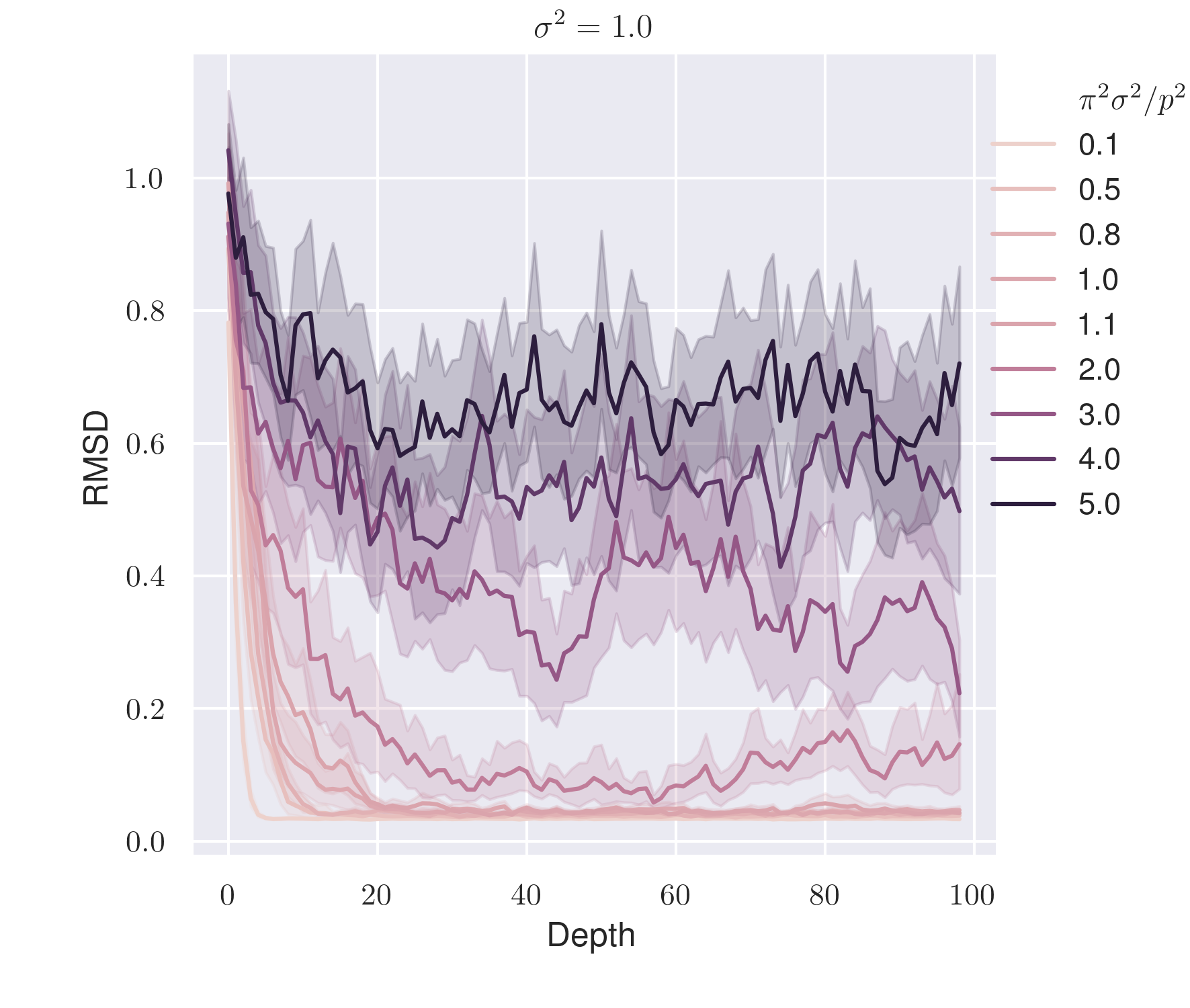}};
            
            \node[] at (0, 1.95) {\small $\kSE$};
            \node[] at (4.35, 1.95) {\small $\kCos$};
        \end{tikzpicture}
        \caption{Trace of RMSDs. RMSDs converge to $0$ when the pathology occurs. }
        \label{fig:experiment_se_cos_1d}
\end{figure}
\subsection{Correctness of recurrence relations}
\label{sec:correct_recur_relations}
We set up a~{\dgp} model with 10 layers with $\kSE$ kernel. The inputs are $x_0=0$ and $x_1=1$. We will track the value $Z_n=\normx{\vf_n(x_0) - \vf_n(x_1)}{2}^2$ for $n=1\dots10$. Given a kernel $k(x,x')$, we can exactly compute the expectations $\expect[Z_n]$. From the model, we collect $2000$ samples for each layer $n$ to obtain the empirical expectation of $\expect[Z_n]$. Then, we would like to compare the true and empirical estimates. Figure~\ref{fig:tracking_expectation} plots the comparisons for $\kSE$ kernel and $\textsc{SM}$ kernel. This numerical experiment supports the claim that our estimation $\expect[Z_n]$ is tight and even close to the true estimation. On the other hand, $\expect[Z_n]$ computed based on~\cite{how_deep} in Equation~(\ref{eq:how_deep_recur}) grows exponentially, and cannot fit in Figure~\ref{fig:tracking_expectation}. The additional plots with different settings of hyperparameter and $m$ can be found in Appendix~\ref{appendix:experiment} (Figure~\ref{fig:full_tracking_expectation_SE} and~\ref{fig:full_tracking_expectation_SM})

\subsection{Justifying the conditions of  pathology}
From $N_\text{data}$ inputs, we generate the outputs of~{\dgps} and measure the root mean squared distance (RMSD) among the outputs 
$\textrm{RMSD}(n)= \sqrt{\frac{1}{N_\text{data}(N_\text{data}-1)}\sum_{i \neq j} \normx{\vf_n(\vx_i) - \vf_n(\vx_j)}{2}^2}$. We record this quantity as we increase $n$. We replicate the procedure $30$ times to aggregate the statistics of $\textrm{RMSD}(n)$. Here, we only consider the case $m=1$.

\paragraf{$\kSE$ kernel}
We set up models in one dimension with inputs of each model in range $(-5,5)$ with $N_{data}=100$. The kernel hyperparameter $\sigma^2$ is set to $1$ while $1/\ell^2$ runs from $0.1$ to $5$. Figure~\ref{fig:experiment_se_cos_1d}a shows the trace of RMSD computed up to layer $100$. When $\sigma^2/\ell^2 > 1$, the models start escaping the pathology. 

\paragraf{$\kCos$ kernel}
With a similar setup to that of $\kSE$, Figure~\ref{fig:experiment_se_cos_1d}b shows that when $\pi^2\sigma^2/ p^2 > 1$, the models do not suffer the pathology.

\begin{figure}[t]
    \centering
    \scalebox{0.8}{
    \begin{tikzpicture}
        \node[inner sep=0pt] at (0,0) {\includegraphics[width=0.195 \textwidth]{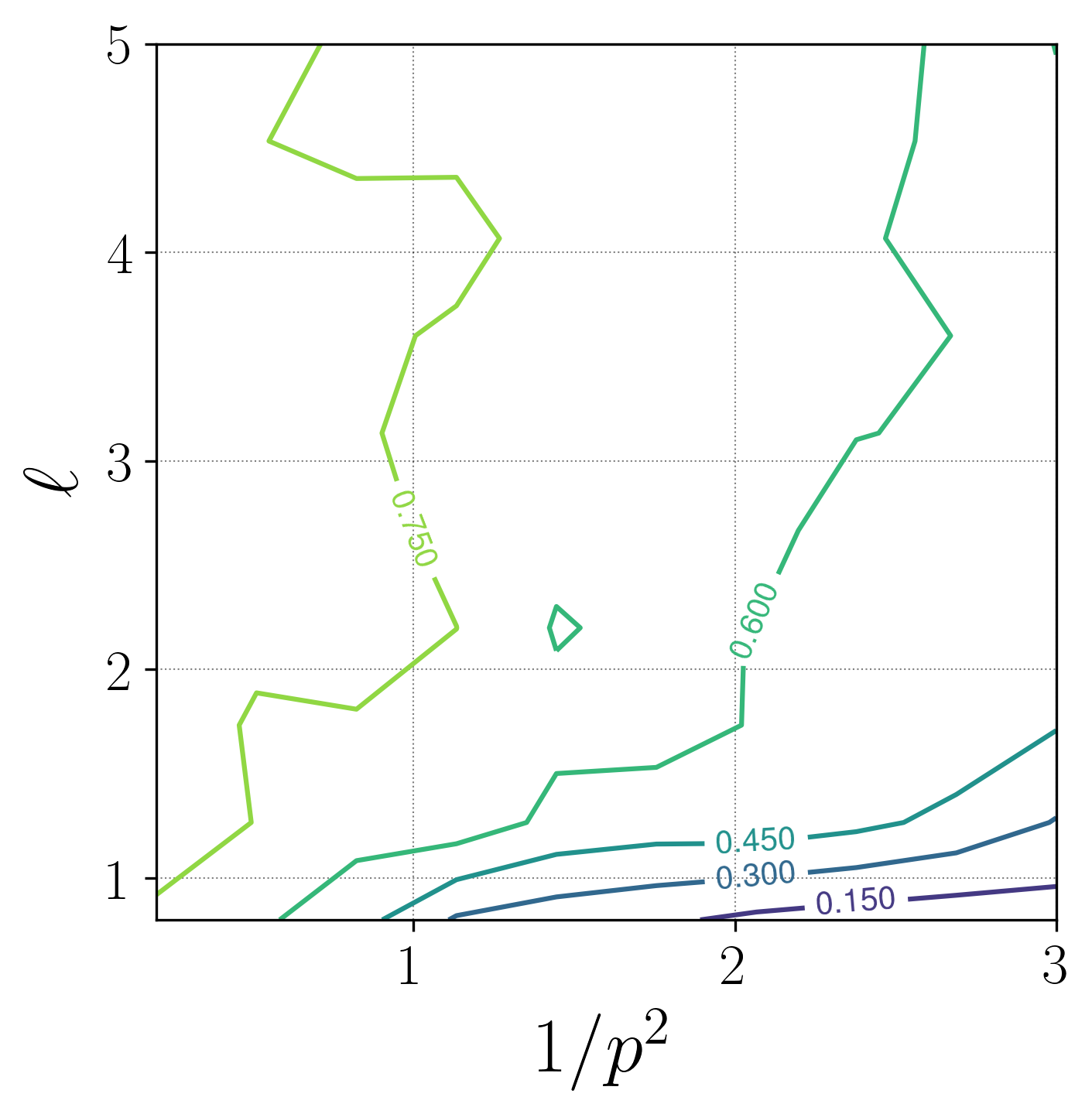}};
        
        \node[inner sep=0pt] at (3.5,0) {\includegraphics[width=0.21\textwidth]{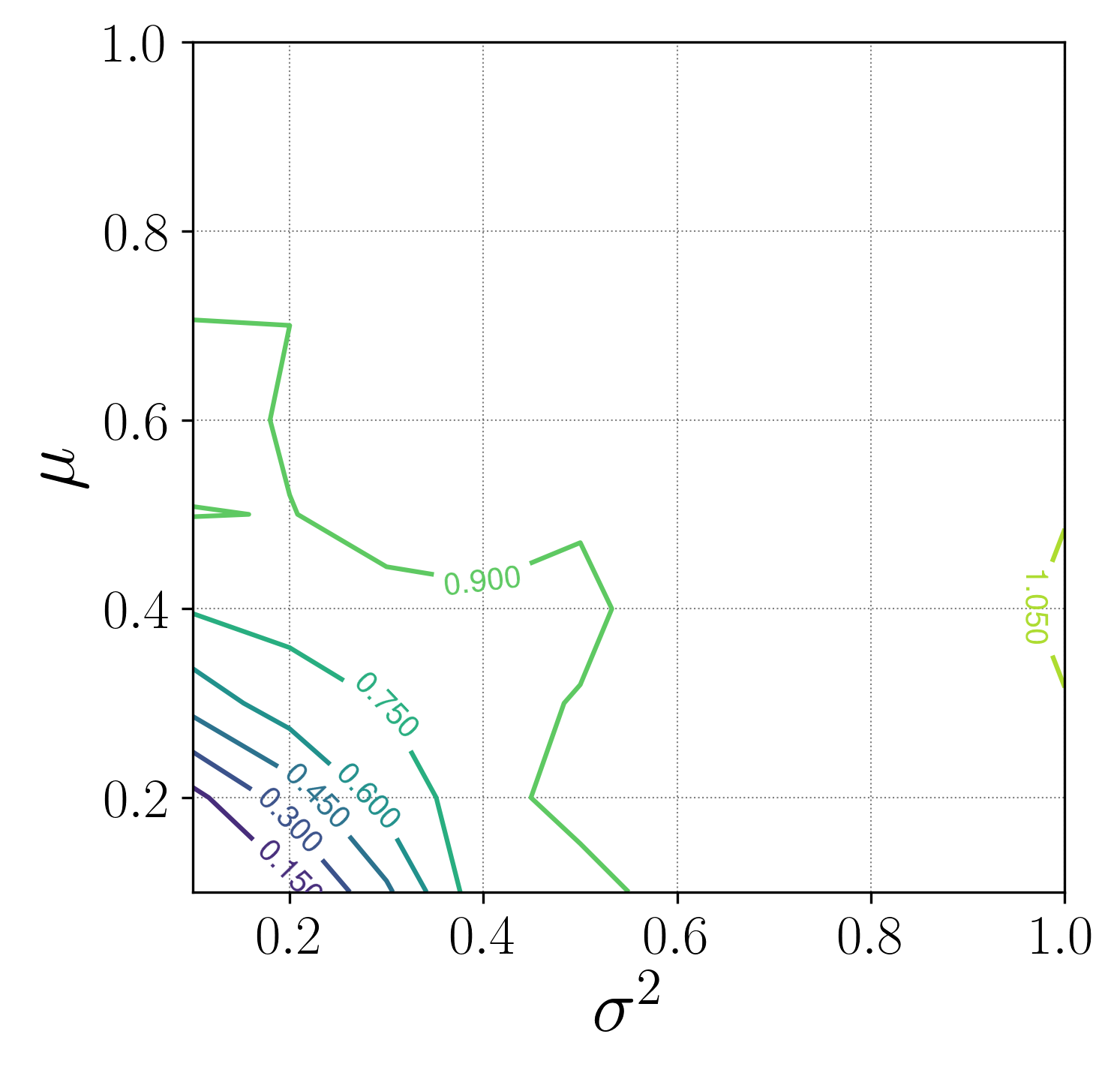}};
        
        \node[inner sep=0pt] at (0,-3.9) {\includegraphics[width=0.2 \textwidth]{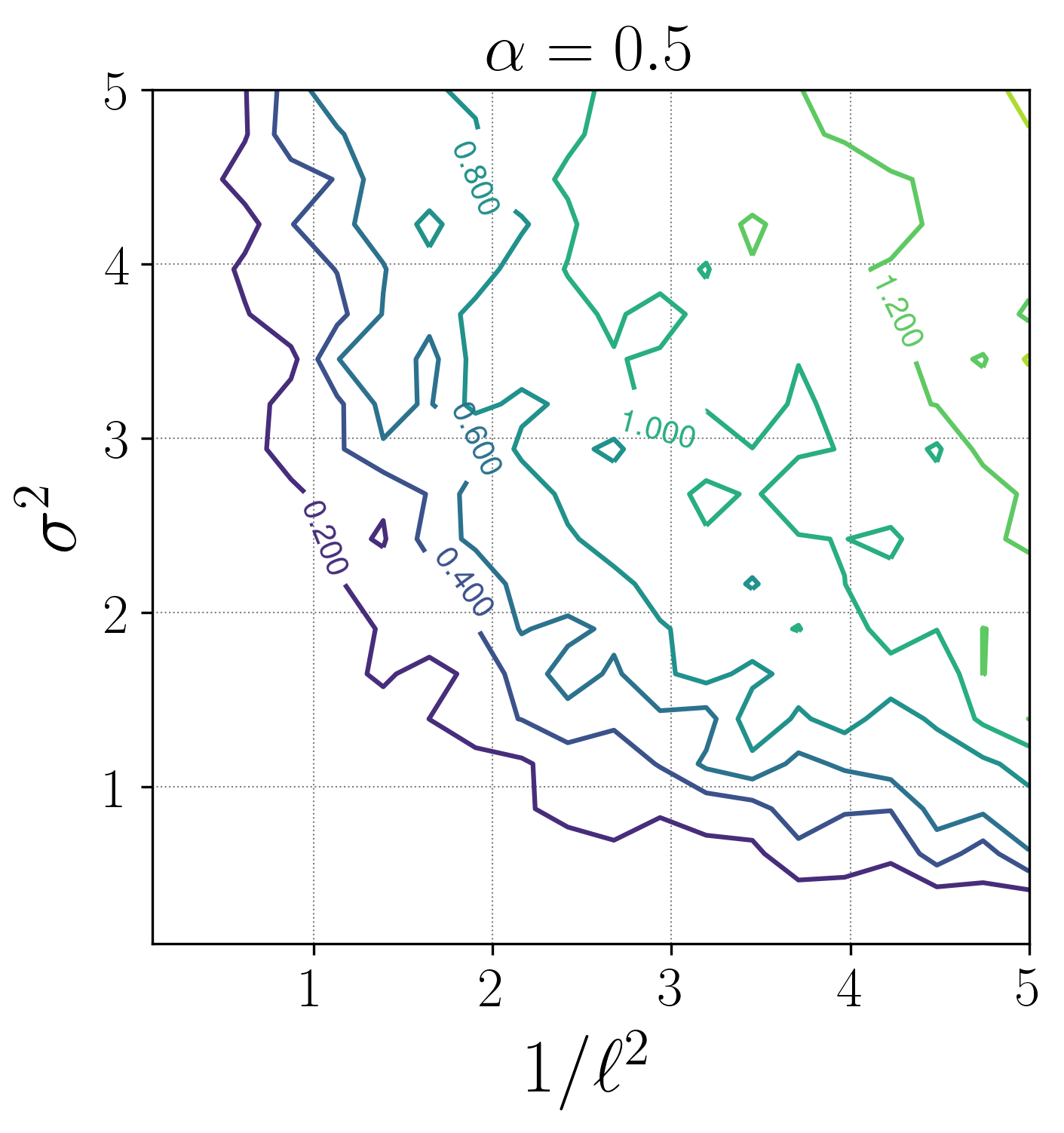}};
        
        \node[inner sep=0pt] at (3.5,-3.9) {\includegraphics[width=0.2 \textwidth]{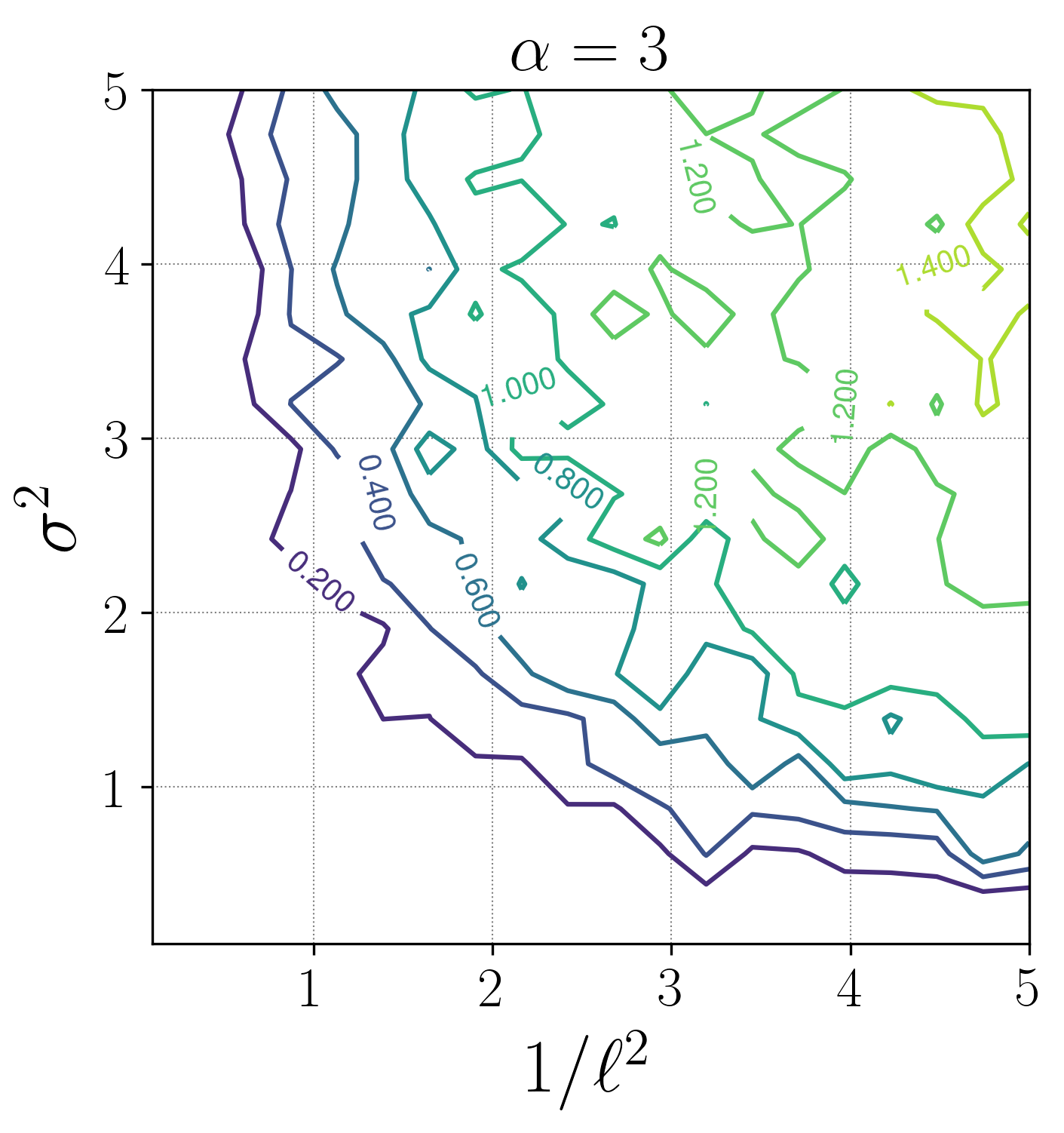}};
        
        \node[] at (0.2, 1.8) {(a) $\kPer$};
        \node[] at (3.7, 1.8) {(b) $\kSM$};
        \node[] at (0.2, -1.85) {(c) $\kRQ, \alpha=0.5$};
        \node[] at (3.7, -1.85) {(d) $\kRQ, \alpha=3$};
    \end{tikzpicture}
    } 
    \caption{Contour plots of RMSDs at layer $100$ for three kernels: $\kPer, \kSM$ and $\kRQ$.}
    \label{fig:experiment_rq_1d}
\end{figure}
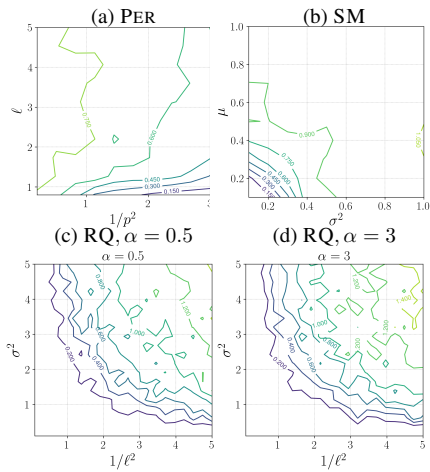

\paragraf{$\kPer$ kernel}
Since the $\kPer$ kernel has three hyperparameters, $\sigma^2, \ell^2, p$, we fix $\sigma^2$, and vary $\ell^2$ and $p$. In this case, we collected the RMSDs at layer $100$. We then compare the contour plot of these RMSDs with the values of the lower bound of $\expect[Z_n]$ computed when $n$ is large. We can find a similarity between Figure~\ref{fig:experiment_rq_1d}a and Figure~\ref{fig:gathering all}b. The lower left of both plots has low values, identified as the region that causes the pathology.

\paragraf{$\kRQ$ kernel}
Analogous to $\kPer$, only the RMSDs at layer $100$ are gathered. We chose two different values of $\alpha = \{0.5, 3\}$, and varying values of $\sigma^2$ and $\ell^2$. Figure~\ref{fig:experiment_rq_1d}c-d shows two contour plots of RMSDs for the two settings of $\alpha$. Both of the two plots share the same area of which the contour level is close to $0$. 

\paragraf{\textsc{SM} kernel\hspace{0.2cm}}
This kernel shows no sight of pathology (Figure~\ref{fig:experiment_rq_1d}b). We can find the similarity between this plot with the contour plot of $\expect[Z_n]$ in Figure~\ref{fig:gathering all}d.

\subsection{Using recurrence relations in~\dgps{}} 
Here, we use the recurrence relation as a tool to analyze~\dgp{} regression models. We learned the models where the number of layers, $N$, ranges from $2$ to $6$ and the number of units per layer, $m$, is from $2$ to $9$.  We trained our models on Boston housing data set~\cite{uci_dataset} and diabetes data set~\cite{diabete_dataset}. For each data set, we train our models with $90\%$ of the data set and hold out the remaining for testing. The inference algorithm is based on~\cite{doubly_deep_gp}. We considered two settings: (1) standard zero-mean~\dgps{} with $\kSE$ kernel; (2) the $\kSE$ kernel hyperparameters are constrained to avoid pathological regions with $\ell^2 \in (0, c_0m\sigma^2]$, constraint coefficient $ 0< c_0 < 1$.  

\begin{figure}[t]
    \centering
    \scalebox{0.8}{
    \begin{tikzpicture}
        
        
        

        \node[rotate=90] at (-6., -2.8) {\small \textbf{Standard}};
        
        \node[inner sep=0pt] at (-3.9,-2.8) {\includegraphics[width=0.22\textwidth]{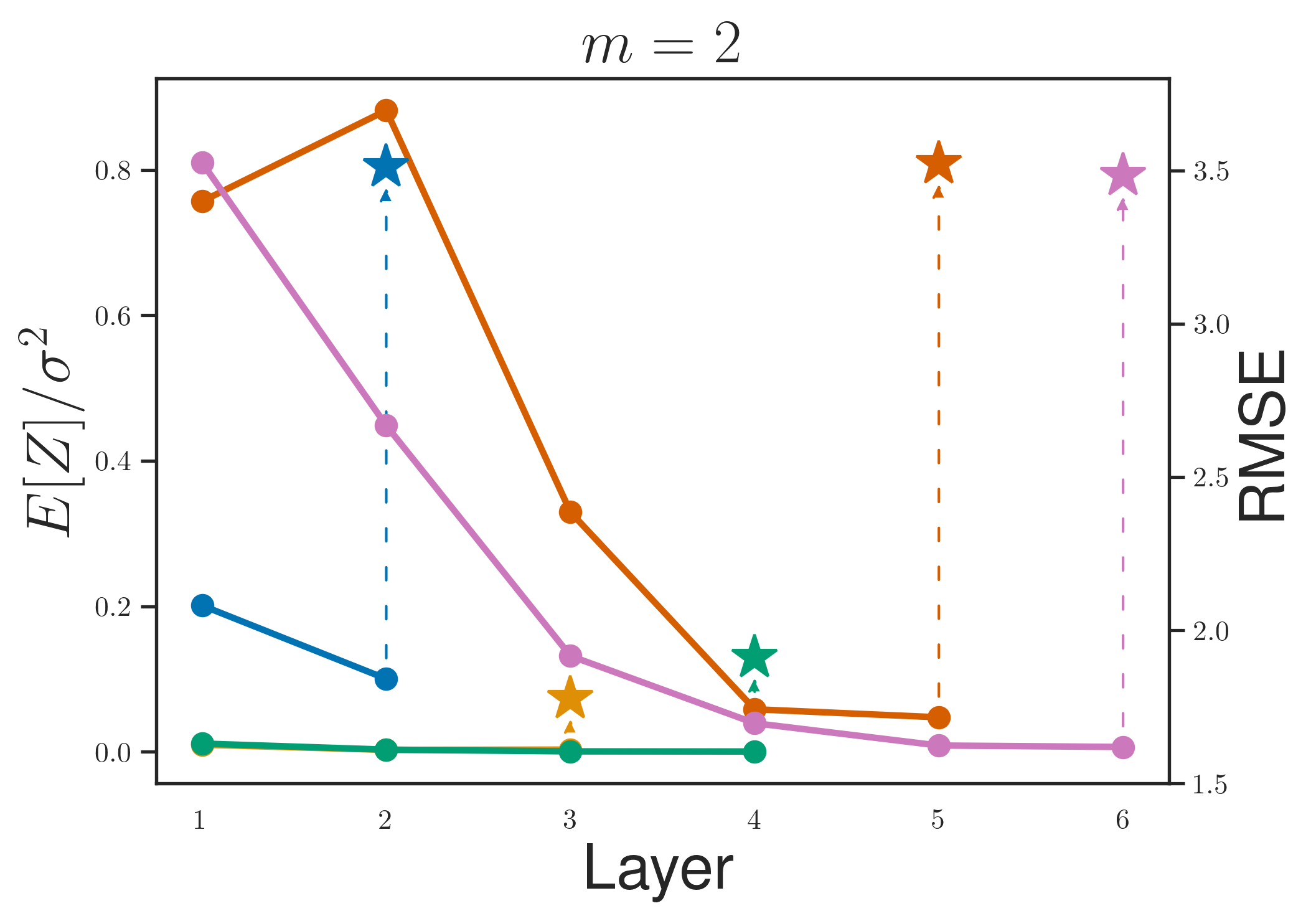}};
        
        \node[inner sep=0pt] at (0,-2.8) {\includegraphics[width=0.22\textwidth]{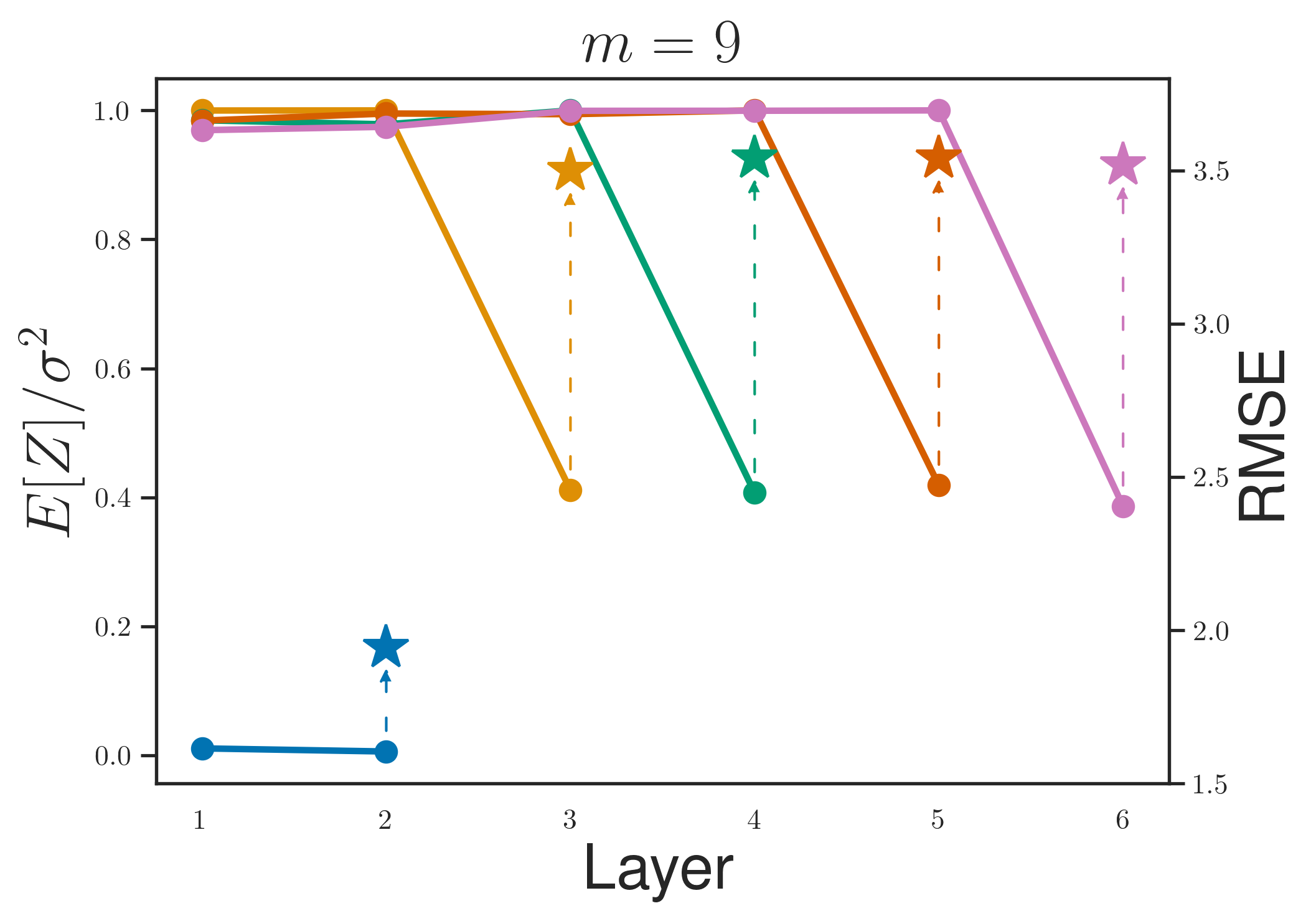}};
        
        \node[rotate=90] at (-6., -5.6) {\small \textbf{Constrained}};
        \node[inner sep=0pt] at (-3.9,-5.6) {\includegraphics[width=0.22\textwidth]{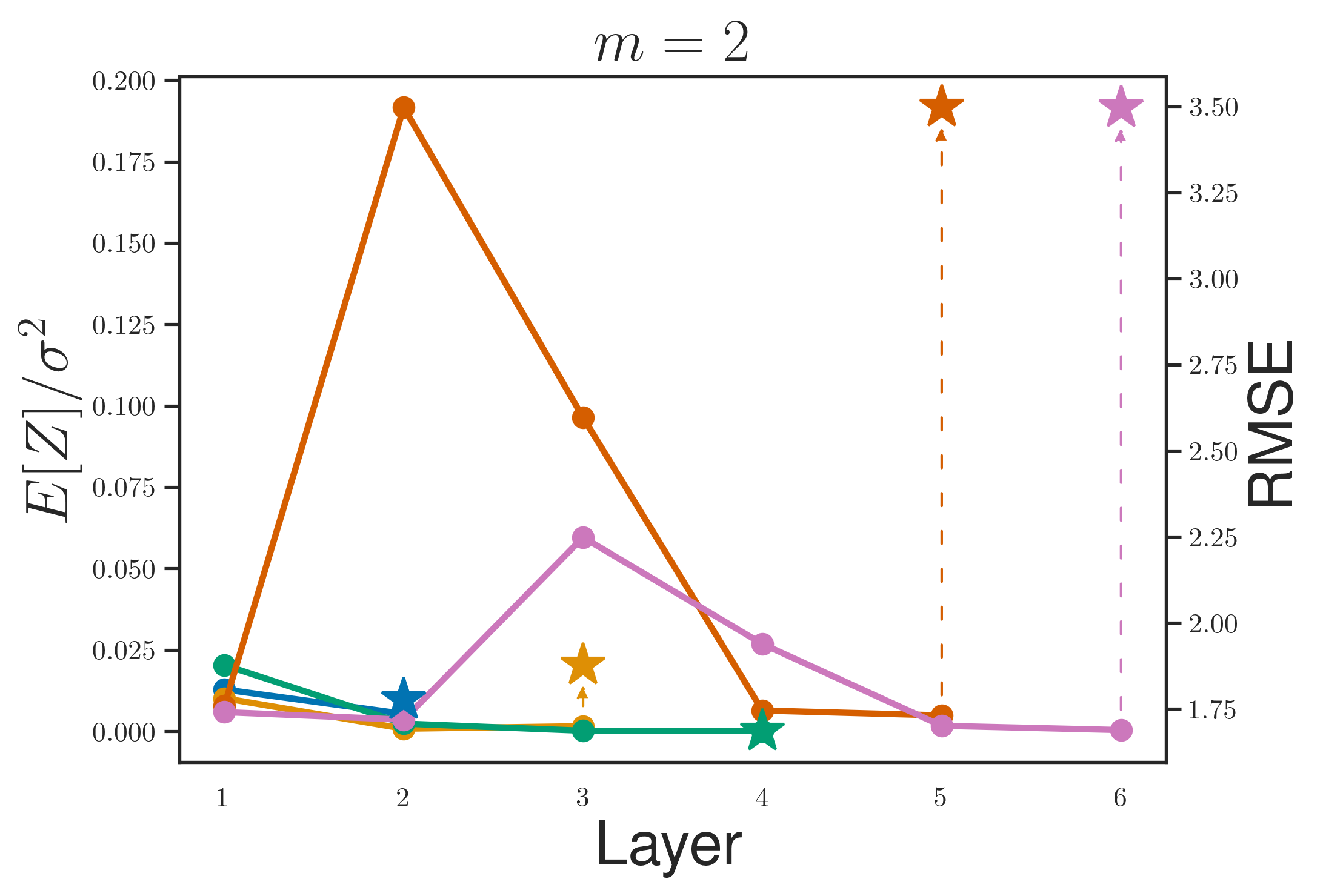}};
        
        \node[inner sep=0pt] at (0,-5.6) {\includegraphics[width=0.22\textwidth]{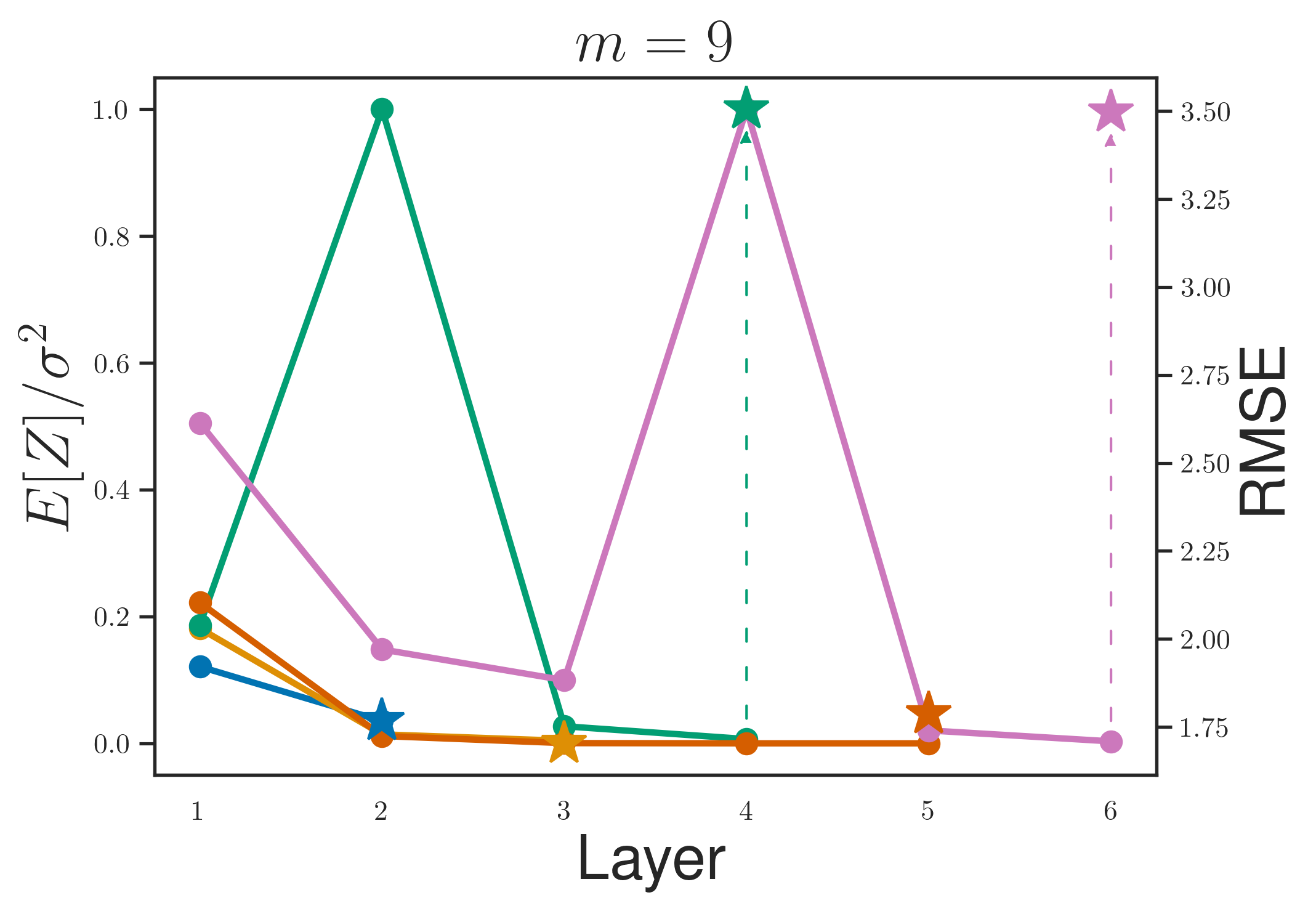}};
        
        \node[inner sep=0pt] at (2.6,-2.5) {\includegraphics[width=0.08\textwidth]{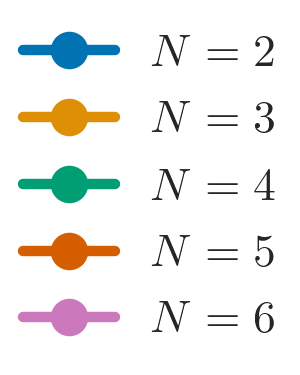}};
        
    \end{tikzpicture}
    } 
    \caption{\small Dual-axis plot of the trajectory $\expect[Z_n]/\sigma^2$ with $n$ running from $1$ to $N$ and RMSE. Solid lines indicate the trajectories of $\expect[Z_n]/\sigma^2$ projected on the left y-axis. Star markers ($\star$) indicate RMSEs projected on the right y-axis. Dashed lines connect the $\expect[Z_n]/\sigma^2$ and RMSE of the same $N$. Here, the constrain coefficient $c_0 = 0.2$. }
    \label{fig:share_kernel}
\end{figure}

Figure~\ref{fig:share_kernel} plots the root mean squared errors (RMSEs) and quantity $\expect[Z_n] /\sigma^2$ which describes changes between layers.  For the case of standard zero-mean~\dgps{}, we can observe that models can not learn effectively at deeper layers and there are drops in terms of $\expect[Z_n]/\sigma^2$ at the last layer. In the case of constraining hyperparameters, we see fewer drops and the results are improved when comparing to non-constrained cases. It seems that the drop pattern of $\expect[Z_n]/\sigma^2$ correlates to model performances. We provide detailed figures and an additional result on the diabetes data set with a similar observation in Appendix~\ref{appendix:experiment}.

\subsection{High-dimensional data set with zero-mean DGPs} 
We test on MNIST data set~\cite{lecun_mnist} with the two models like previous experiments. The number of units per layer, $m$, is chosen as $m=30$. We consider the number of layers, $N=2,3,4$.

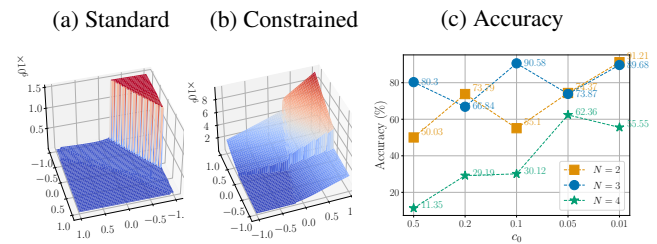
\begin{figure}[ht]
    \centering
    \begin{tikzpicture} 
    \node[inner sep=0pt] at (0, 1.7) {\small{(a) Standard}};
    \node[inner sep=0pt] at (0,0) {\includegraphics[width=0.15\textwidth]{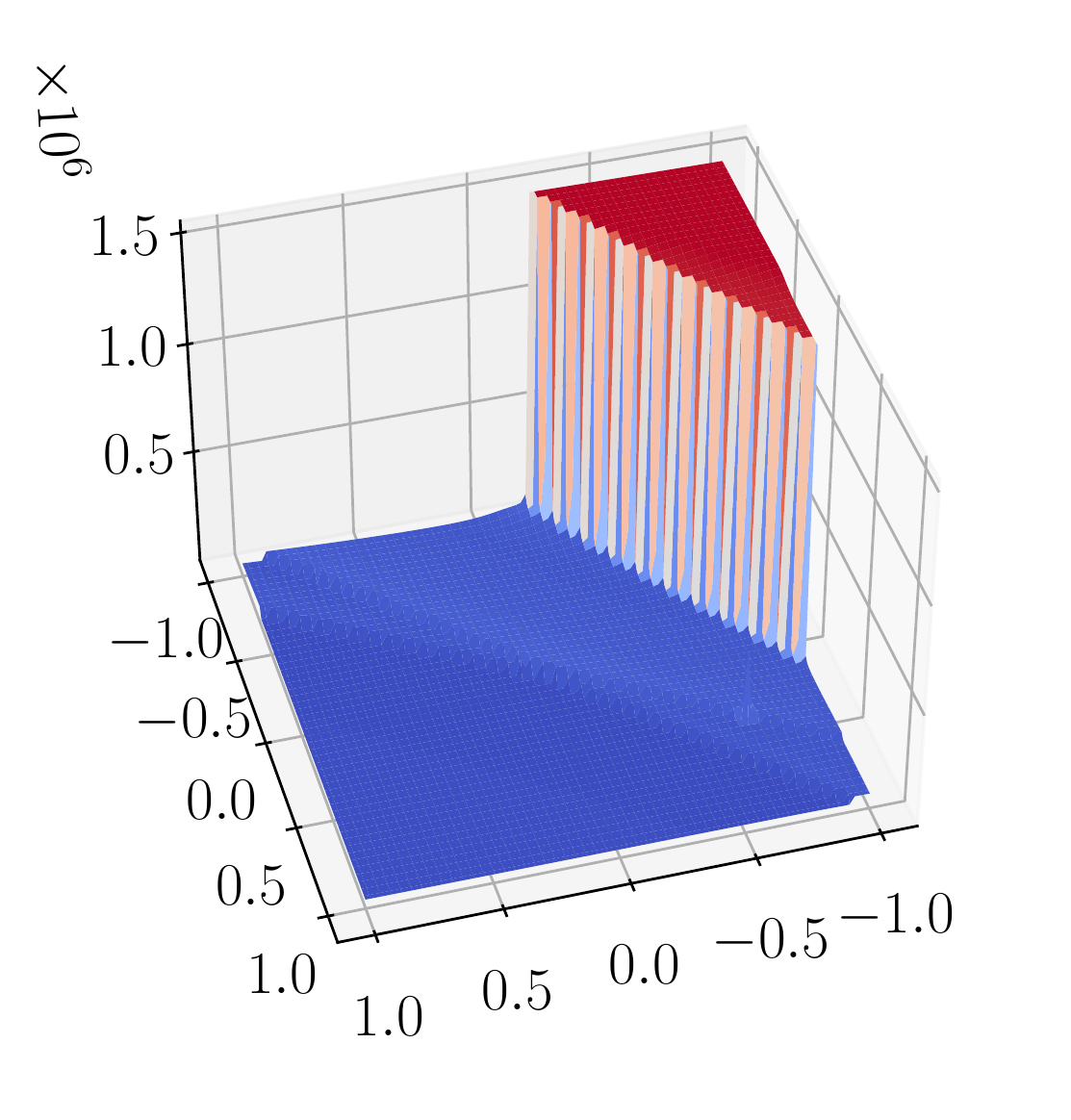}};
    
    \node[inner sep=0pt] at (2.3, 1.7) {\small{(b) Constrained}};
    \node[inner sep=0pt] at (2.3,0) {\includegraphics[width=0.15\textwidth]{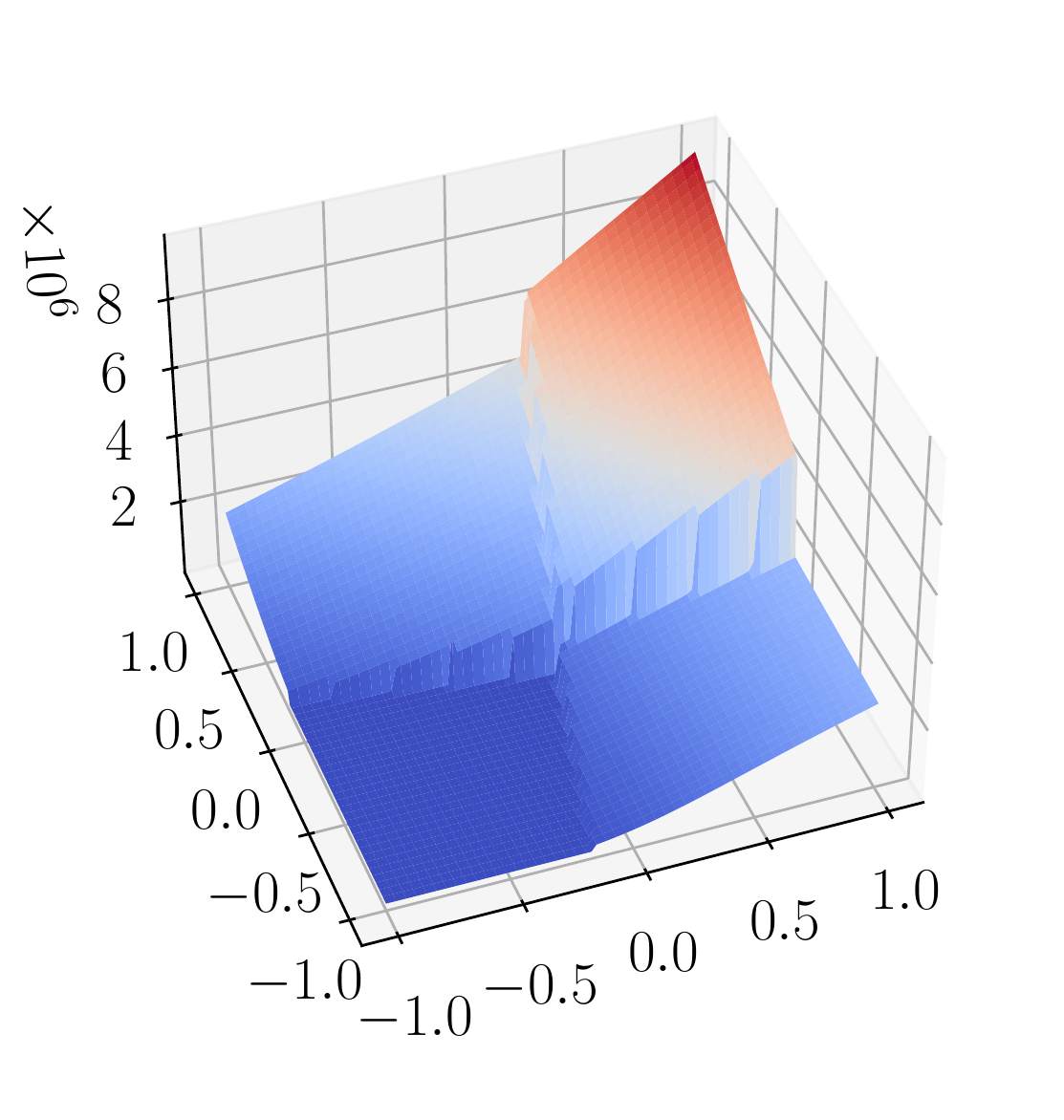}};
    
    \node[inner sep=0pt] at (5.2, 1.7) {\small{(c) Accuracy}};
    \node[inner sep=0pt] at (5.2, 0) {\includegraphics[width=0.225\textwidth]{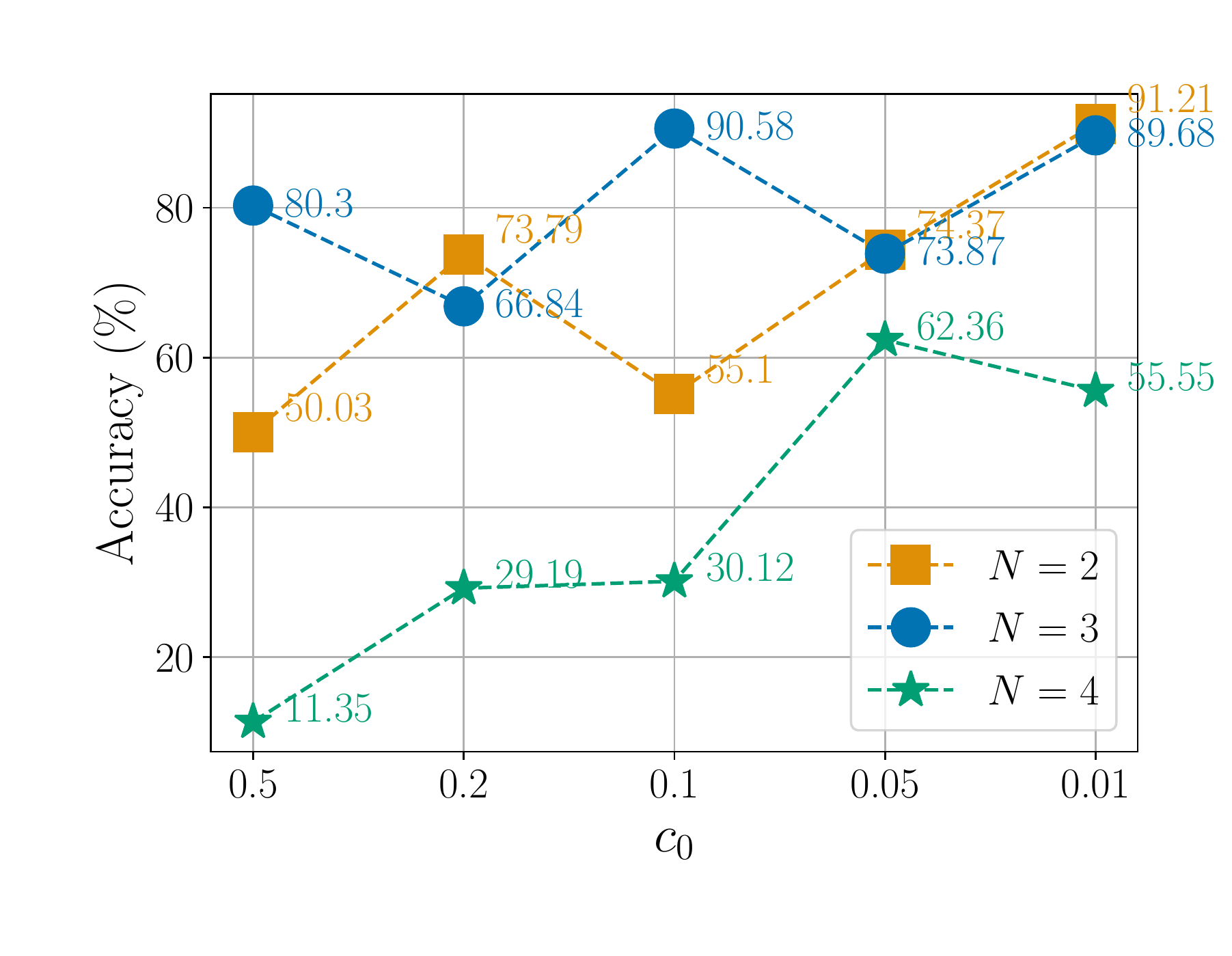}};
    \end{tikzpicture}
    \caption{\small (a-b) Loss landscape of two models. (c) Classification accuracy with respect to the number of layers, $N$, and constrain coefficients, $c_0$.}
    \label{fig:loss_landscape_accuracy}
\end{figure}
The standard zero-mean~\dgp{} without any regularization fails to learn from data with accuracy $\approx 10\%$. This means that the output of this model is just a flat function, making this 10-class classifier have such an accuracy. On the other hand, the constrained zero-mean~\dgp{} can alleviate the model performance with accuracy at best $91.21\%$. Figure~\ref{fig:loss_landscape_accuracy}c provides the results with different settings of $c_0$.\\
\indent To have a better understanding of the above models, we visualize the loss landscape~\cite{visualloss} of the two cases in Figure~\ref{fig:loss_landscape_accuracy}. The standard zero-mean~\dgp{} easily falls into unsafe pathological hyperparameters during optimization and cannot escape the unsafe state (see Figure~\ref{fig:loss_landscape_accuracy}a). In contrary, the loss landscape of constrained~\dgps{} (Figure~\ref{fig:loss_landscape_accuracy}b) shows an improved loss surface. However, we note that it still has a flat region where the optimization cannot be improved. \\
\indent Our result is not as good as the accuracy ($98.06\%$) of nonzero-mean~\dgps{} reported in~\cite{doubly_deep_gp}. However, we emphasize that the main contribution of our work is not to demonstrate the classifier performance but to show the importance of incorporating the theoretical insights into practice. This shows that learning zero-mean~\dgps{} is potentially possible.

\section{Conclusion}
We have presented a new analysis of the existing issue of~\dgp{} for a number of kernel functions via analyzing the chaotic properties of corresponding nonlinear systems which models the state of magnitudes between layers. We believe that such analysis can be beneficial in kernel structure discovery tasks~\cite{ckl,Ghahramani15_nature,hwang2016a,tong19a} for~\dgp{}. Our analysis not only provides a better understanding of the rate of convergence to fixed points but also considers a number of kernel types. Finally, our findings are verified by numerical experiments. 

\clearpage
\section*{Acknowledgements}
This work was supported by Institute of Information \& communications Technology Planning \& Evaluation (IITP) grant funded by the Korea government (MSIT)  
(No.2019-0-00075, Artificial Intelligence Graduate School Program (KAIST)).

\bibliography{ref}{}

\appendix
\onecolumn
\onecolumn

\begin{table}[t]
\centering
\caption{Summary of the recurrence relations}
\label{tab:all_kernel}
\begin{tabular}{@{}lll@{}}
\toprule

Squared exponential ($\kSE$) & $\sigma^2 \exp\left(-{\normx{\vx - \vx'}{}^2}/{2\ell^2}\right)$ & $u_n = 2m\sigma^2\left(1-(1 + {u_{n-1}}/{m\ell^2})^{-m/2}\right)$\\ \midrule

 Rational quadratic ($\kRQ$)& $ \left(1 + {\normx{\vx - \vx'}{}^2}/{(2\alpha\ell^2)}\right)^{-\alpha}$ & $u_n = 2m(1 - {}_2F_{0}(\alpha;\frac{m}{2}; \frac{-u_{n-1}}{\alpha \ell^2}))$ \\ \midrule
 Cosine ($\kCos$) & $\sigma^2 \cos\left({\pi \normx{\vx - \vx'}{2}}/{p}\right)$ & $u_n = 2\sigma^2\left(1 - {}_1F_{1}(\frac{m}{2},\frac{1}{2}, -\frac{2\pi^2}{p^2} u_{n-1})\right)$\\ \midrule
 Periodic ($\kPer$)& $\sigma^2 \exp\left(-\frac{2\sin^2\left({\pi\normx{\vx - \vx'}{2}}/{p}\right)}{\ell^2}\right)$  & $u_n = \frac{2m\sigma^2}{\ell^2} \left(1 - {}_1F_{1}(\frac{m}{2},\frac{1}{2}, -\frac{2\pi^2}{p^2} u_{n-1})\right)$  \\ \midrule
 Spectral mixture (SM) & $\exp(-2\pi^2 \sigma^2 r^2) \cos(2\pi \mu r)$ & {$\!\begin{aligned}
     u_{n} = 2m &\bigg\{1 - \exp\left(-\frac{2m\pi^2\mu^2 u_{n-1}}{1 + 4\pi^2\sigma^2 u_{n-1}}\right) \\
     &(1 + 4\pi^2\sigma^2 u_{n-1})^{-m/2} \bigg\}
    \end{aligned}$}\\ \bottomrule
\end{tabular}

\end{table}

\section{Mathematical background}
This section contains supporting theorems and results needed for proofs in the main text.
\subsection{Some probability background}

\begin{lem}[First Borel-Cantelli lemma]
	Let $E_1,E_2,...$ be a sequence of events in some probability space. If the sum of the probability of $E_n$ is finite
	$$\sum_{n=1}^\infty \mathbb{P}(E_n) < \infty,$$
	then the probability that finitely many of them occur is 0, that is,
	$$\mathbb{P}\left(\limsup\limits_{n \rightarrow \infty}E_n\right) = 0.$$
\end{lem}

\subsection{Generalized hypergeometric function}
\label{appendix:hypergeometric}
A generalized hypergeometric function is in the form of
\begin{equation*}
    {}_pF_q(a_1,a_2,\dots,a_p;b_1, b_2, \dots, b_q; z) = \sum_{n=0}^\infty \frac{a_1^{\bar{n}}a_2^{\bar{n}} \dots a_p^{\bar{n}}}{b_1^{\bar{n}}b_2^{\bar{n}} \dots b_q^{\bar{n}}}\frac{z^n}{n!}.
\end{equation*}

Here, $x^{\overline{n}}$ is the \emph{rising factorial} defined as 
\begin{equation*}
    x^{\overline{n}} = \prod_{k=0}^{n-1} (x + k).
\end{equation*}

The function ${}_1F_1(a_1, a_2; z)$ is used in the case of cosine kernel.

The function ${}_2F_0(a_1, a_2; z)$ is used in the case of rational quadratic kernel.

\section{The recurrence relation for the case of $\kPer$}
\label{appendix:periodic}

 The periodic kernel (\kPer) resembles $\kCos$ kernel, and is written in the form of
 \begin{equation*}
 \textstyle
     \kPer(\vx, \vx') = \sigma^2 \exp\left(-\frac{2\sin^2\left({\pi\normx{\vx - \vx'}{2}}/{p}\right)}{\ell^2}\right).
 \end{equation*} 
  In this case, we do not have an exact recurrence under equality. Instead, we find the lower bound of $\expect[Z_n]$. It is done by using $e^x \geq 1 + x$:
 \begin{align*}
    \exp\left(-\frac{2\sin^2(\pi r /p )}{\ell^2}\right) & \geq g(\cos(\frac{2\pi r}{p})).
 \end{align*}
where $r = \normx{\vx-\vx'}{2}$, and $g(\cos(2\pi r / p)) = 1 - \ell^{-2} + \ell^{-2} \cos(2\pi r / p)$. We can see that  the $\kPer$ kernel now is bounded  in terms of $\kCos$ kernels and use the readily obtained result of $\kCos$ kernel to get the recurrence.

The function $g$ is obtained based on $\exp(x) \geq 1 + x$:
\begin{equation*}
    g(\cos(\frac{2\pi r}{p})) = 1 -\frac{1}{\ell^2} + \frac{1}{\ell^2} \cos(\frac{2 \pi r}{p}).
\end{equation*}

The bound of $\expect[Z_n]$ is recursively computed from $\expect[Z_{n-1}]$:
\begin{equation*}
     \expect[Z_n] \leq h(\expect[Z_{n-1}|]),
\end{equation*}
where 
\begin{align*}
    h(\expect[Z_{n-1}]) = & \frac{2m\sigma^2}{\ell^2} \left(1 - {}_1F_{1}(\frac{m}{2},\frac{1}{2}, -\frac{2\pi^2}{p^2} \expect[Z_{n-1})\right).
\end{align*}

\section{Spectral mixture kernel}
\label{appendix:sm}

We consider the case the spectral mixture kernel has one-dimensional inputs and one mixture. We rewrite the kernel function as:
\begin{align*}
    \exp(-2\pi^2 \sigma^2 r^2) \cos(2\pi \mu r) &= \frac{1}{2}\{\exp(-2\pi^2 \sigma^2 r^2 +  2\pi \mu i r) + \exp(-2\pi^2 \sigma^2 r^2 -  2\pi \mu i r)\}\\
    & = \frac{1}{2} \exp(-\frac{\mu^2}{2\sigma^2}) \left\{\exp(-2\pi^2\sigma^2 (r - \frac{i \mu}{2\pi \sigma^2})^2) + \exp(-2\pi^2\sigma^2 (r + \frac{i \mu}{2\pi \sigma^2})^2)\right\}.
\end{align*}
This leads to our change in variables in the main text where we denote
\begin{equation*}
    w^2 = \exp(-\frac{\mu^2}{2\sigma^2}), \quad v^2 = 2\pi^2\sigma^2, \quad u = \frac{i \mu}{2\pi \sigma^2}.
\end{equation*}
Because $\frac{(\sqrt{Z_{n-1}} \pm i\mu / (2\pi\sigma^2))^2}{s_{n-1}}$ is distributed according to a non-central Chi-square distribution with degree of freedom $1$ and the noncentrality parameter $\lambda = - \mu^2 / (4\pi^2\sigma^4 s_{n-1})$. The moment-generating function is

\begin{equation*}
    M_{\chi'^2_1}(t) = \expect\left[\exp\left(t \frac{(\sqrt{Z_{n-1}} \pm i\mu / (2\pi\sigma^2))^2}{s_{n-1}}\right)\right] = (1 - 2t)^{-1/2}\exp(\frac{\lambda t}{1 - 2t}). 
\end{equation*}
Choosing $t = -2\pi^2\sigma^2 s_{n-1}$, we have
\begin{equation*}
    \expect\left[\exp\left(-2\pi^2\sigma^2 (\sqrt{Z_{n-1}} \pm \frac{i \mu}{2\pi \sigma^2})^2\right)\right] = (1 + 4\pi^2\sigma^2 s_{n-1})^{-1/2} \exp\left(\frac{\frac{\mu^2}{2\sigma^2}}{1 + 4\pi^2\sigma^2 s_{n-1}}\right)
\end{equation*}
We can obtain the recurrence relation as
\begin{align*}
    u_{n} &= 2 \left\{1 - \exp(-\frac{\mu^2}{2\sigma^2})\exp\left(\frac{\frac{\mu^2}{2\sigma^2}}{1 + 4\pi^2\sigma^2 u_{n-1}}\right) (1 + 4\pi^2\sigma^2 u_{n-1})^{-1/2} \right\}\\
     &=2 \left\{1 - \exp\left(-\frac{2\pi^2\mu^2 u_{n-1}}{1 + 4\pi^2\sigma^2 u_{n-1}}\right) (1 + 4\pi^2\sigma^2 u_{n-1})^{-1/2} \right\}.
\end{align*}

In the case of high-dimensional~\dgps{}, the $\kSM$ kernel takes $m$-dimensional inputs. Because all dimensions are independent, we can obtain the expectation  rely on the probabilistic independence between input dimensions to obtain the expectation as the product of the expectation in each dimension. Hence, we can have the recurrence in the form:
\begin{equation*}
    u_{n} = 2m \left\{1 - \exp\left(-\frac{2m\pi^2\mu^2 u_{n-1}}{1 + 4\pi^2\sigma^2 u_{n-1}}\right) (1 + 4\pi^2\sigma^2 u_{n-1})^{-m/2} \right\}.
\end{equation*}
Note that we assume all dimensions share the same parameter $\sigma^2$ and $\mu$. 

\section{The recurrence relation for the case of rational quadratic kernel}
\label{appendix:rq} 

Now, we study the rational quadratic ($\kRQ$) kernel. This kernel is obtained from the $\kSE$ kernel by marginalizing the inverse lengthscale of $\kSE$ kernel~\cite{Rasmussen_GPM}:
\begin{equation*}
    \kRQ(\vx, \vx') = \left(1 + {\normx{\vx - \vx'}{}^2}/{(2\alpha\ell^2)}\right)^{-\alpha}.
\end{equation*}
 We use the power series expansion $(1 + x)^{-\alpha} = \sum_{k=0}^{\infty} {-\alpha \choose k} x^k$ for this kernel
\begin{align*}
    \expect[Z_n] = 2\sigma^2 - 2\sigma^2 \sum_{k=0}^\infty {-\alpha \choose k} \frac{\expect[Z_{n-1}^k|\vf_{l-2}]}{(2\alpha\ell^2)^k}.
\end{align*}
Next, we use the high-order moment of Chi-squared distribution. As $\frac{Z_{n-1}}{s_{n-1}}|\vf_{n-2} \sim \chi^2_m$, it is known that
$
    \expect\left[\frac{Z_{l-1}^k}{s_{l-1}^k}|\vf_{l-2}\right] = 2^k \frac{\Gamma(k + \frac{m}{2})}{\Gamma(\frac{m}{2})}. 
$
Then, we can obtain the corresponding recurrence relation as \begin{align*}
    \expect\left[\left(1 + \frac{Z_{n-1}}{2\alpha \ell^2}\right)^{-\alpha} | \vf_{n-2}\right] &= \sum_{k=0}^\infty {-\alpha \choose k} \frac{\expect[Z_{n-1}^k|\vf_{n-2}]}{(2\alpha\ell^2)^k} \\
    & = \sum_{k=0}^\infty \frac{(-1)^k \alpha^{\bar{k}}}{k!} \frac{2^k \frac{\Gamma(k + m/2)}{\Gamma(m/2)}\expect[Z_{n-1}|\vf_{n-2}]^k}{(2\alpha\ell^2)^k} && \text{(use high-order moment of Chi-squared)} \\
    & =  \sum_{k=0}^\infty {\alpha^{\bar{k}} \left(\frac{m}{2}\right)^{\bar{k}}} \frac{(-1)^k\expect[Z_{n-1}|\vf_{n-2}]^k}{(\alpha\ell^2)^k k!} && \text{(by a property of rising factorial) } \\
    & = {}_2F_{0}(\alpha;\frac{m}{2}; \frac{-\expect[Z_{n-1}|\vf_{n-2}]}{\alpha \ell^2}). && \text{(by definition of hypergeometric function)}
\end{align*}
Consequently, we obtain the recurrence between layers as 
\begin{equation*}
    u_n = 2m(1 - {}_2F_{0}(\alpha;\frac{m}{2}; \frac{-u_{n-1}}{\alpha \ell^2})),
\end{equation*}
where ${}_2F_0 (\cdot;\cdot;\cdot)$ is one of the hypergeometric functions (see Definition in Appendix~\ref{appendix:hypergeometric}). This ${}_2F_0 (\cdot;\cdot;\cdot)$ function has a close connection to the exponential integral function $Ei(x) = \int_{-x}^{+\infty} \frac{e^{-t}}{t} dt$. This is related to the way of constructing the $\kRQ$ kernel from $\kSE$ kernel.

\section{More on bifurcation plots}

\label{appendix:more_bifurcation}
Figure~\ref{fig:se_bifurcation_m_2_3} shows the bifurcation plots of $\kSE$ kernel in high-dimensional cases ($m=2,3$).

\begin{figure}
    \centering
    \includegraphics[width=0.8\textwidth]{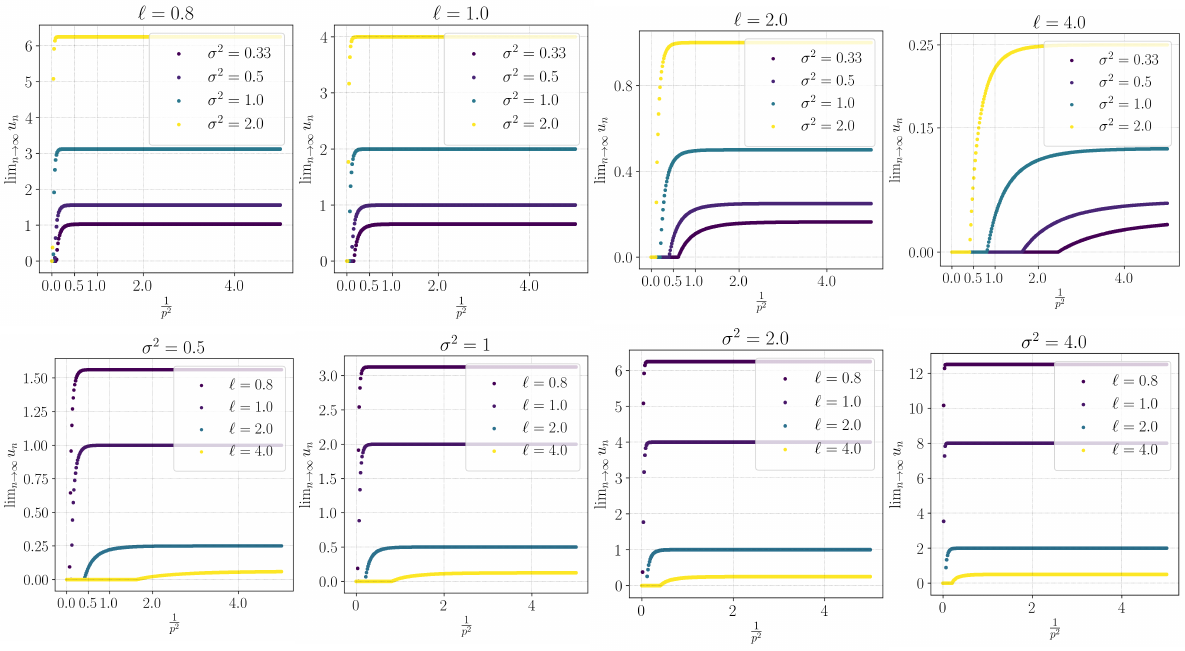}
    \caption{Bifurcation plot of the recurrence of periodic kernel for $m=1$. First row: From left to right, $\ell$ is varied. Second row: $\sigma^2$ is varied.}
    \label{fig:per_bifurcation}
\end{figure}

Figure~\ref{fig:per_bifurcation} shows the bifurcation plots of $\kPer$ kernel.

\begin{figure}[t]
    \centering
    \begin{tikzpicture}
        \node[] at (0,0) {\includegraphics[width=0.2\textwidth]{figure/rq_alpha_3_contour.png}};
        
        \node[] at (4,0) {\includegraphics[width=0.2\textwidth]{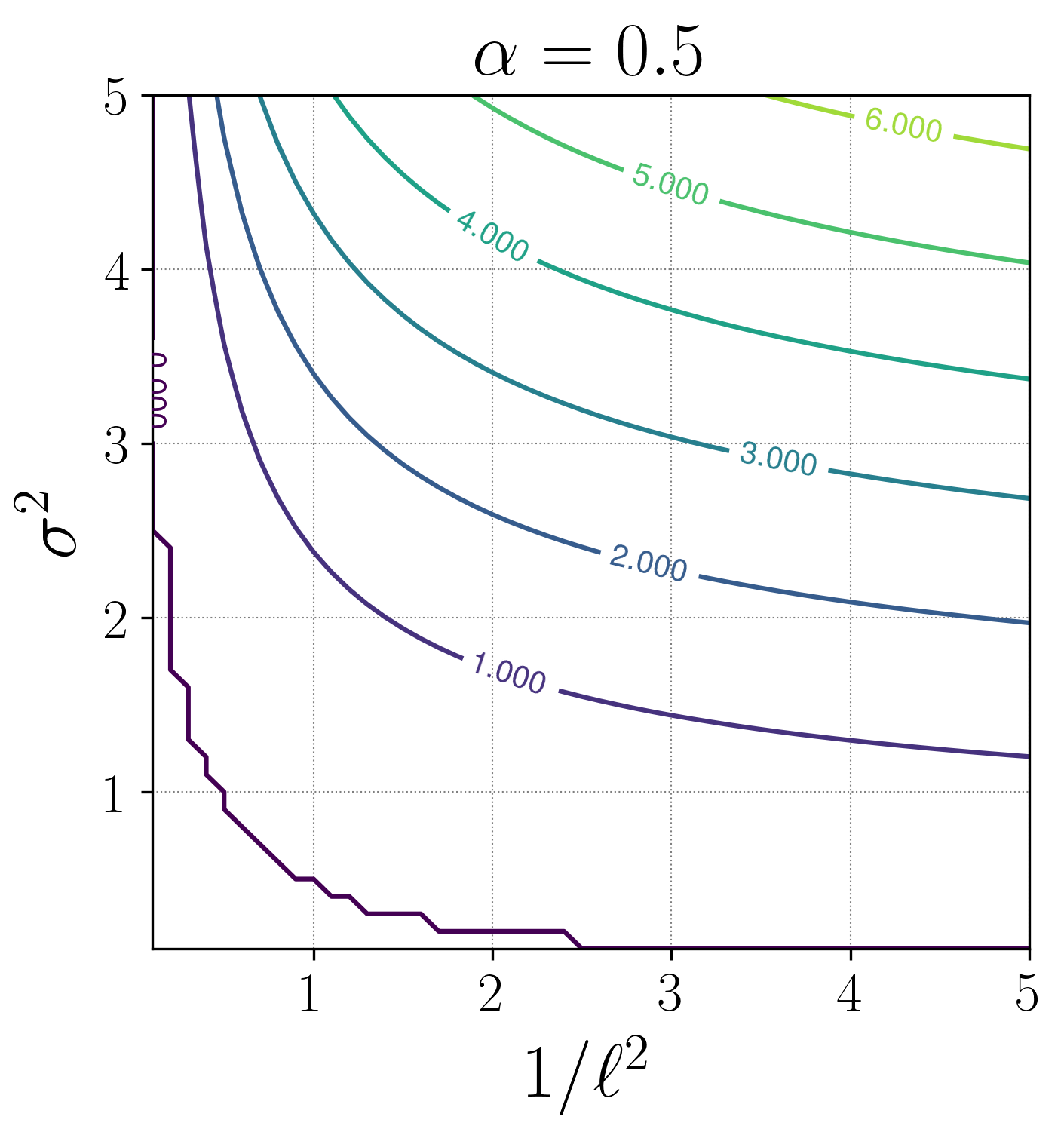}};
    \end{tikzpicture}
    \caption{$\kRQ$: Contour plots of $\expect[Z_n]$ at $n=300$. The two contour plots share the same zero-value level. So that $\alpha$ does not decide the condition overcome the pathology.}
    \label{fig:appendix_rq}
\end{figure}

Figure~\ref{fig:appendix_rq} shows the contour plots of $\kRQ$ kernel for two cases of $\alpha$, indicating that changing $\alpha$ does not affect the condition to escape the pathology.


\begin{figure}
    \centering 
    \scalebox{1.5}{
    \begin{tikzpicture}
    \node[] at (0, 0) {
        \includegraphics[width=0.115\textwidth]{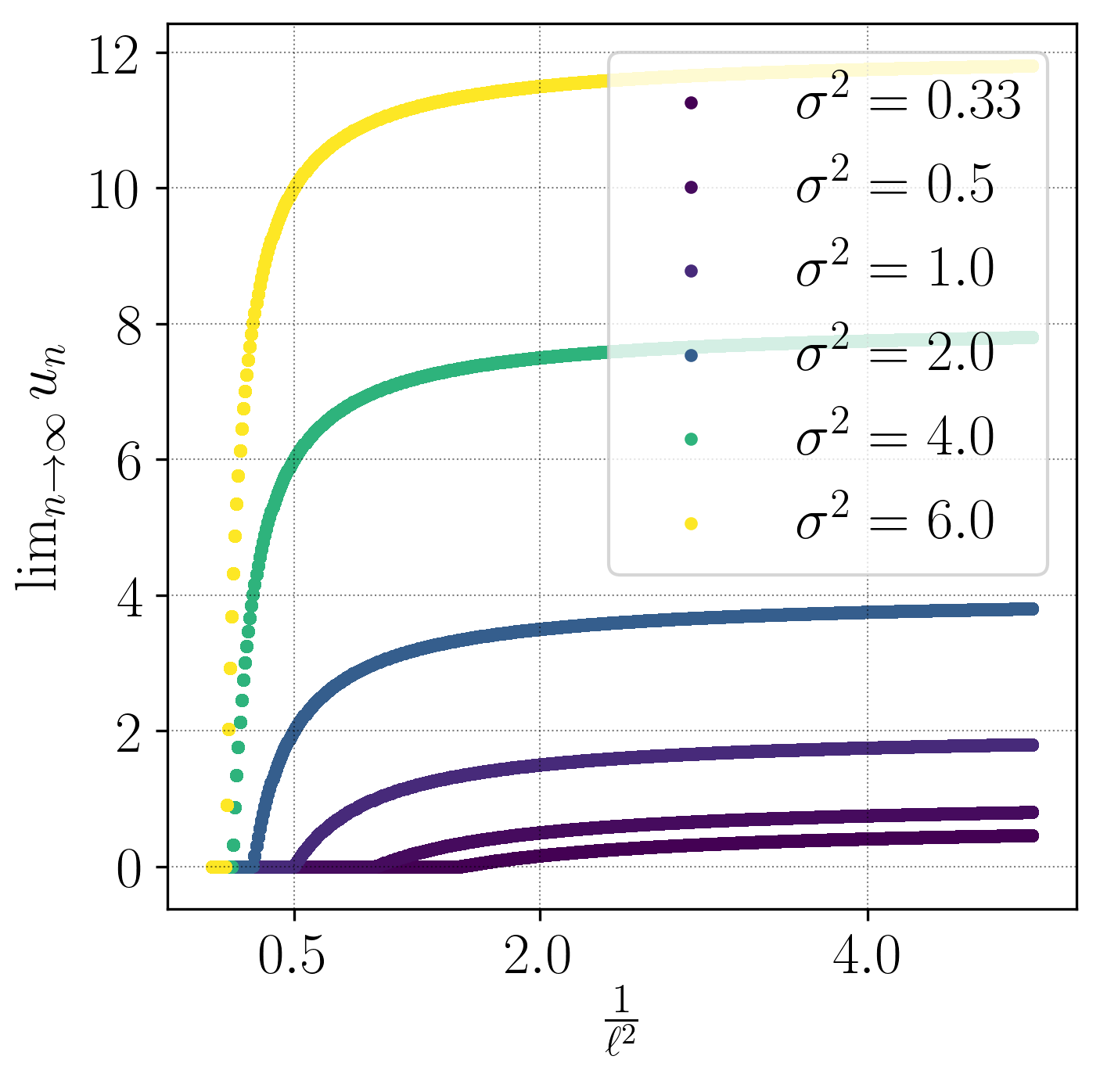}
    };
    \node[] at(0,-1.1) {{\tiny(a)}};
    
    \node[] at (2.,0) {
        \includegraphics[width=0.115\textwidth]{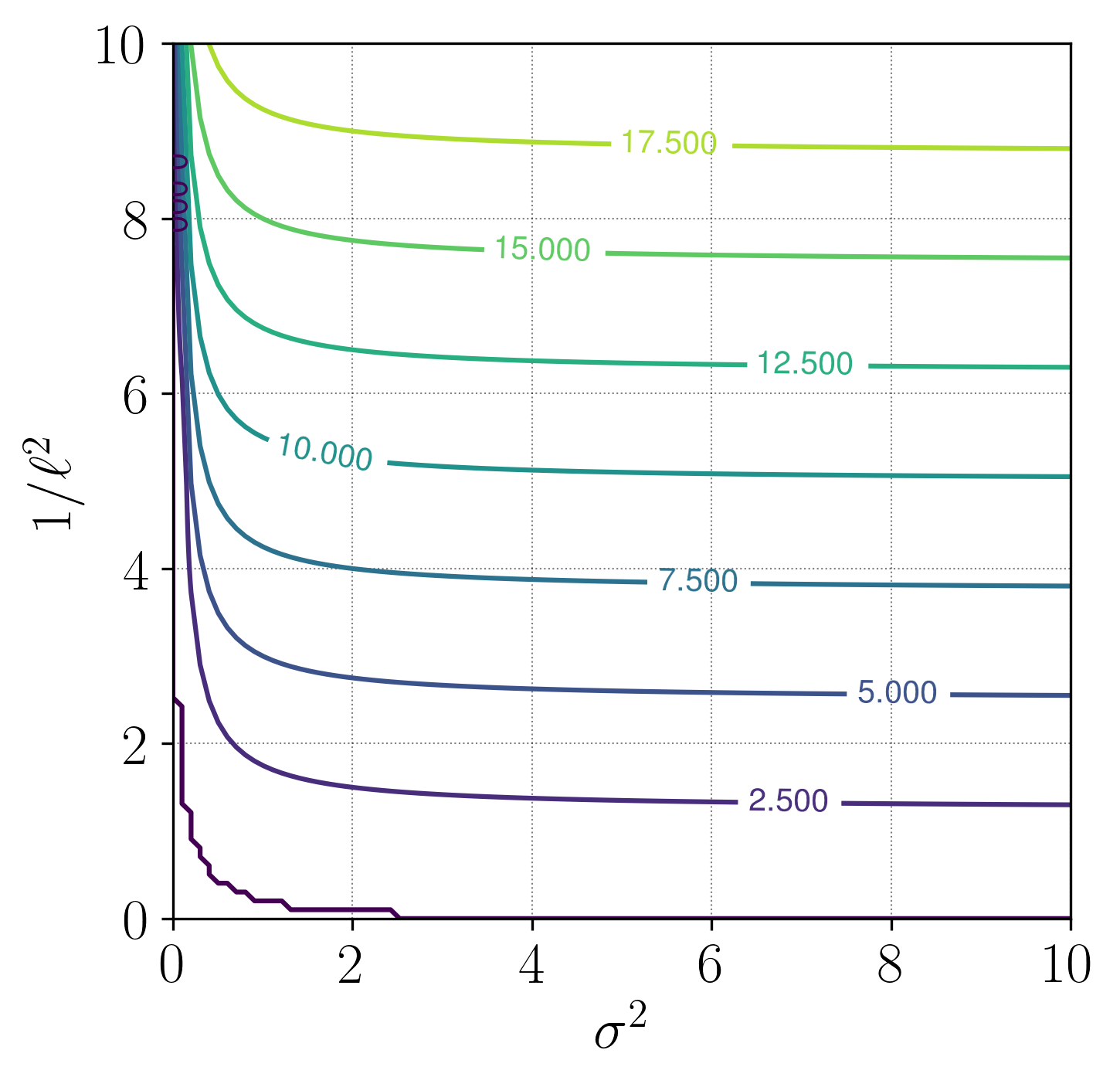}
    };
    \node[] at(2,-1.1) {\tiny(b)};
    
    \node[] at (4, 0) {
        \includegraphics[width=0.115\textwidth]{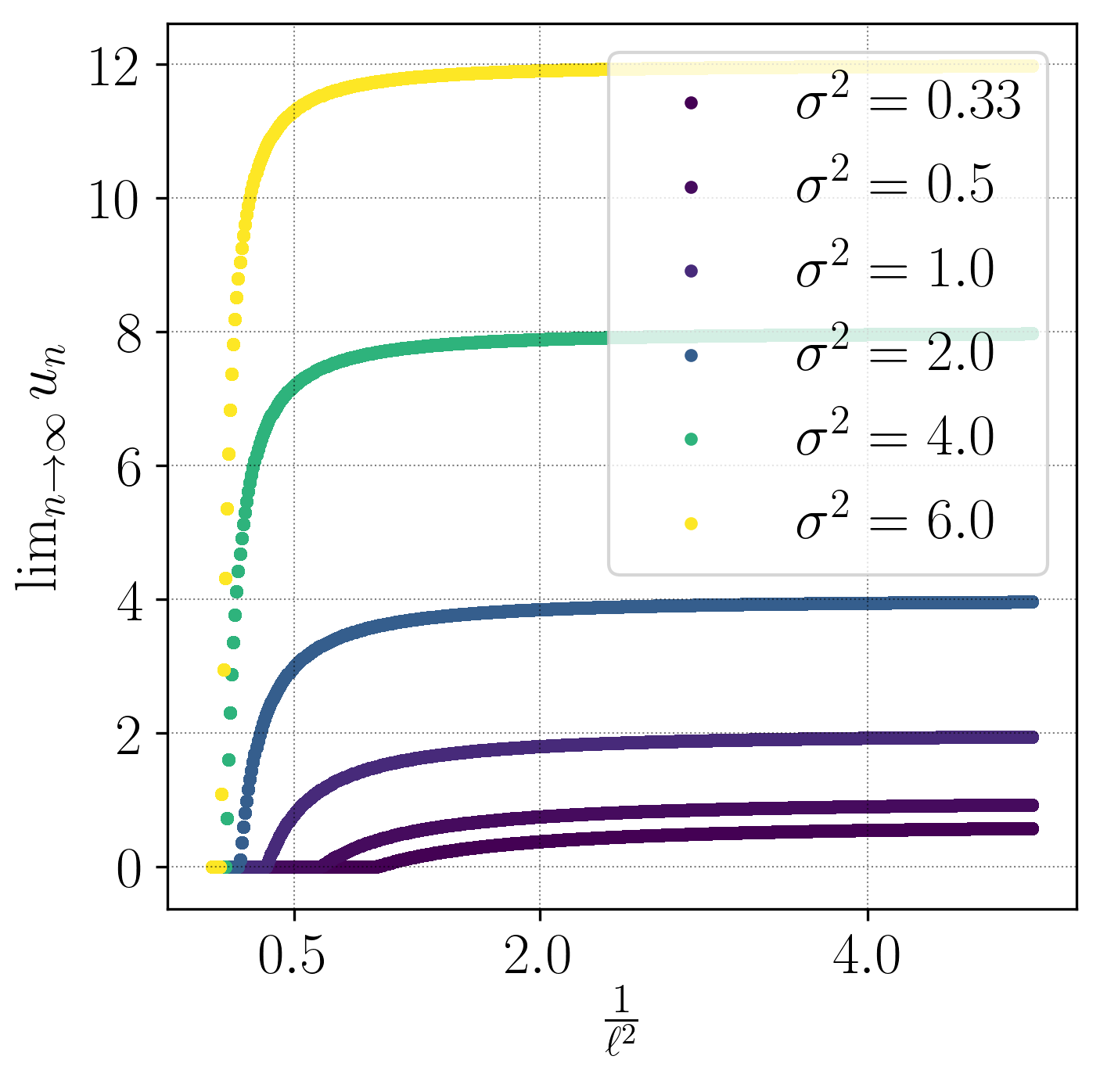}
    };
    \node[] at(4,-1.1) {\tiny(c)};
    
    \node[] at (6.,0) {
        \includegraphics[width=0.115\textwidth]{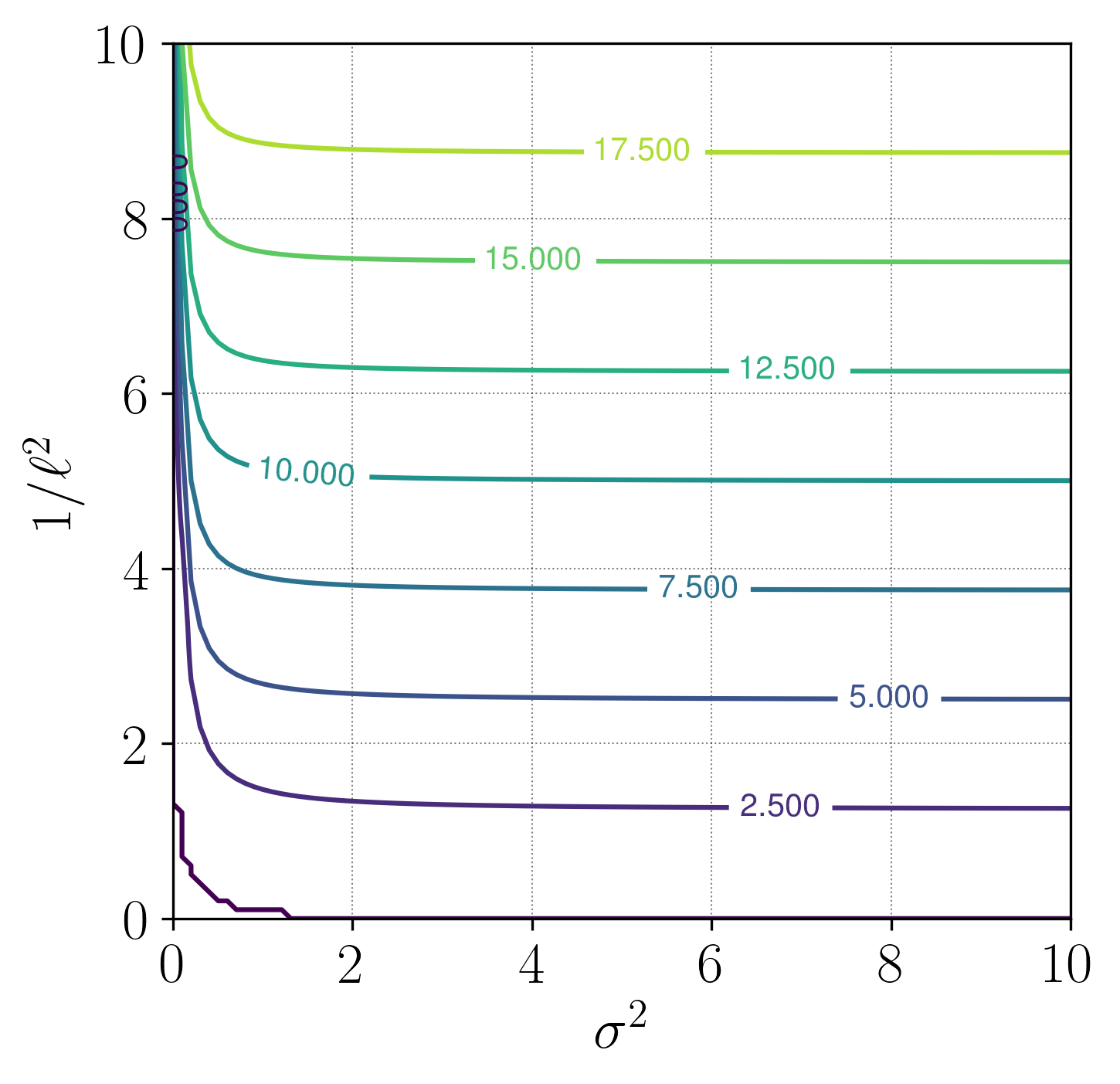}
    };
    \node[] at(6,-1.1) {\tiny(d)};
\end{tikzpicture}
}
    \caption{Bifurcation and contour plot of $\kSE$ kernel for two cases $m=2,3$. (a)-(b): $m=2$. (c)-(d): $m=3$.}
    \label{fig:se_bifurcation_m_2_3}
\end{figure}
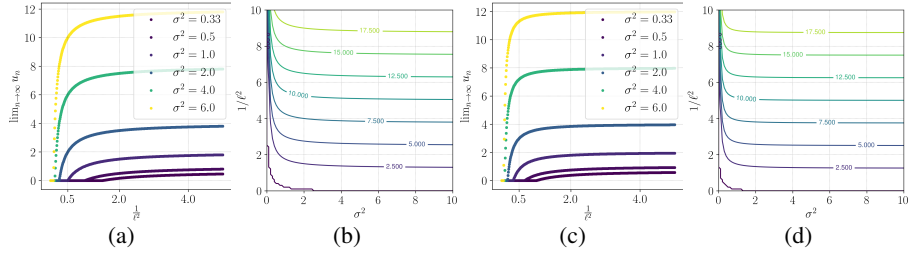{}

\section{More on experiment}
\label{appendix:experiment}
\subsection{High-dimensional~\dgps{} with $\kSE$}
\begin{figure}
    \centering
    \includegraphics[width=0.34\textwidth]{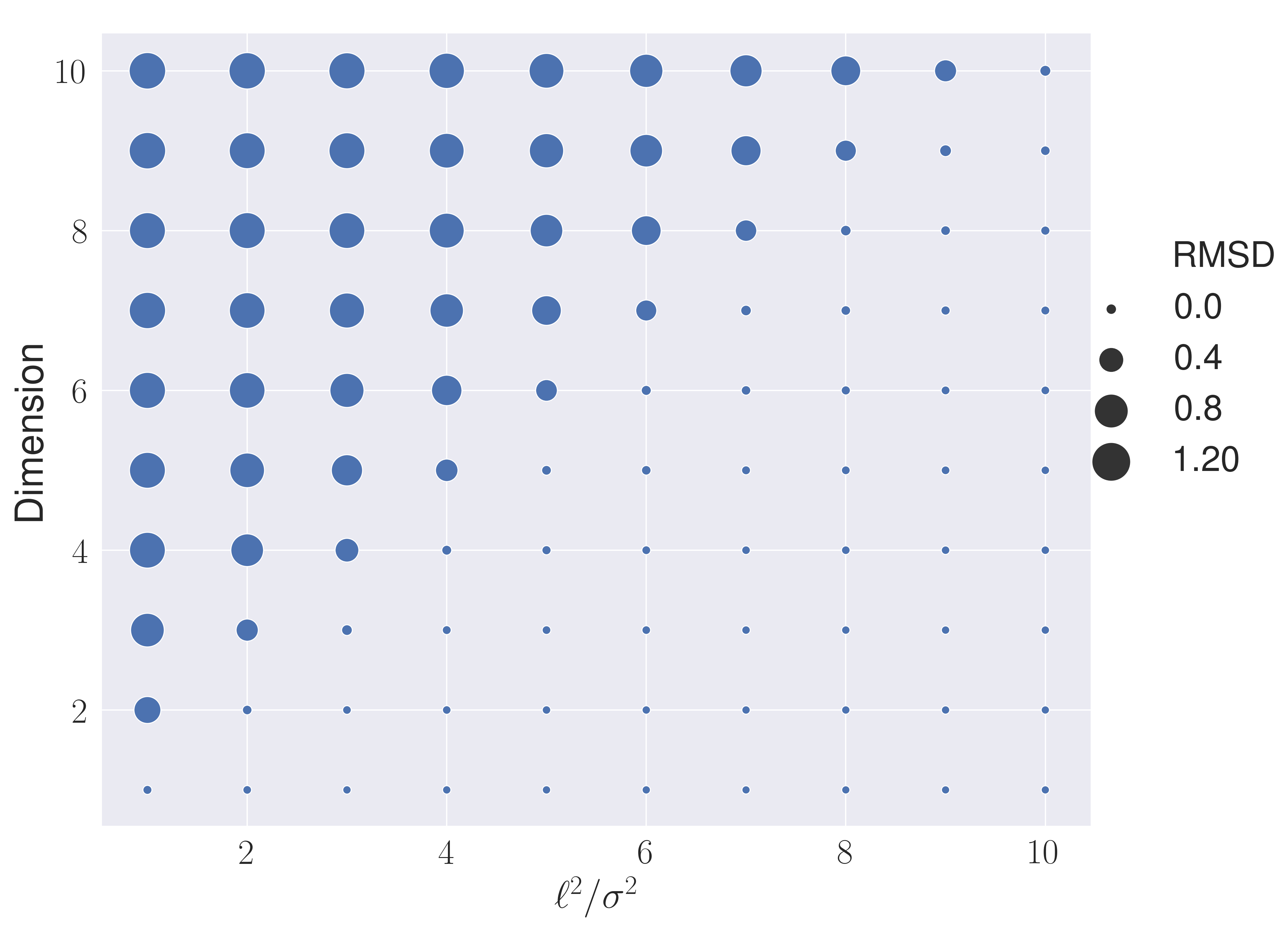}
    \caption{Plot of RMSDs at layer $100$ as the dimension $m$ and $\ell^2/\sigma^2$ change.}
    \label{fig:experiment_se_high_dim}
\end{figure}

In this experiment, we track the RMSDs at layer $100$ with different settings of dimension (number of units) at a layer, $m$, and the ratio $\ell^2/\sigma^2$. Figure~\ref{fig:experiment_se_high_dim} shows the values of RMSDs, illustrating that choosing $m > \ell^2/\sigma^2$ will help~{\dgp} overcome the pathology.

\subsection{Correctness of recurrence relations}
\begin{figure}
    \centering
    \begin{tikzpicture}
        \node[inner sep=0pt] at (0,0) {\includegraphics[width=0.25\textwidth]{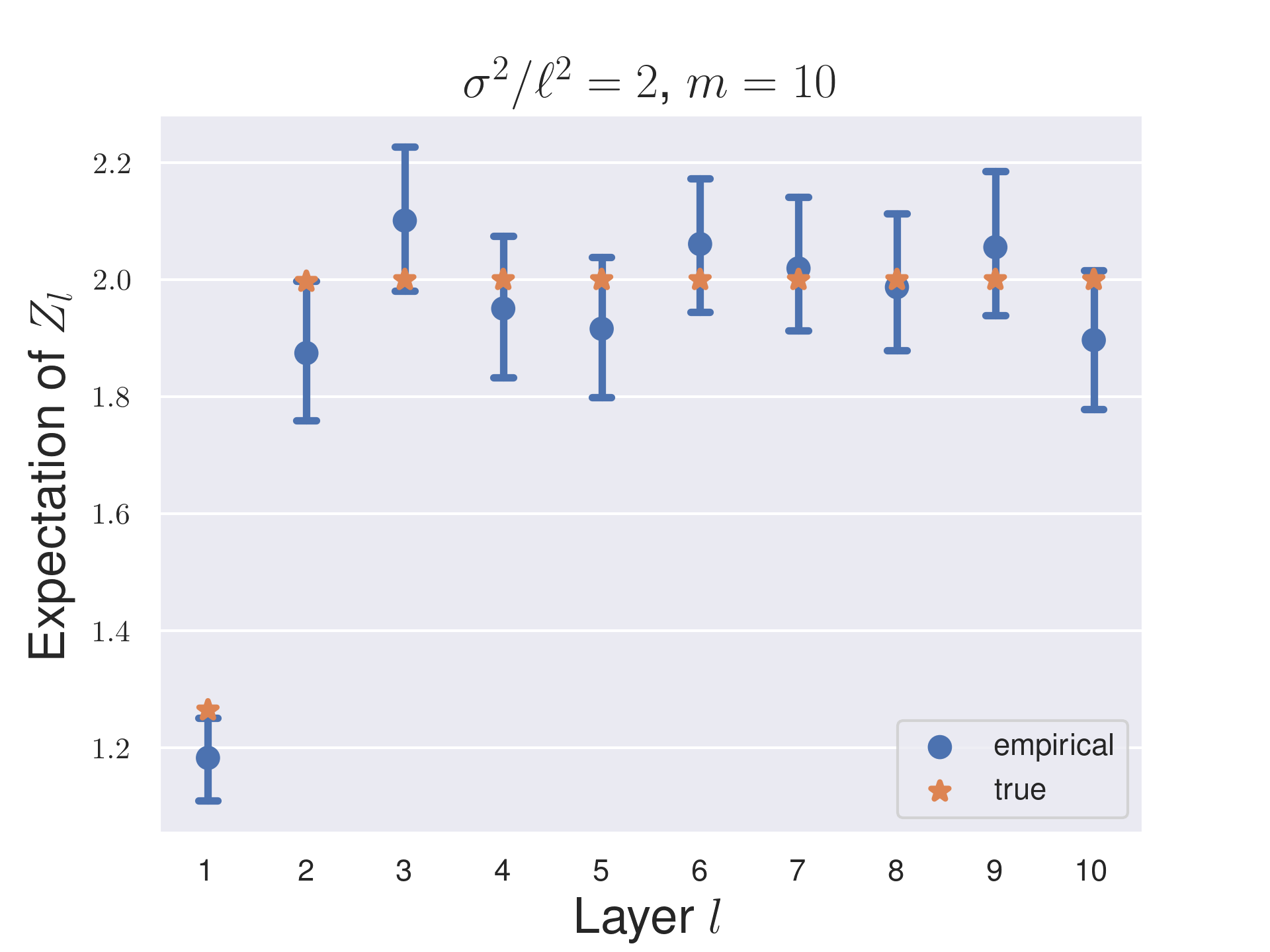}};
        
        \node[inner sep=0pt] at (4.5,0) {\includegraphics[width=0.25\textwidth]{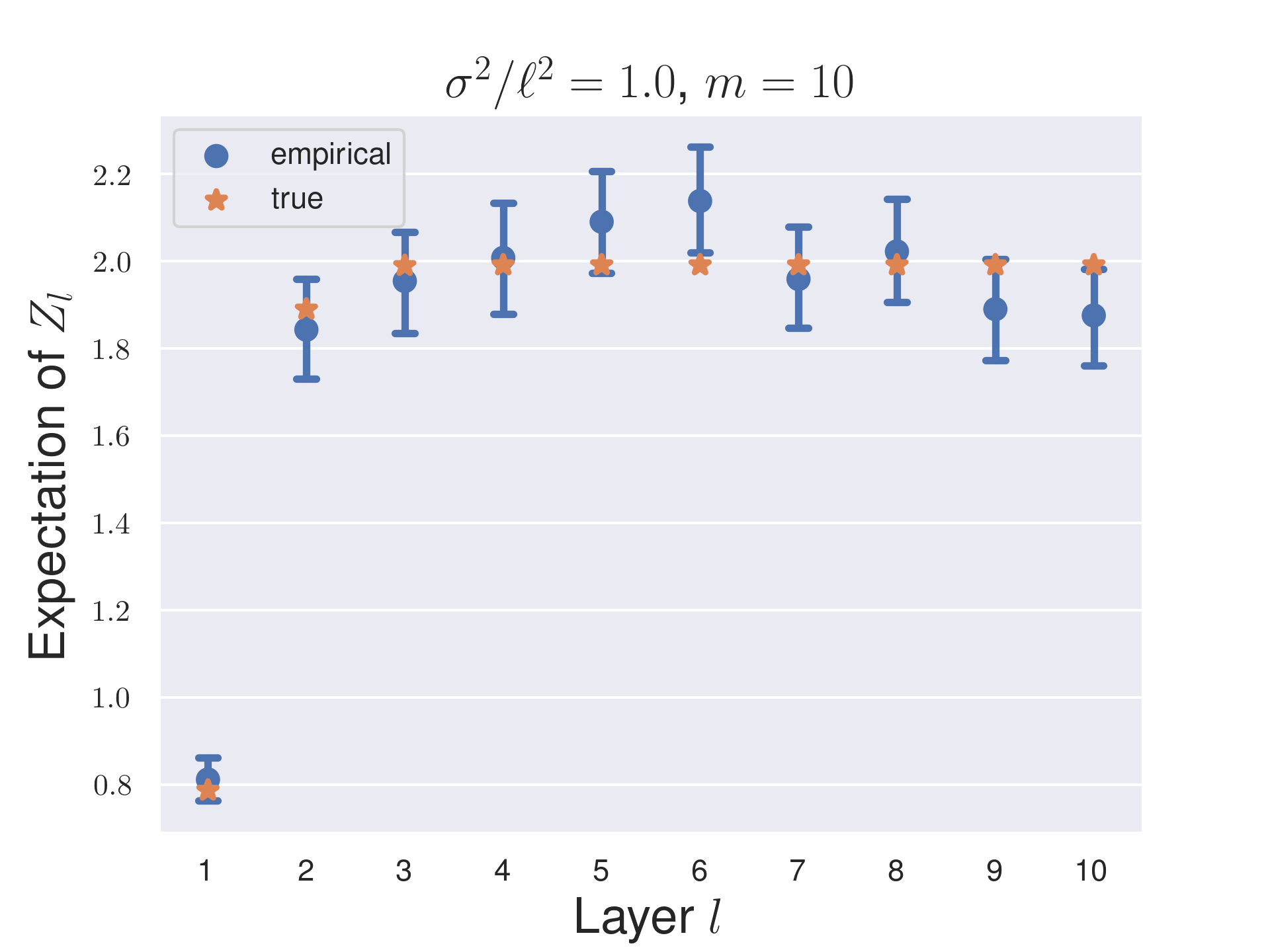}};
        
        \node[inner sep=0pt] at (9,0) {\includegraphics[width=0.25\textwidth]{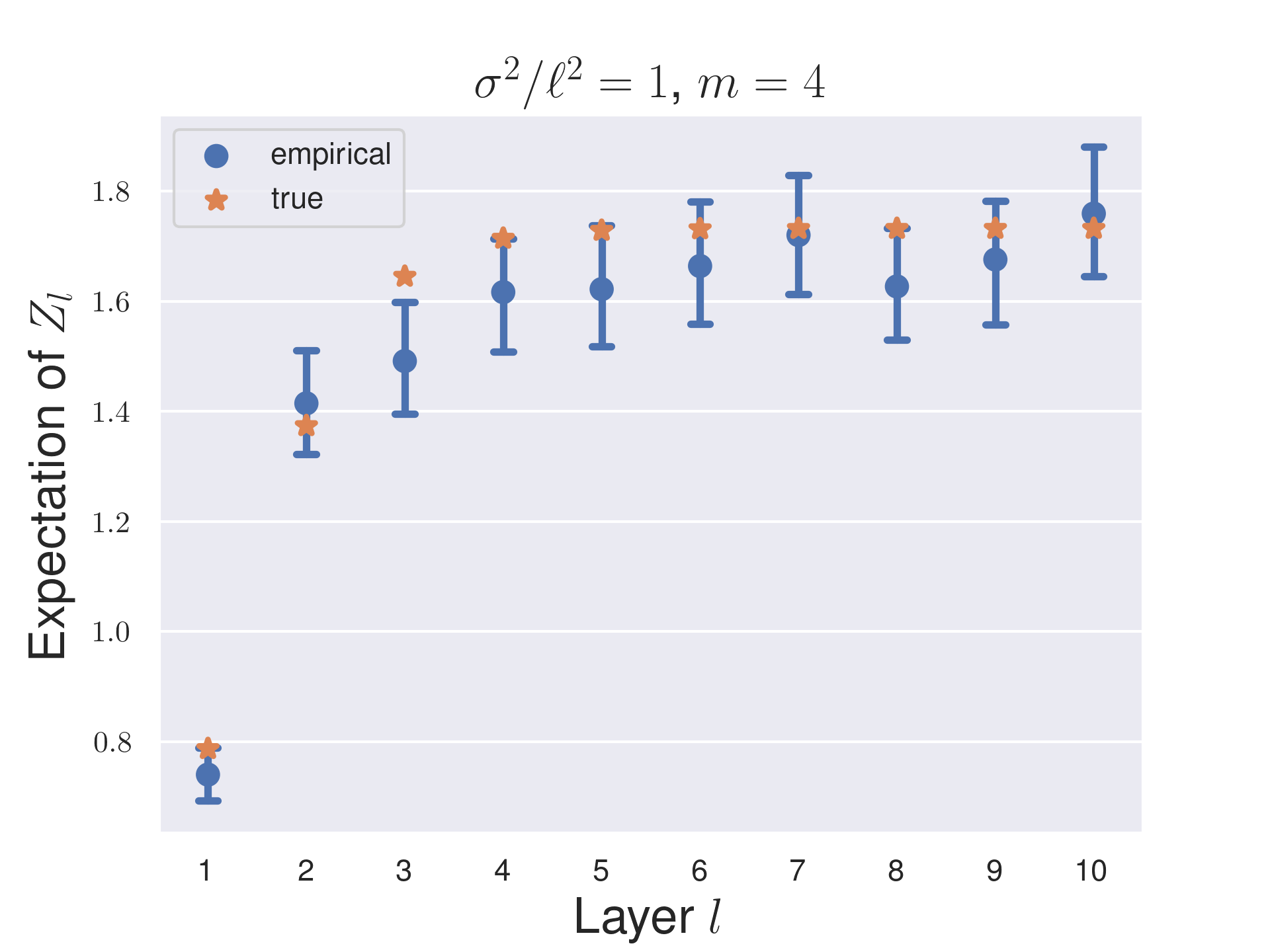}};
        
        \node[inner sep=0pt] at (13.5,0) {\includegraphics[width=0.25\textwidth]{figure/r_0_5_m_10.png}};
    \end{tikzpicture}
    \caption{High-dimensional $\kSE$: $\expect[Z_n]$ computed from recurrence vs. empirical estimation of $\expect[Z_n]$.}
    \label{fig:full_tracking_expectation_SE}
\end{figure}

\begin{figure}
    \centering
    \begin{tikzpicture}
        \node[inner sep=0pt] at (0,0) {\includegraphics[width=0.24\textwidth]{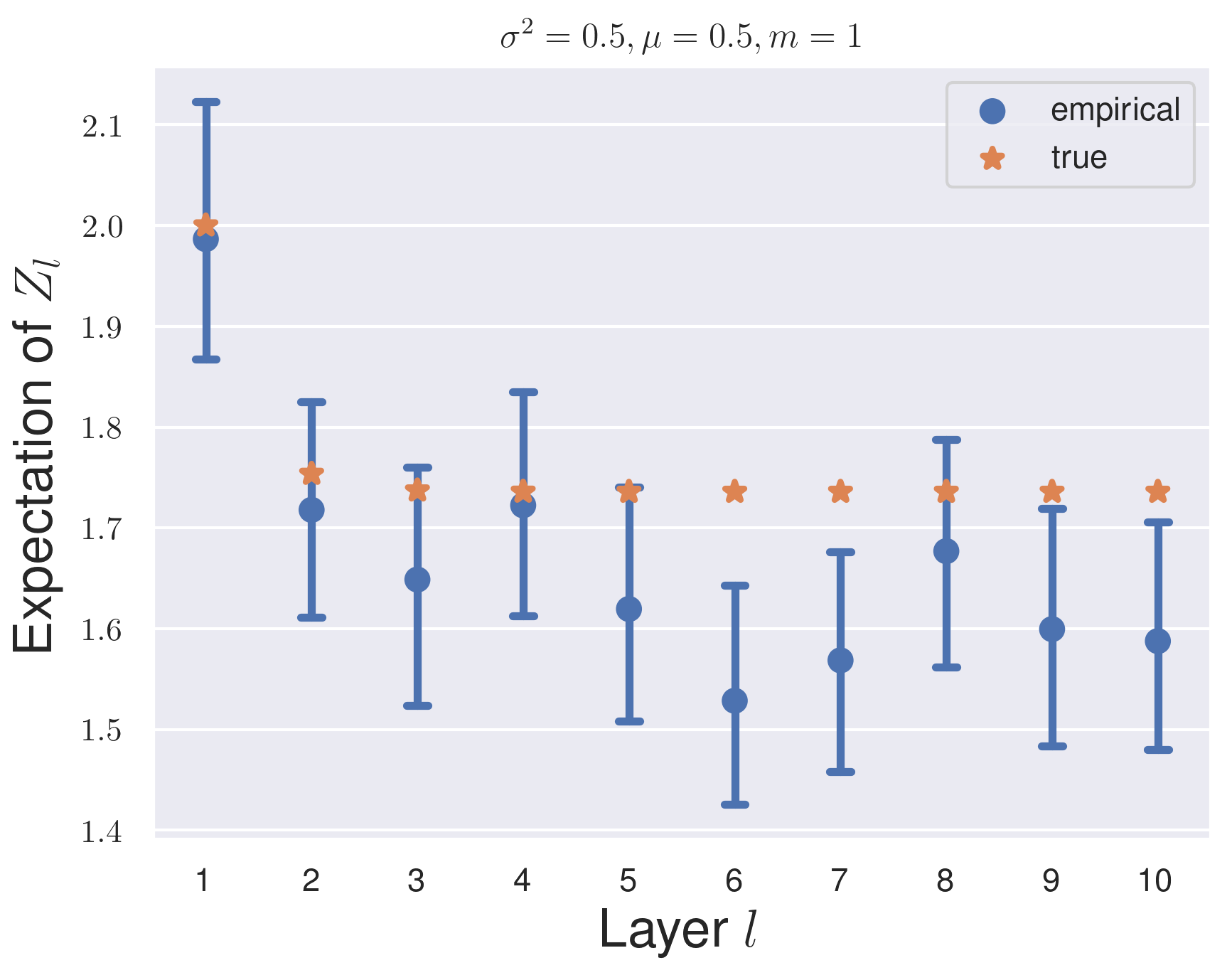}};
        
        \node[inner sep=0pt] at (4.5,0) {\includegraphics[width=0.24\textwidth]{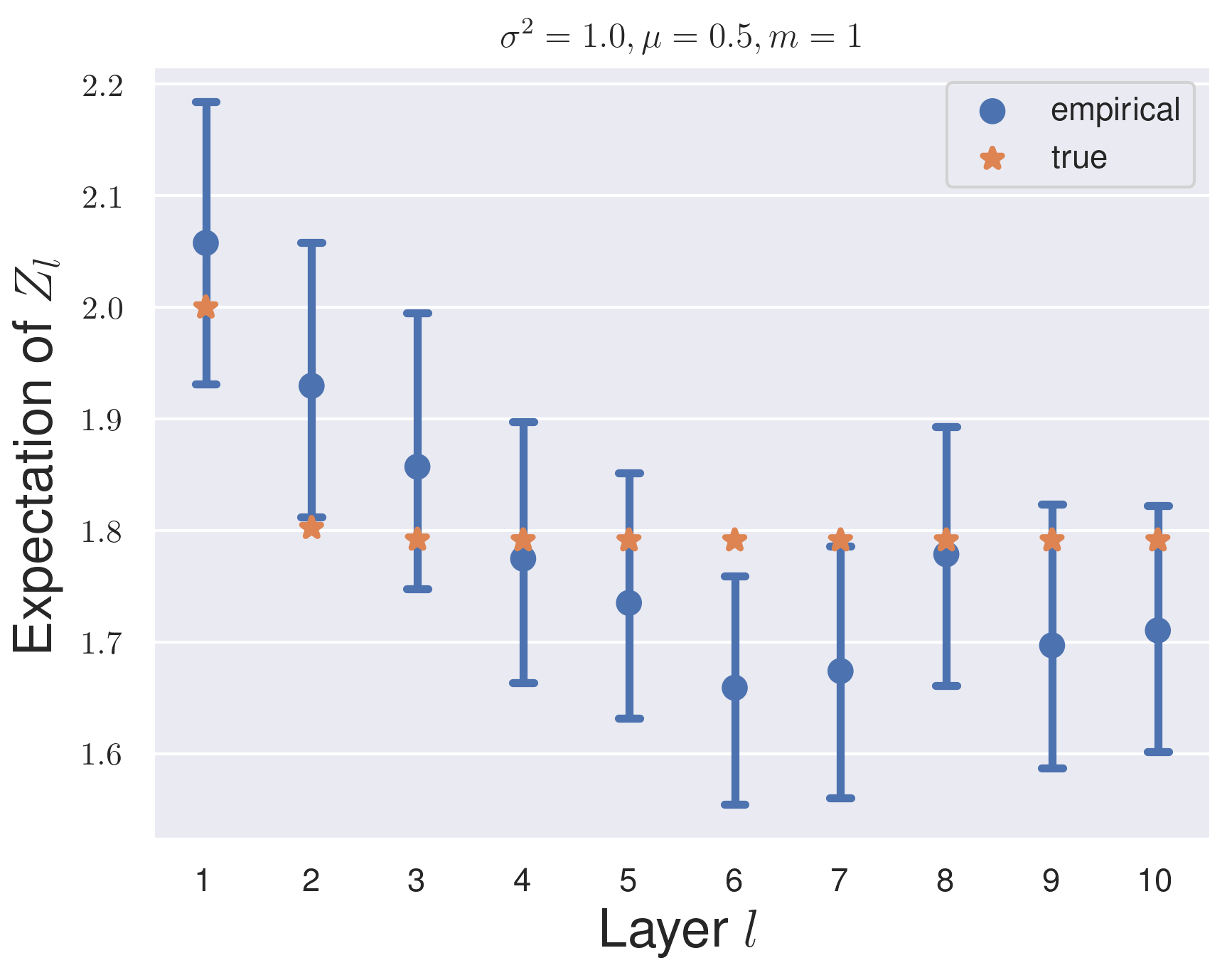}};
        
        \node[inner sep=0pt] at (9,0) {\includegraphics[width=0.24\textwidth]{figure/SM_1_0_1_0_m_1.png}};
        
        \node[inner sep=0pt] at (13.5,0) {\includegraphics[width=0.24\textwidth]{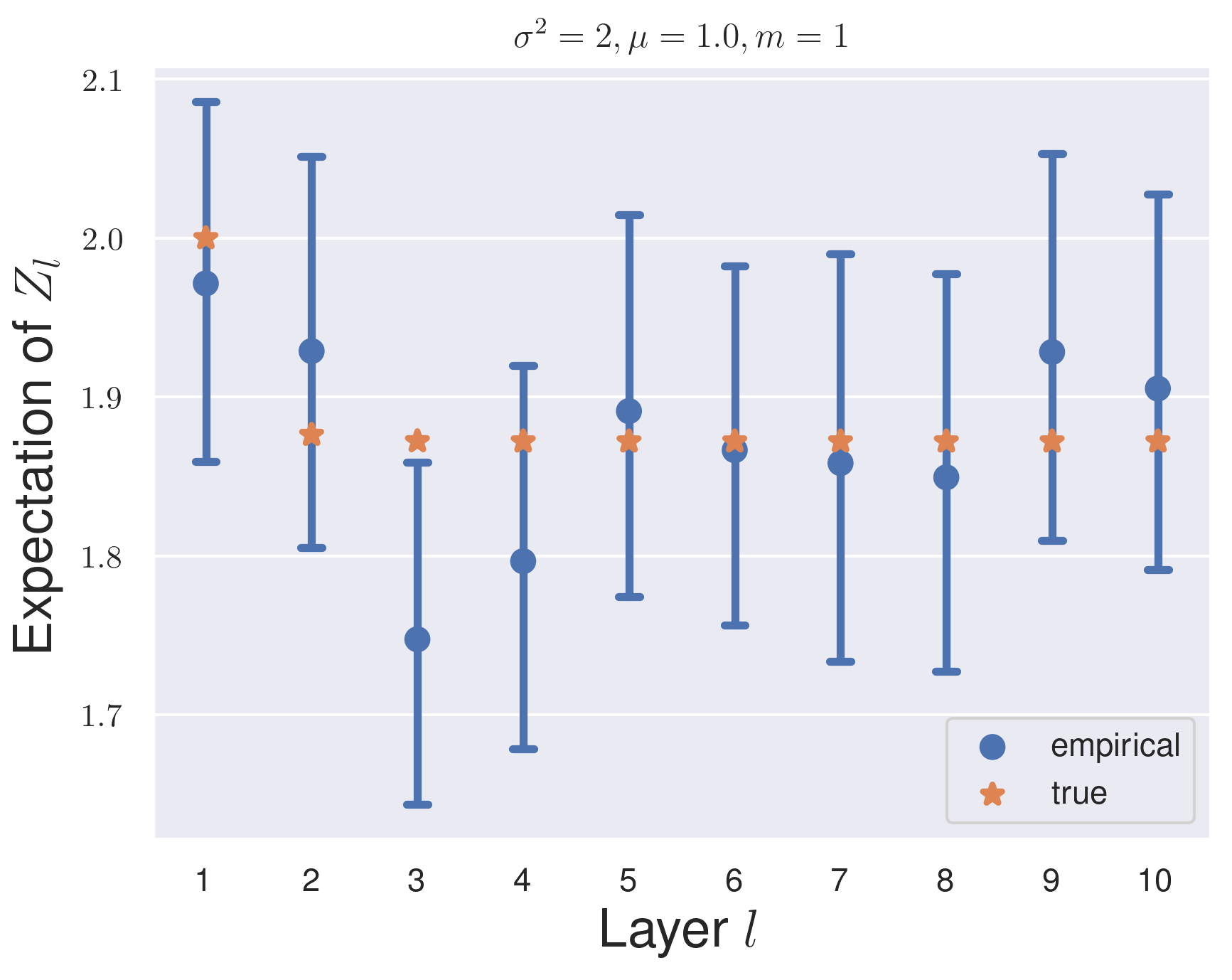}};
    \end{tikzpicture}
    \caption{$\textsc{SM}$ kernel: $\expect[Z_n]$ computed from recurrence vs. empirical estimation of $\expect[Z_n]$.}
    \label{fig:full_tracking_expectation_SM}
\end{figure}

Figure~\ref{fig:full_tracking_expectation_SE} and~\ref{fig:full_tracking_expectation_SM} contain the comparisons between the true $\expect[Z_n]$ versus its empirical estimation. 

\subsection{Monitoring~\dgp{} for Boston housing and diabetes data set}
The Boston housing data set concerns housing values in suburbs of Boston, consisting of 506 instances with 14 attributes. The diabetes data set~\cite{diabete_dataset} measuring disease progression contains 442 instances with 10 attributes.

\begin{figure}
    \centering
    \begin{tikzpicture}
        
        \node[inner sep=0pt] at (-4,0) {\includegraphics[width=0.22\textwidth]{figure/boston_nosharing/m_2_dgp.png}};
        
        \node[inner sep=0pt] at (0,0) {\includegraphics[width=0.22\textwidth]{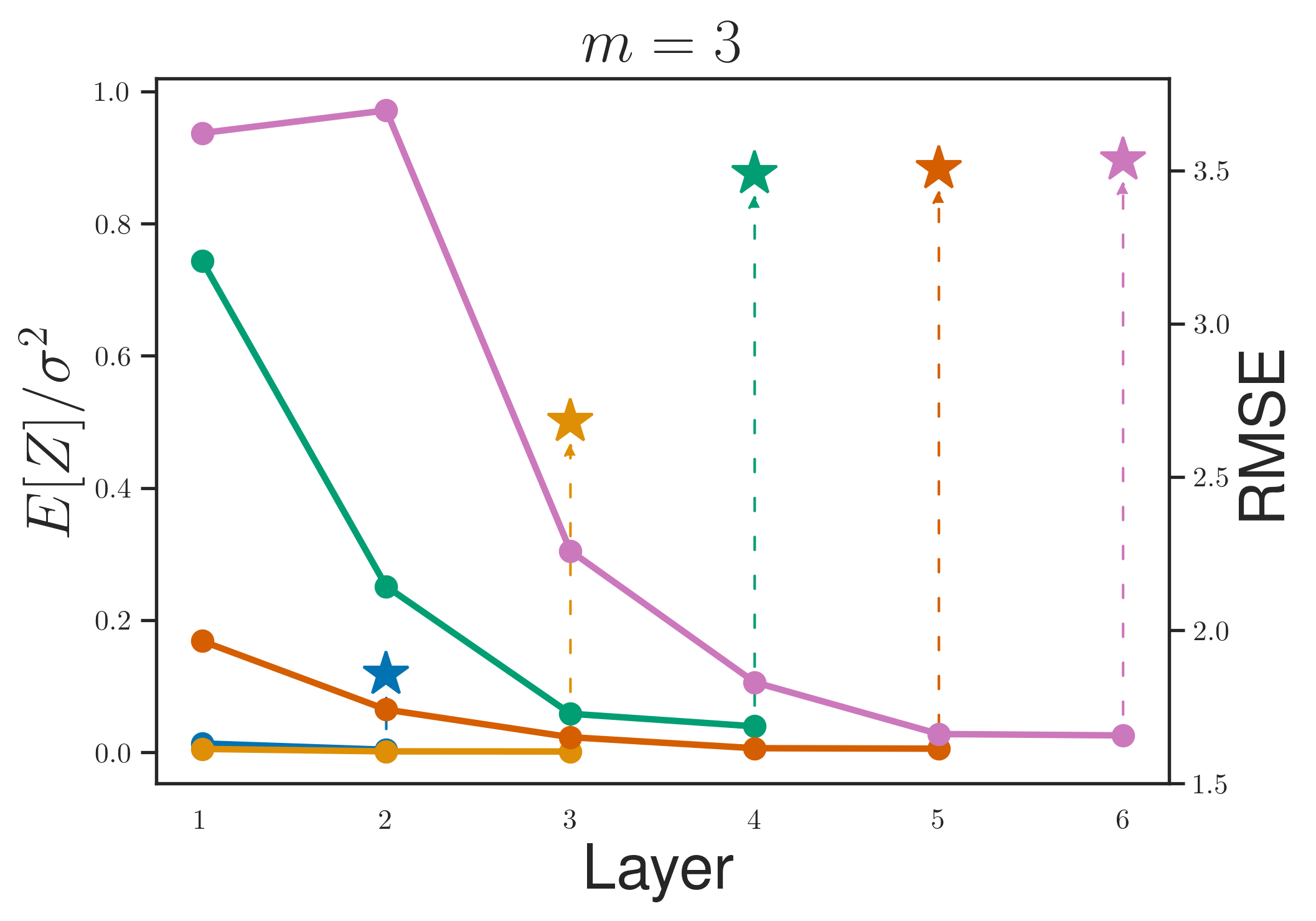}};
        
        \node[inner sep=0pt] at (4,0) {\includegraphics[width=0.22\textwidth]{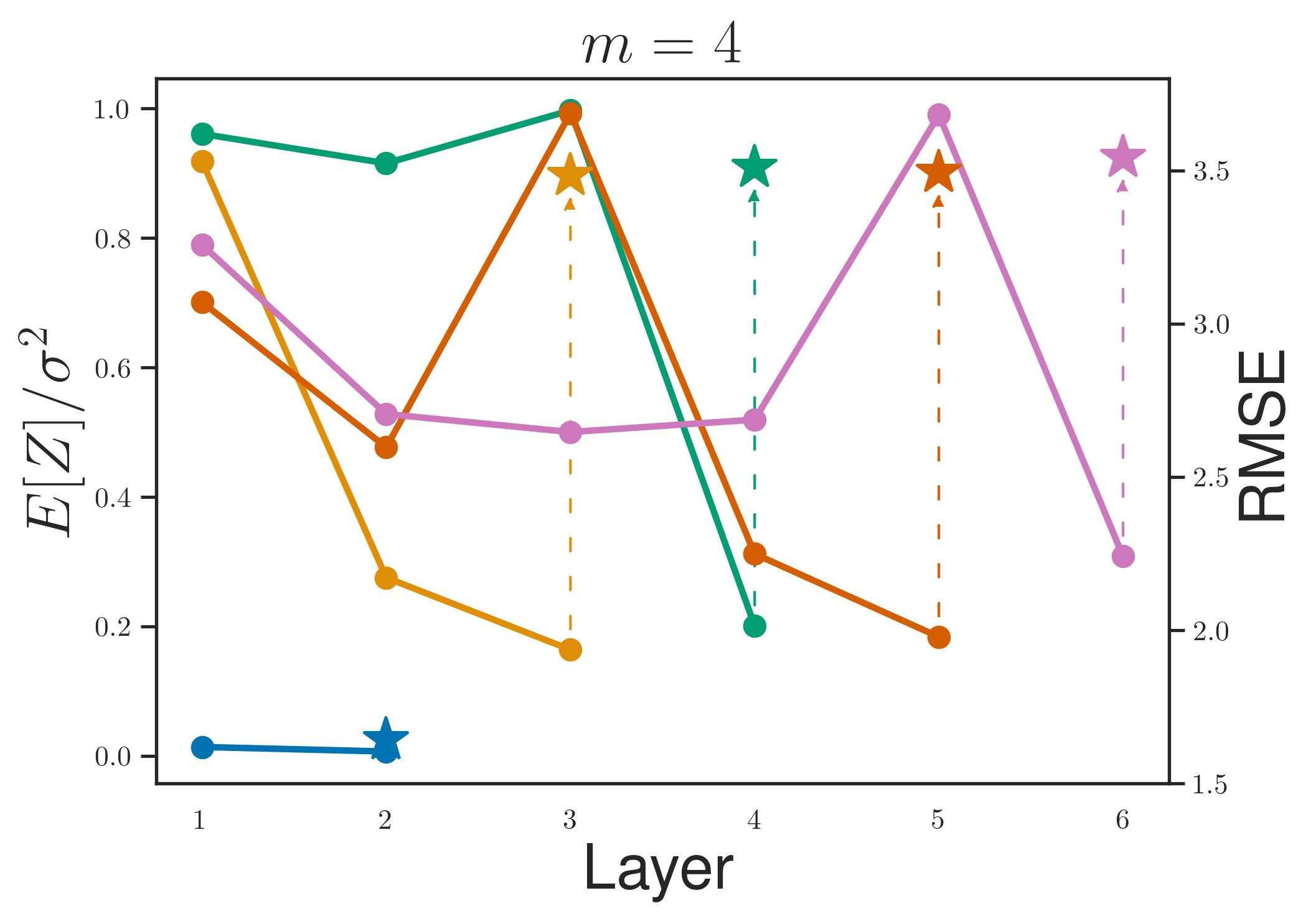}};
        
        \node[inner sep=0pt] at (8,0) {\includegraphics[width=0.22\textwidth]{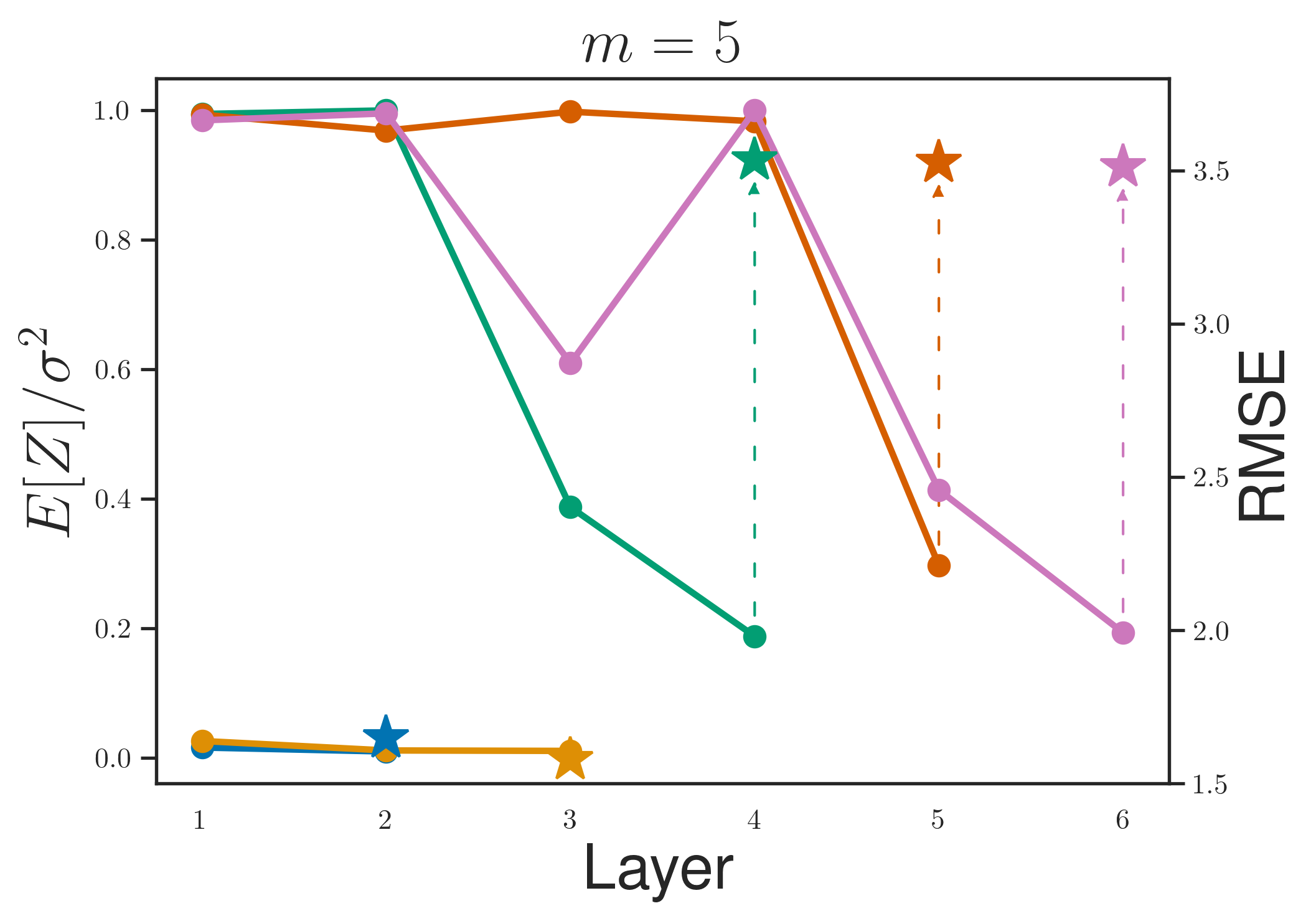}};
        
        \node[inner sep=0pt] at (-4,-2.8) {\includegraphics[width=0.22\textwidth]{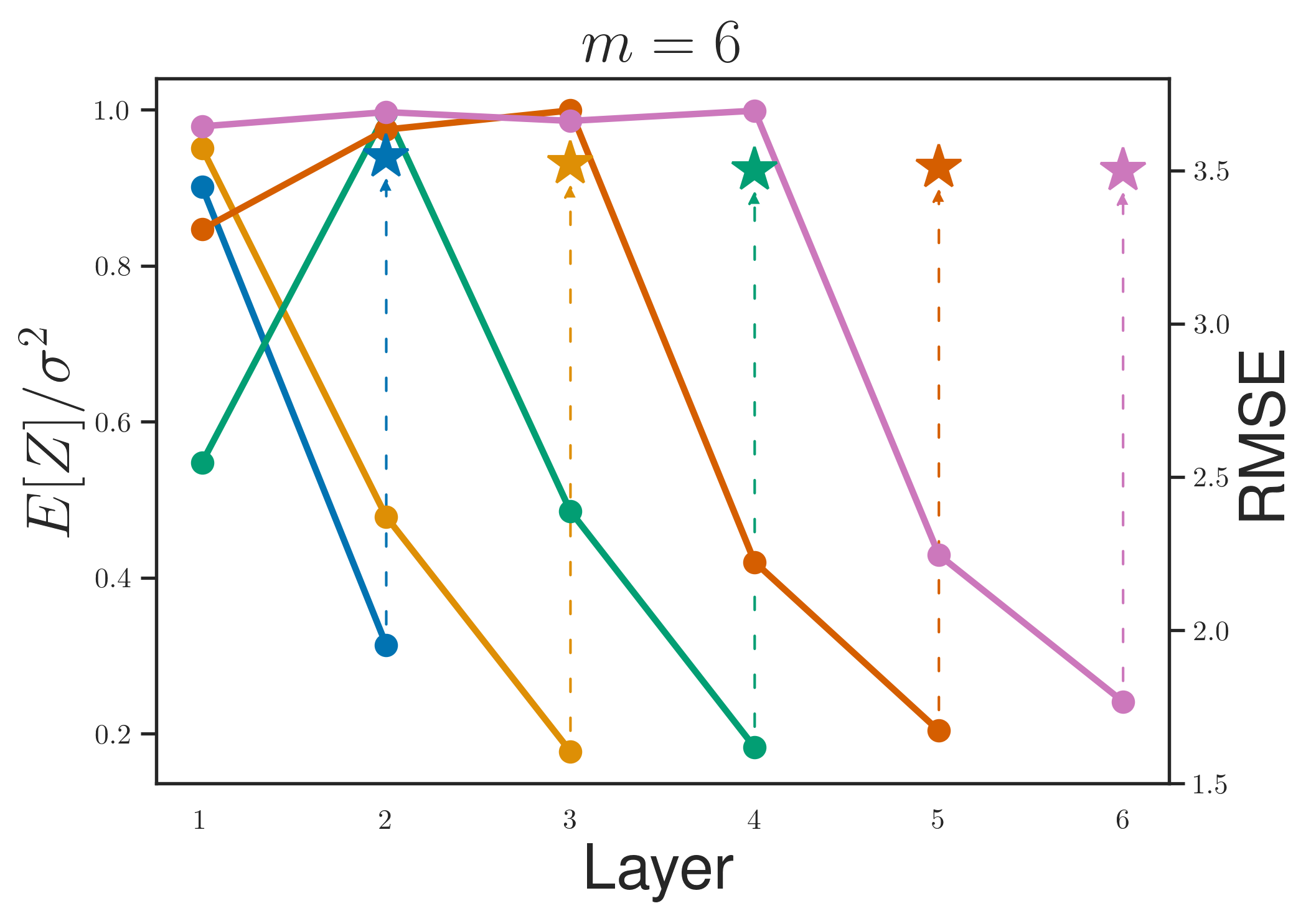}};
        
        \node[inner sep=0pt] at (0,-2.8) {\includegraphics[width=0.22\textwidth]{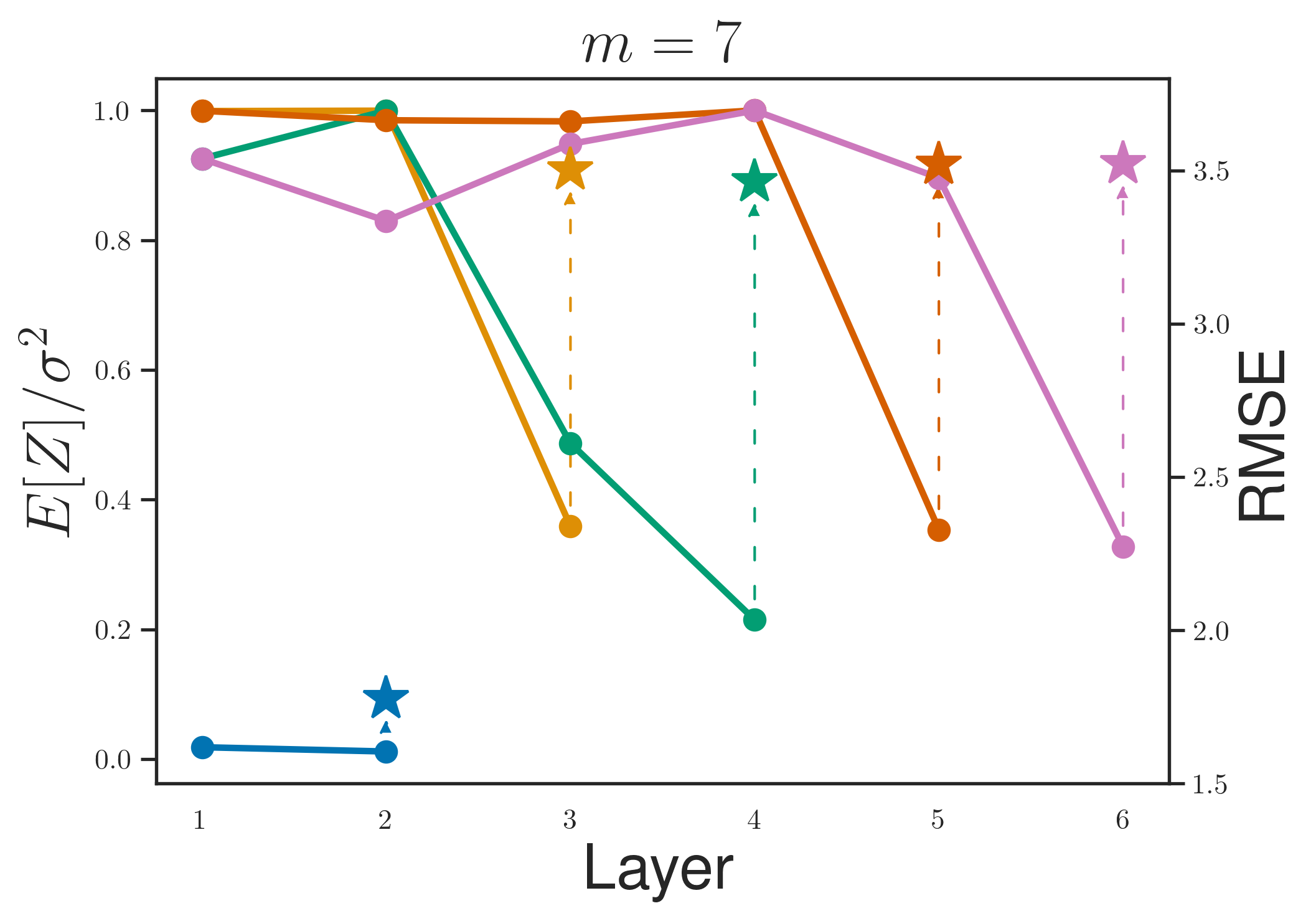}};
        
        \node[inner sep=0pt] at (4,-2.8) {\includegraphics[width=0.22\textwidth]{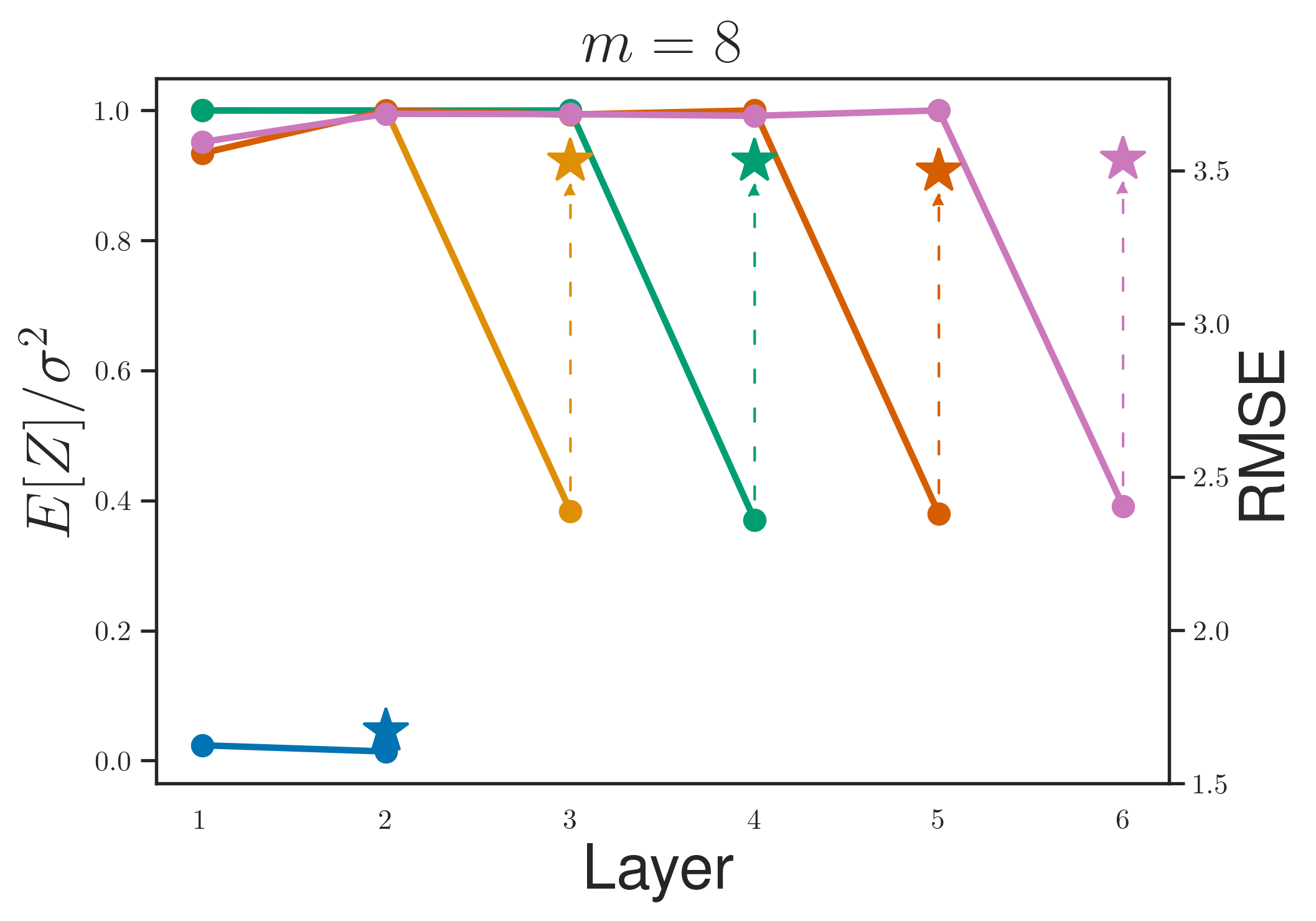}};
        
        \node[inner sep=0pt] at (8,-2.8) {\includegraphics[width=0.22\textwidth]{figure/boston_nosharing/m_9_dgp.png}};

    \end{tikzpicture}
    \caption{\textbf{Standard zero-mean DGPs}: Results of Boston housing data set}
    \label{fig:plot_boston_no_sharing}
\end{figure}

Figure~\ref{fig:plot_boston_no_sharing} contains the results from the Boston housing data set with \textbf{standard} zero-mean~\dgps{}.

\begin{figure}[t]
    \centering
    \begin{tikzpicture}
        
        \node[inner sep=0pt] at (-4.,0) {\includegraphics[width=0.22\textwidth]{figure/boston_nosharing/m_2_dgp_01.png}};
        
        \node[inner sep=0pt] at (0,0) {\includegraphics[width=0.22\textwidth]{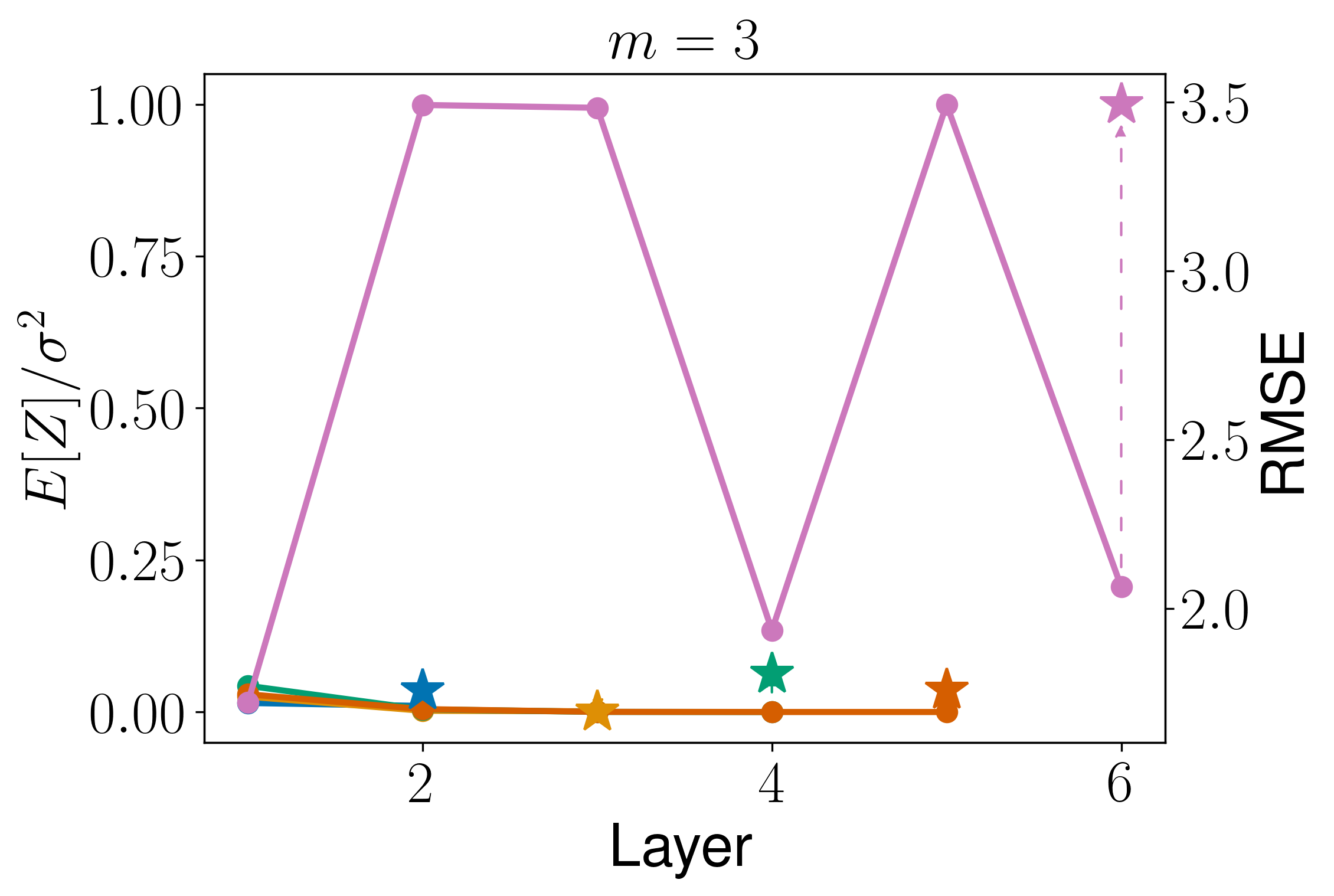}};
        
        \node[inner sep=0pt] at (4,0) {\includegraphics[width=0.22\textwidth]{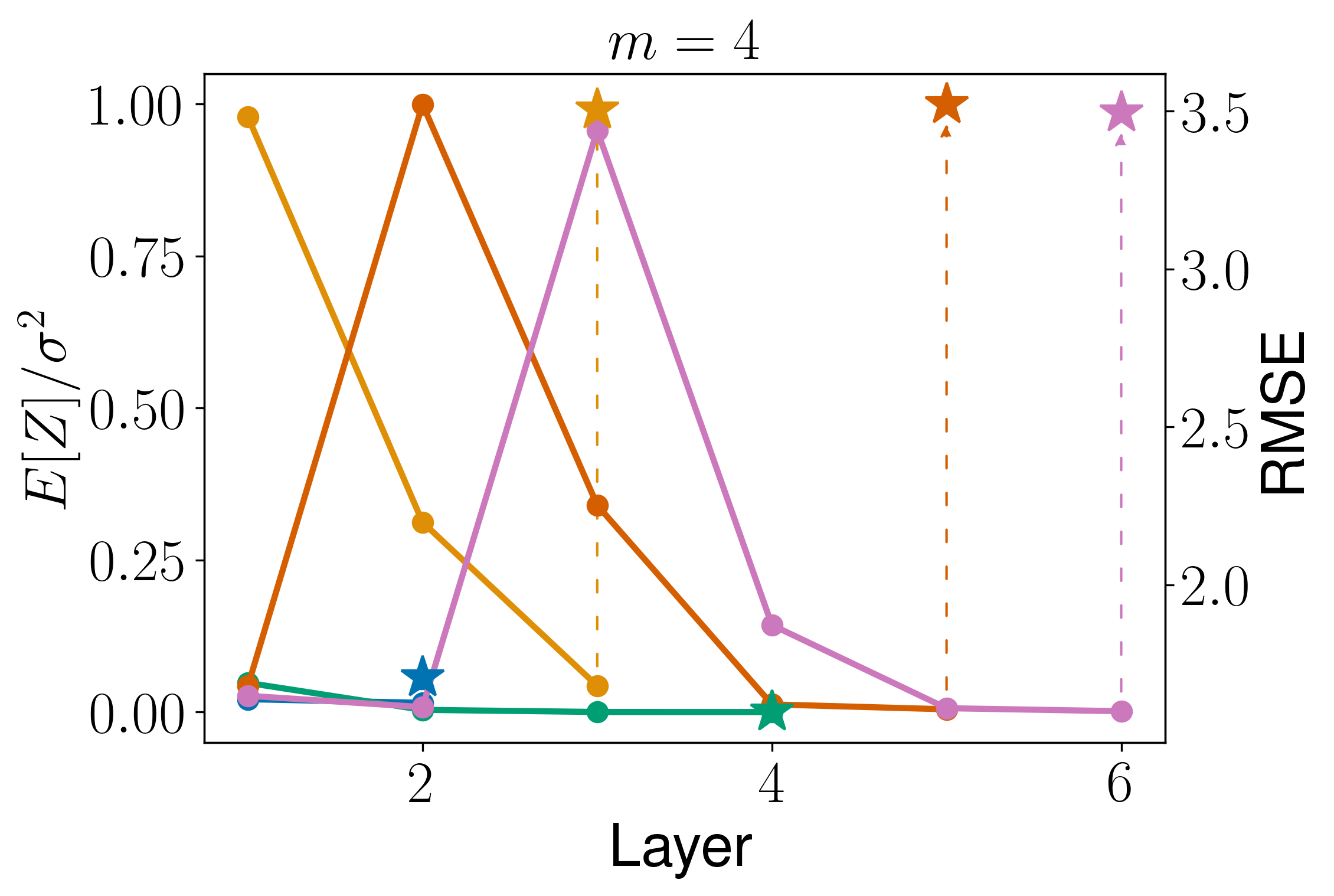}};
        
        \node[inner sep=0pt] at (8,0) {\includegraphics[width=0.22\textwidth]{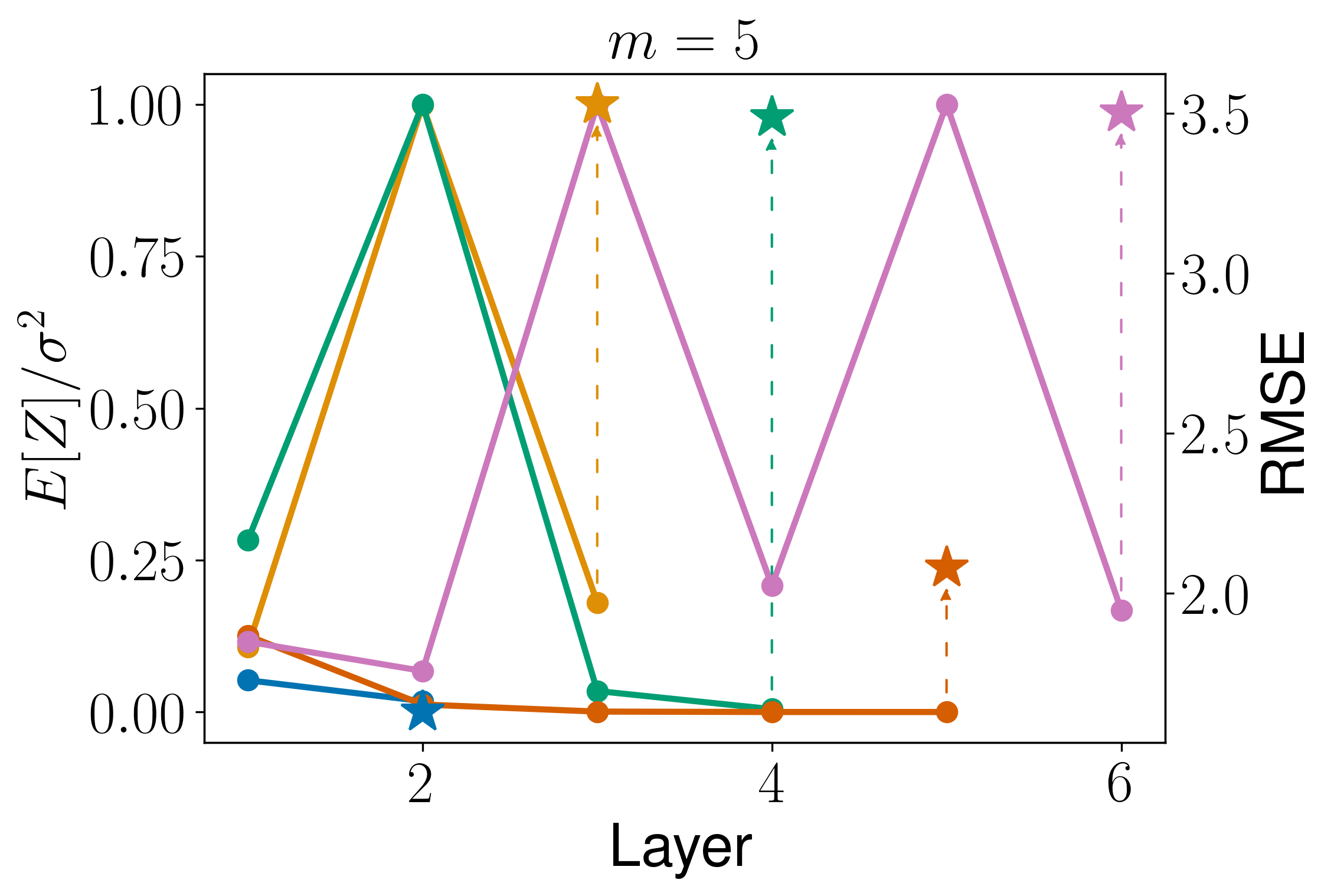}};
        
        \node[inner sep=0pt] at (-4,-2.8) {\includegraphics[width=0.22\textwidth]{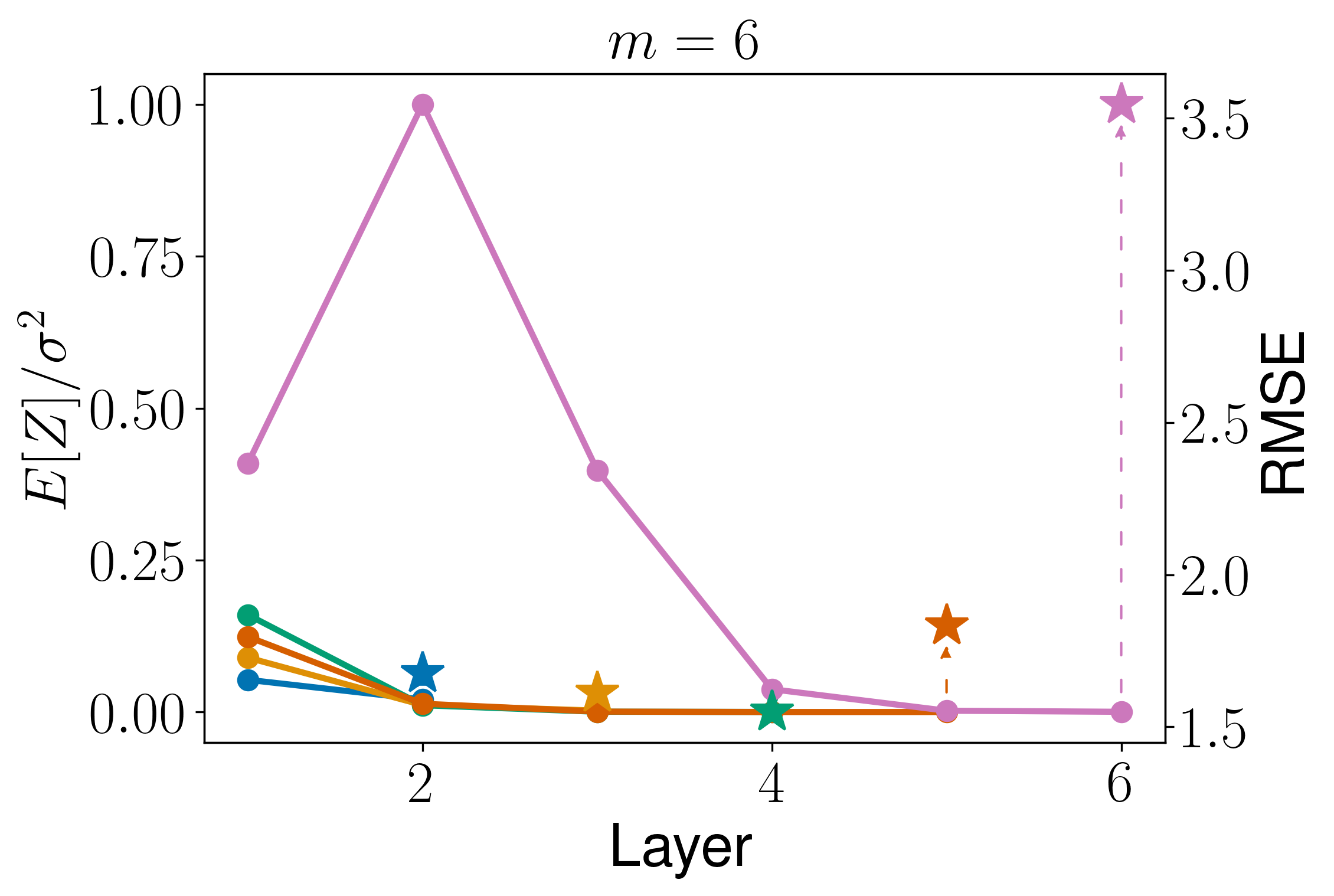}};
        
        \node[inner sep=0pt] at (0,-2.8) {\includegraphics[width=0.22\textwidth]{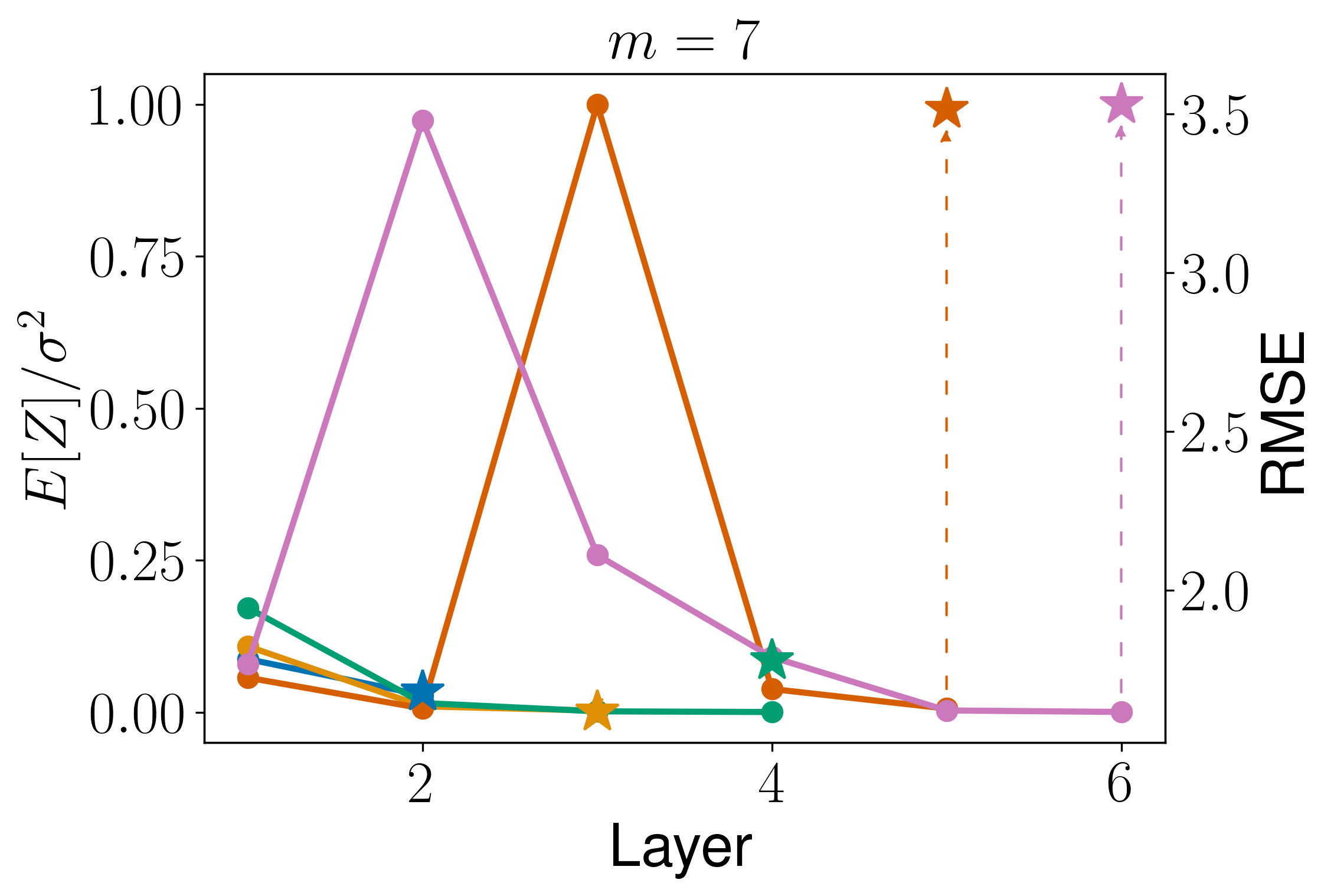}};
        
        \node[inner sep=0pt] at (4,-2.8) {\includegraphics[width=0.22\textwidth]{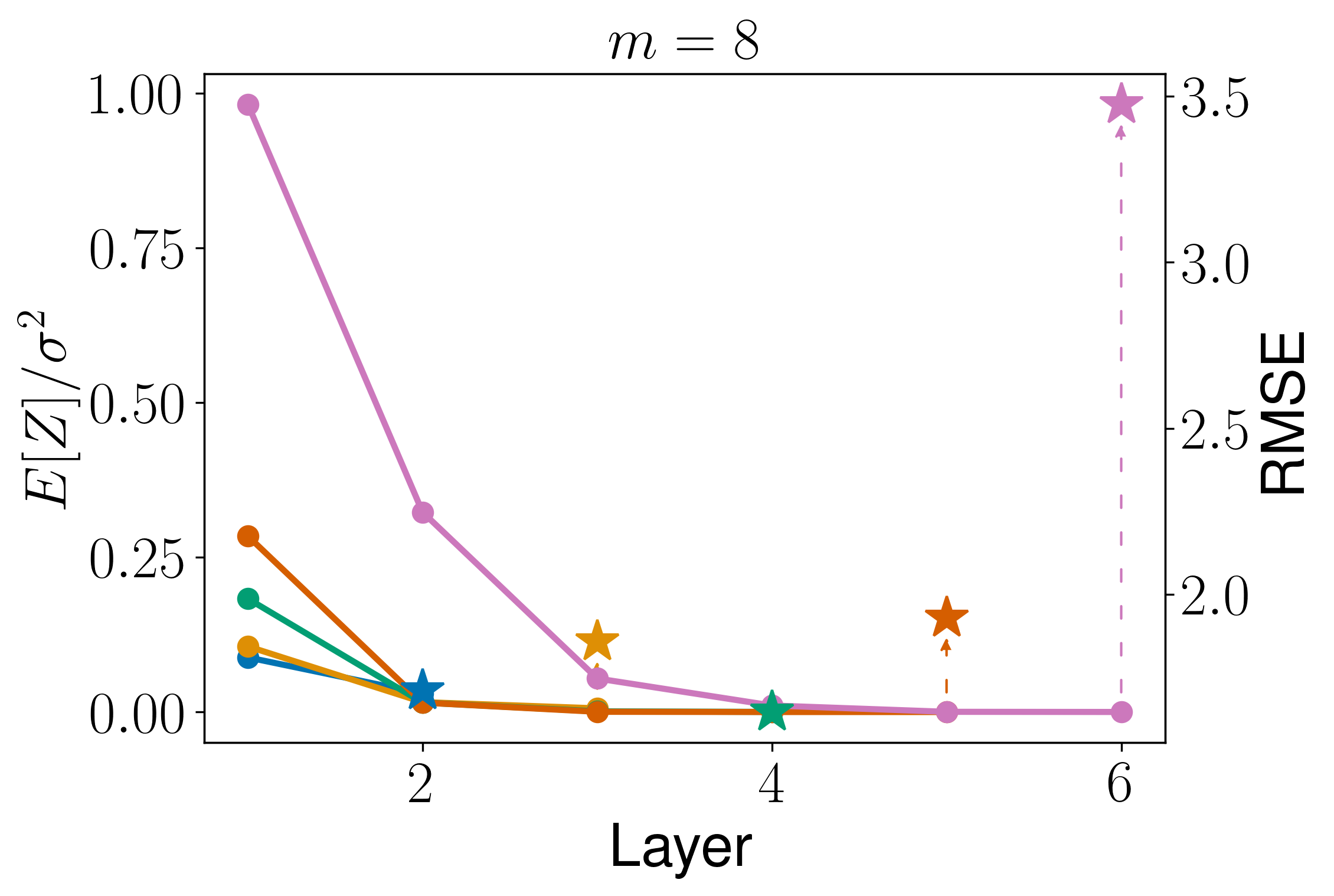}};
        
        \node[inner sep=0pt] at (8,-2.8) {\includegraphics[width=0.22\textwidth]{figure/boston_nosharing/m_9_dgp_03.png}};

    \end{tikzpicture}
    \caption{\textbf{Constrained~\dgps{}}: Results of Boston housing data set}
    \label{fig:plot_boston}
\end{figure}

Figure~\ref{fig:plot_boston} contains the results from the Boston housing data set with \textbf{constrained} zero-mean~\dgps{} where $c_0 = 0.2$.

\begin{figure}
    \centering
    \begin{tikzpicture}
        
        \node[inner sep=0pt] at (-4.,0) {\includegraphics[width=0.22\textwidth]{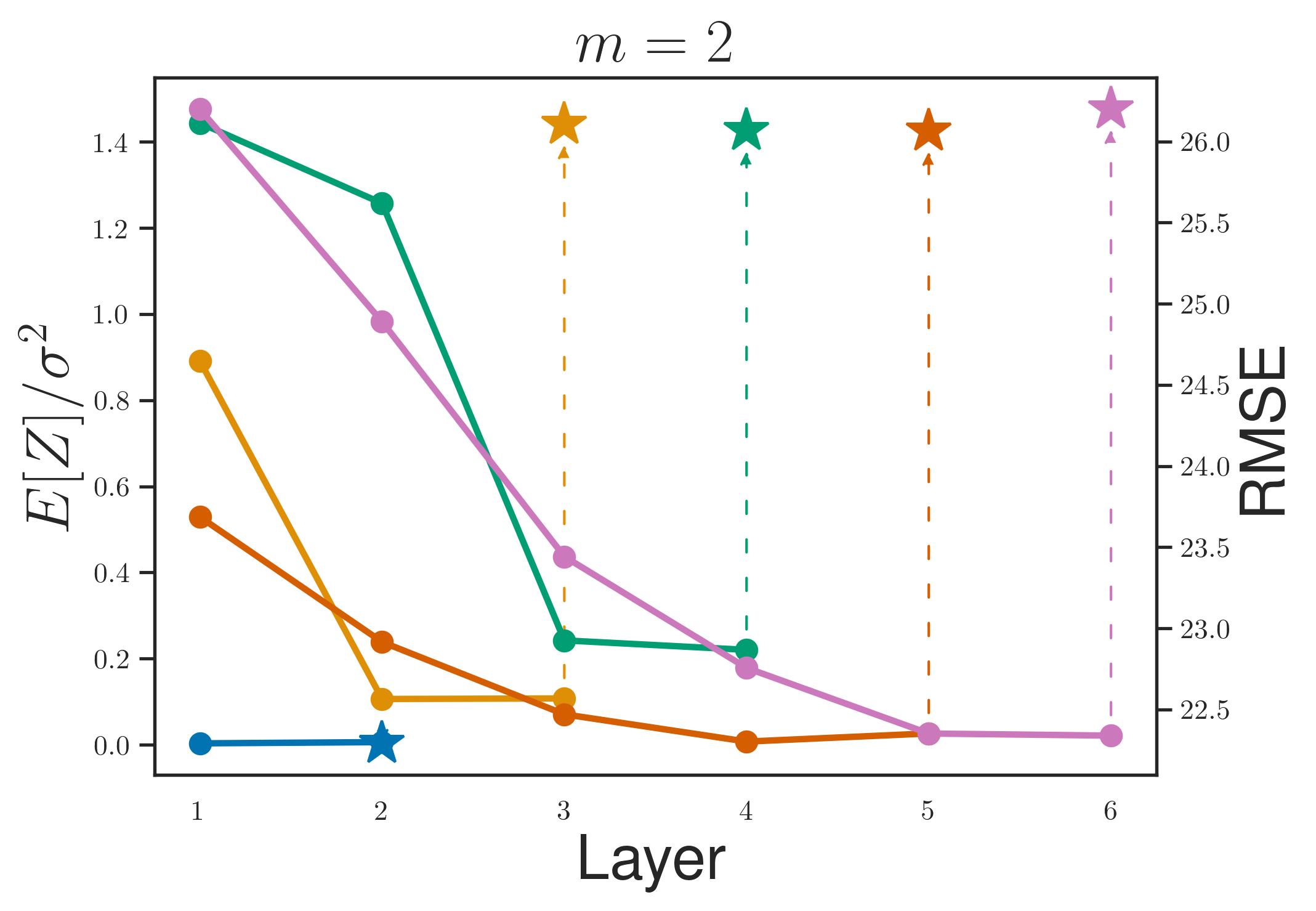}};
        
        \node[inner sep=0pt] at (0,0) {\includegraphics[width=0.22\textwidth]{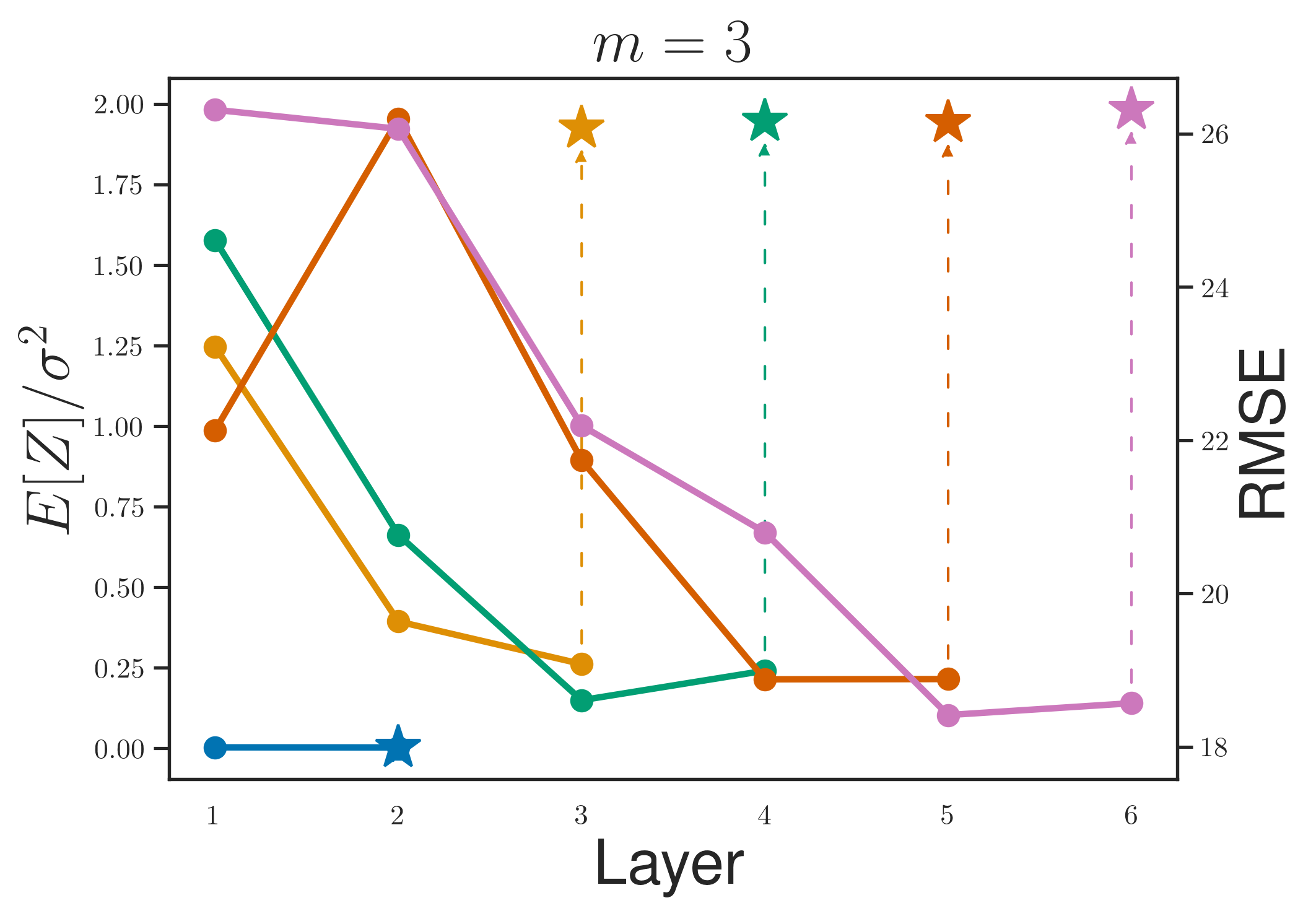}};
        
        \node[inner sep=0pt] at (4,0) {\includegraphics[width=0.22\textwidth]{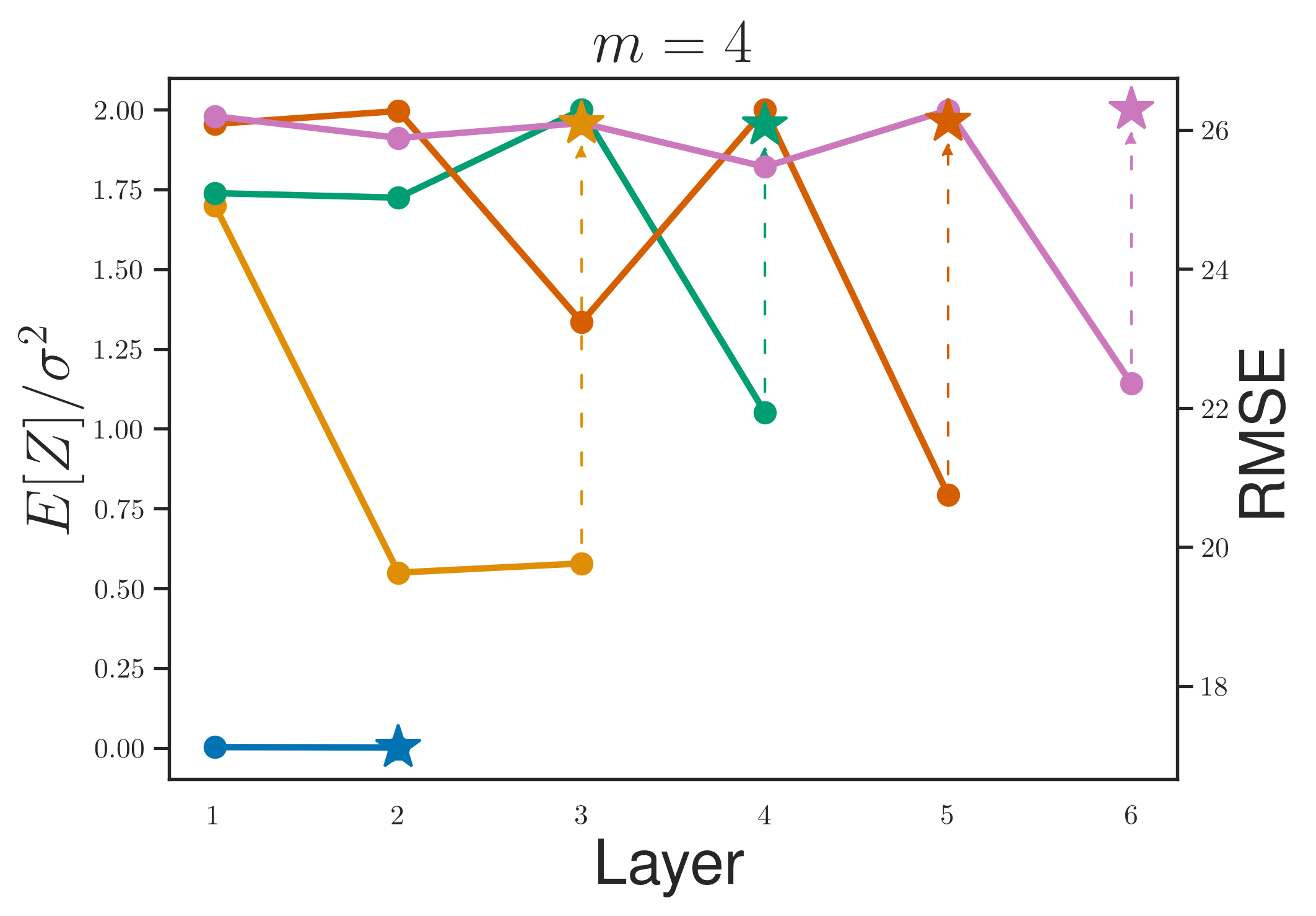}};
        
        \node[inner sep=0pt] at (8,0) {\includegraphics[width=0.22\textwidth]{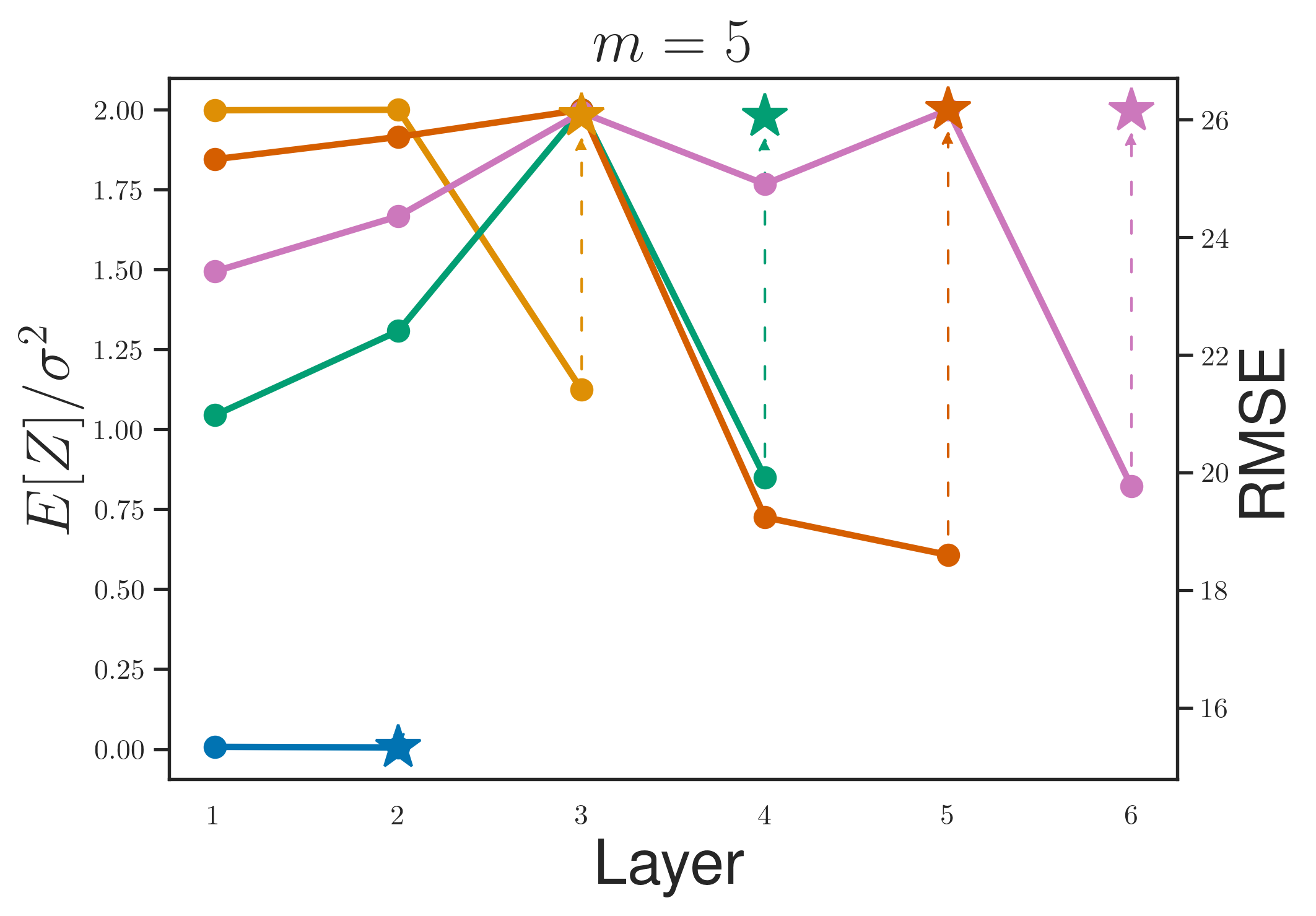}};
        
        \node[inner sep=0pt] at (-4,-2.8) {\includegraphics[width=0.22\textwidth]{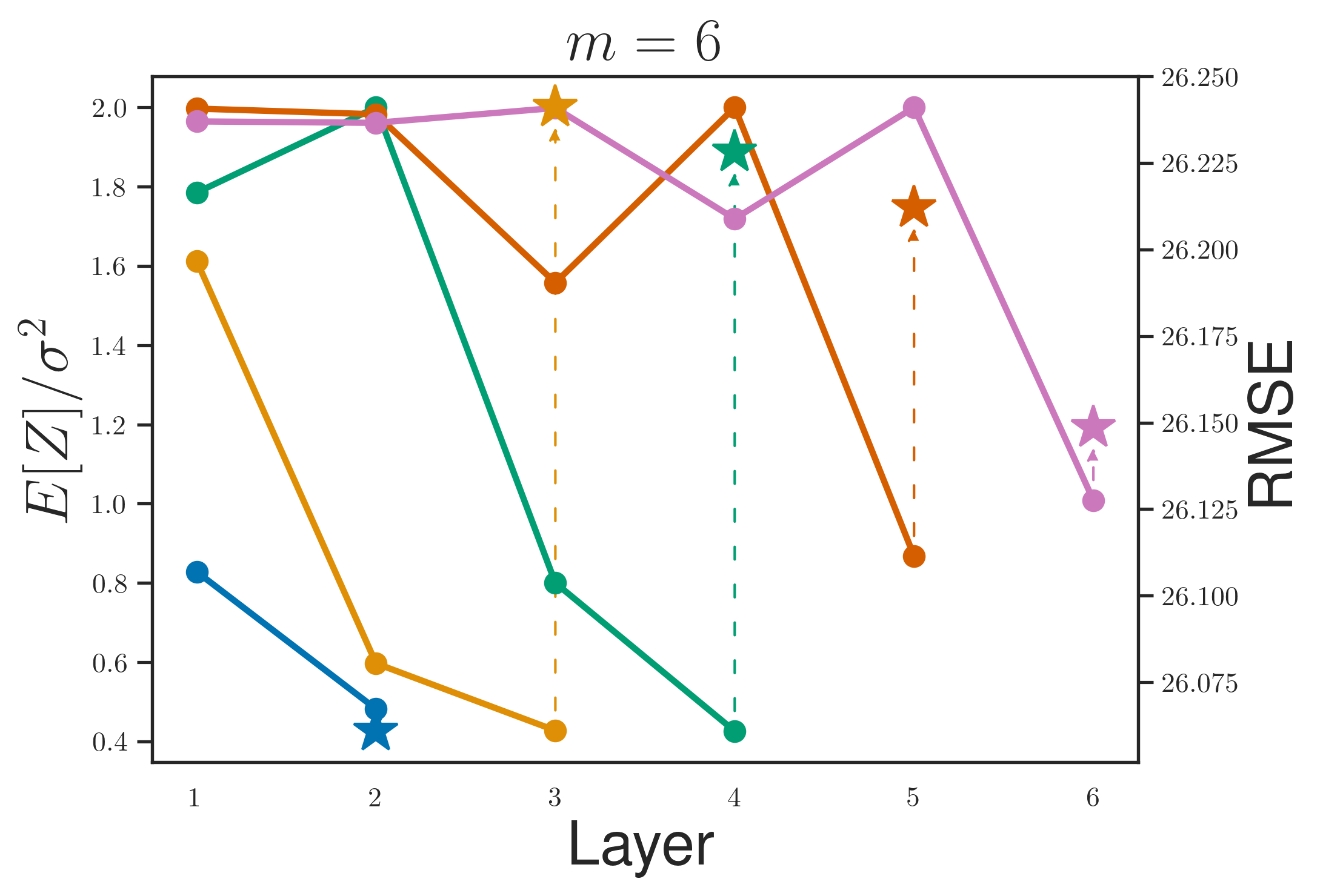}};
        
        \node[inner sep=0pt] at (0,-2.8) {\includegraphics[width=0.22\textwidth]{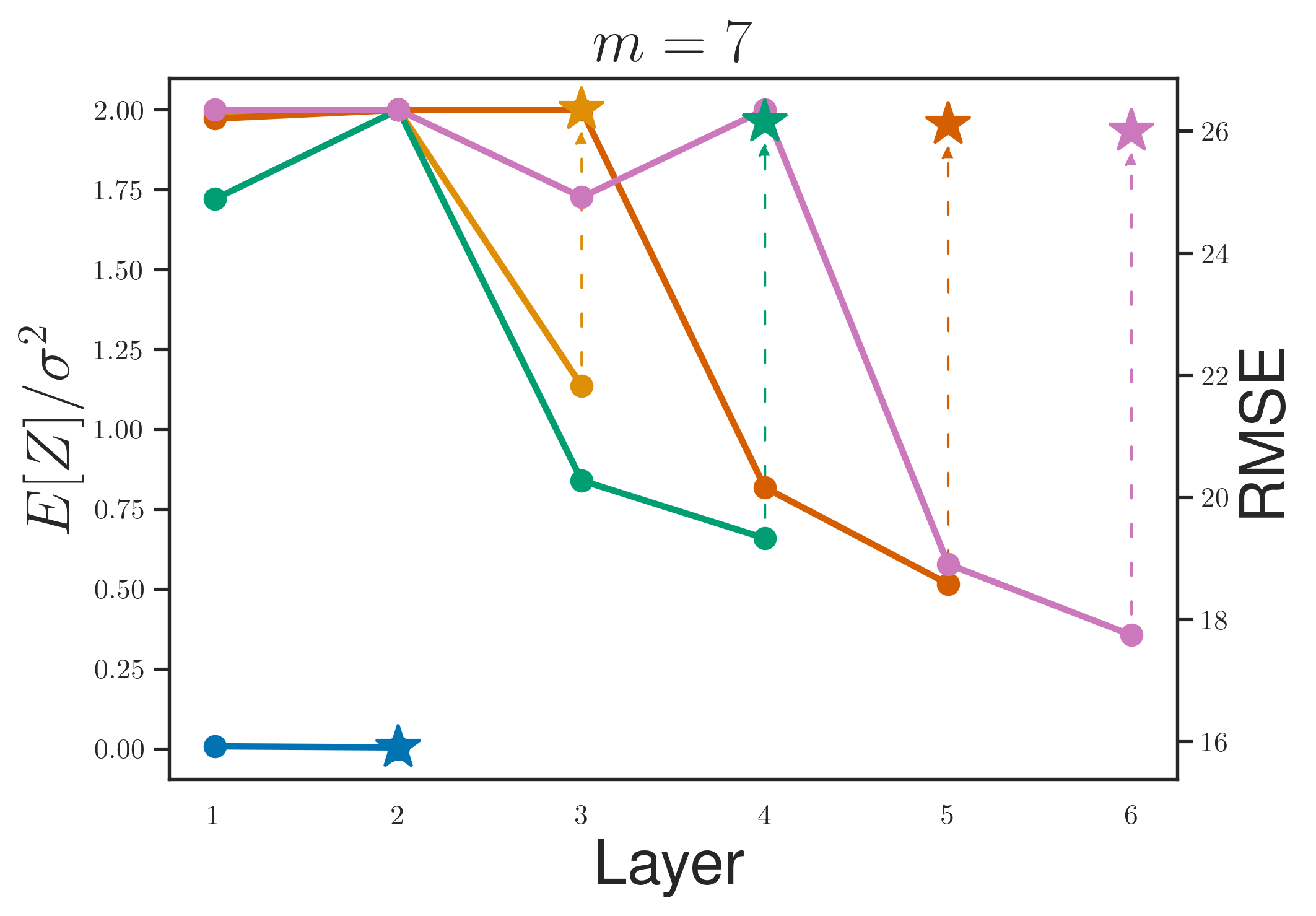}};
        
        \node[inner sep=0pt] at (4,-2.8) {\includegraphics[width=0.22\textwidth]{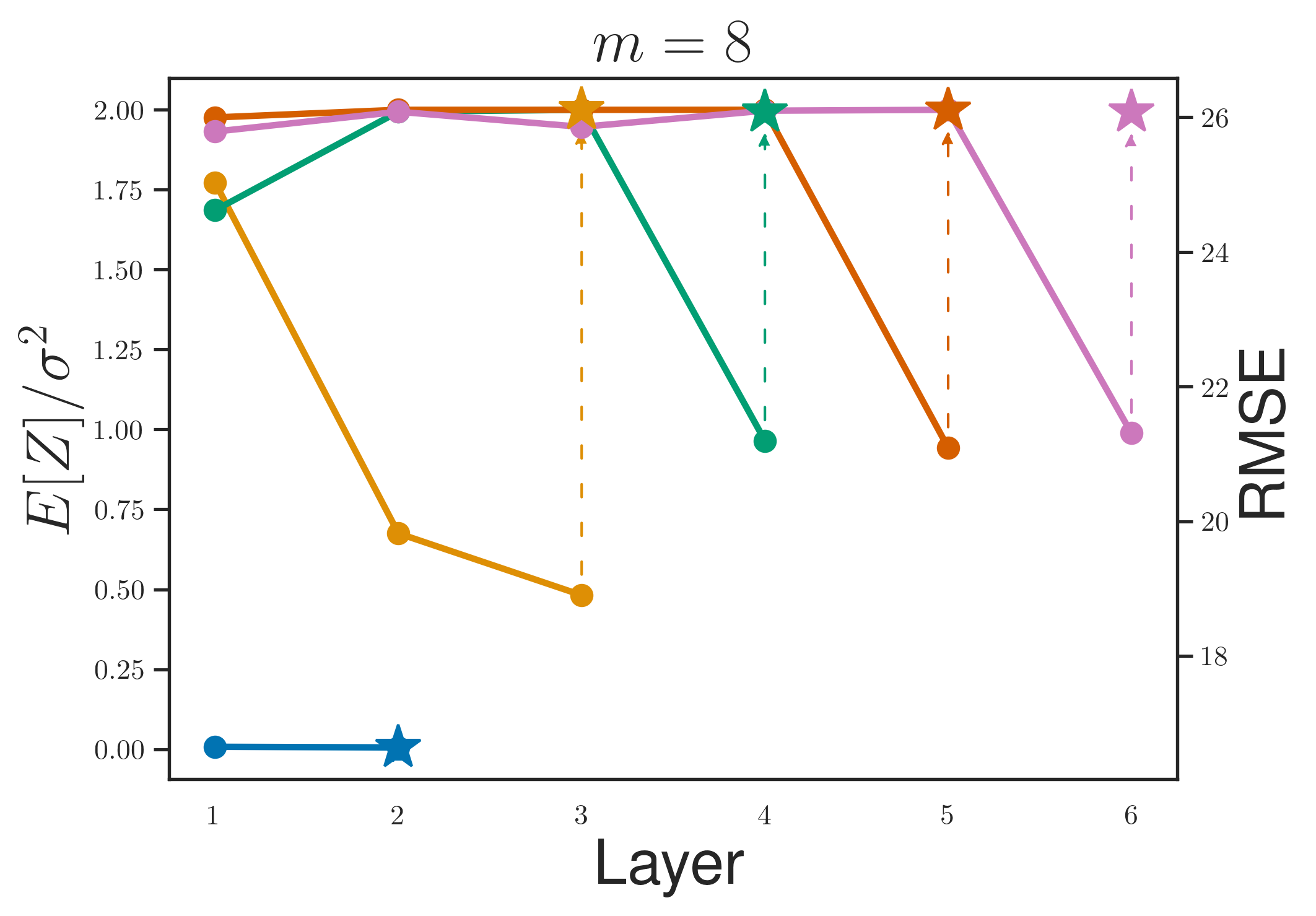}};
        
        \node[inner sep=0pt] at (8,-2.8) {\includegraphics[width=0.22\textwidth]{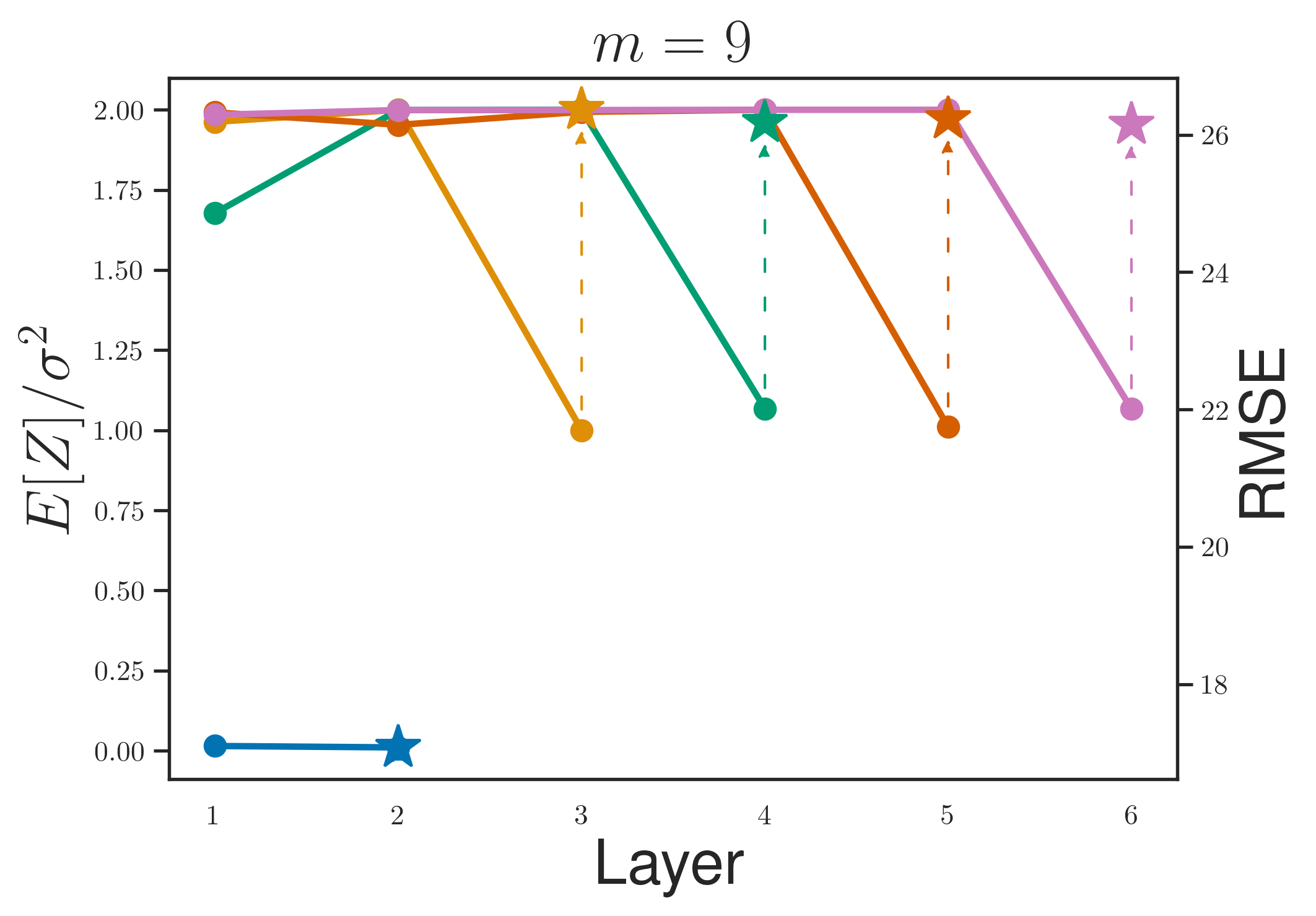}};

    \end{tikzpicture}
    \caption{\textbf{Standard zero-mean DGPs}: Results of diabetes data set}
    \label{fig:plot_diabetes_no_sharing}
\end{figure}

Figure~\ref{fig:plot_diabetes_no_sharing} contains the results from the diabetes data set with \textbf{standard} zero-mean~\dgps{}.

\begin{figure}
    \centering
    \begin{tikzpicture}
        
        \node[inner sep=0pt] at (-4.,0) {\includegraphics[width=0.22\textwidth]{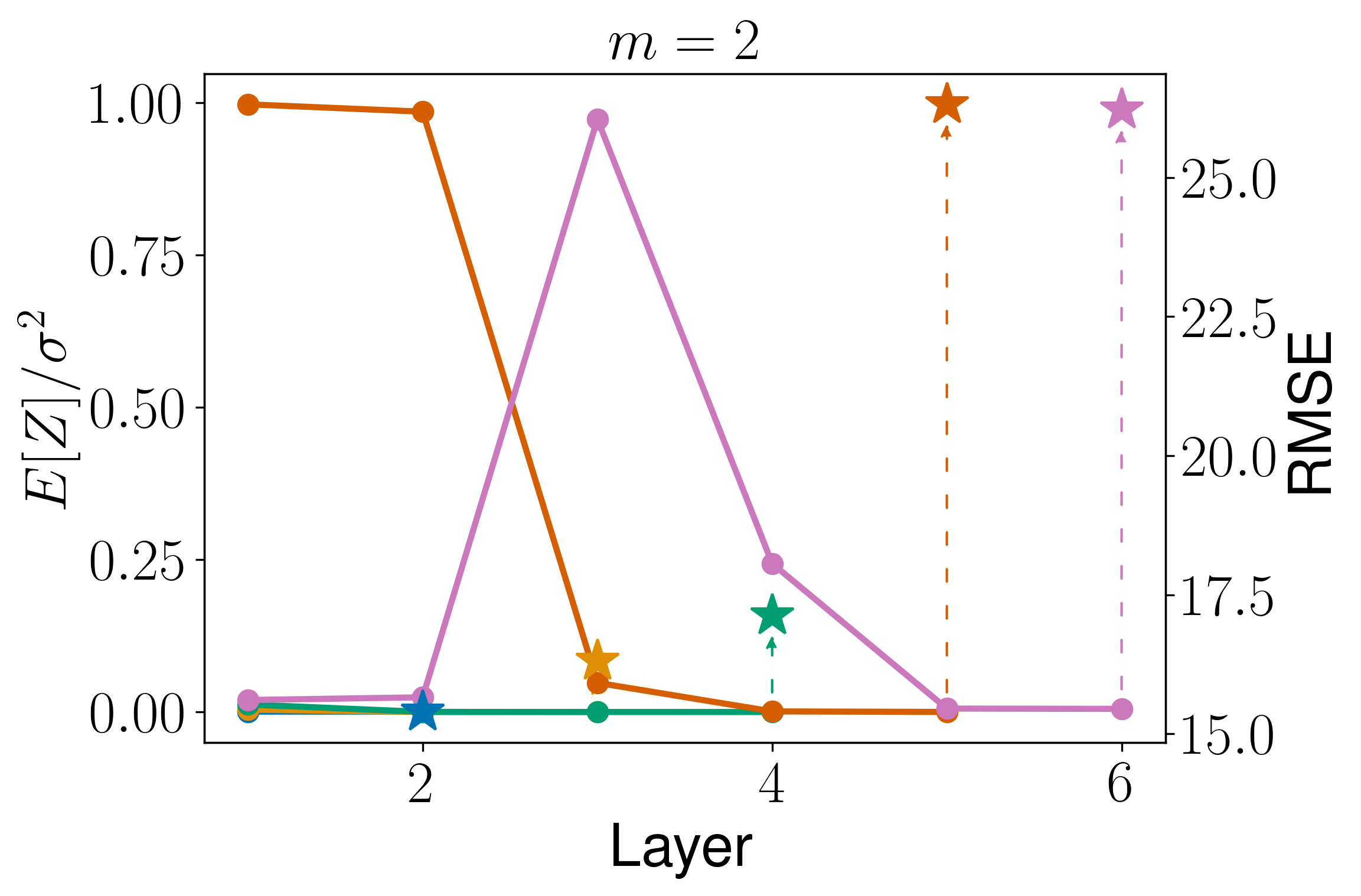}};
        
        \node[inner sep=0pt] at (0,0) {\includegraphics[width=0.22\textwidth]{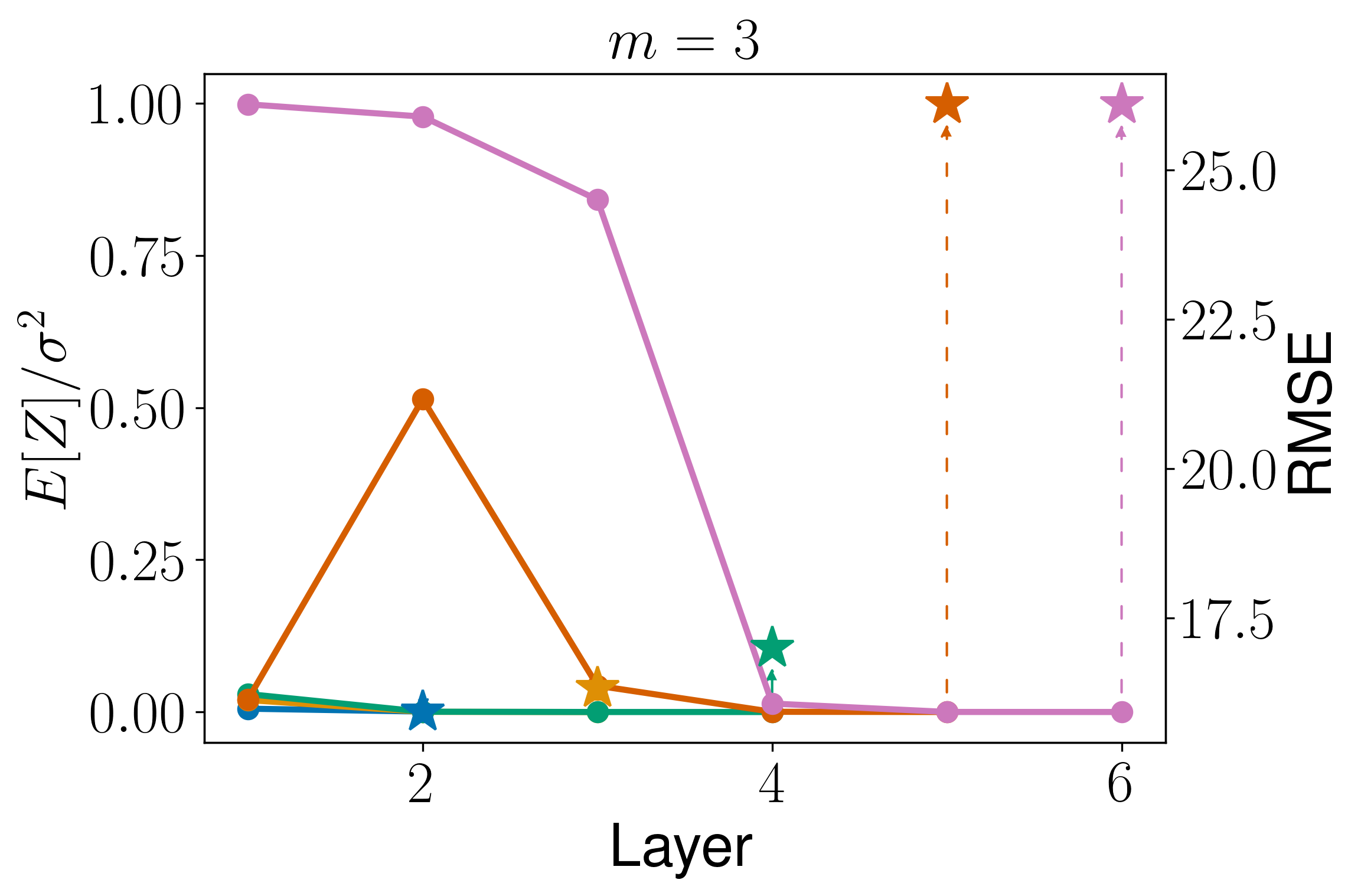}};
        
        \node[inner sep=0pt] at (4,0) {\includegraphics[width=0.22\textwidth]{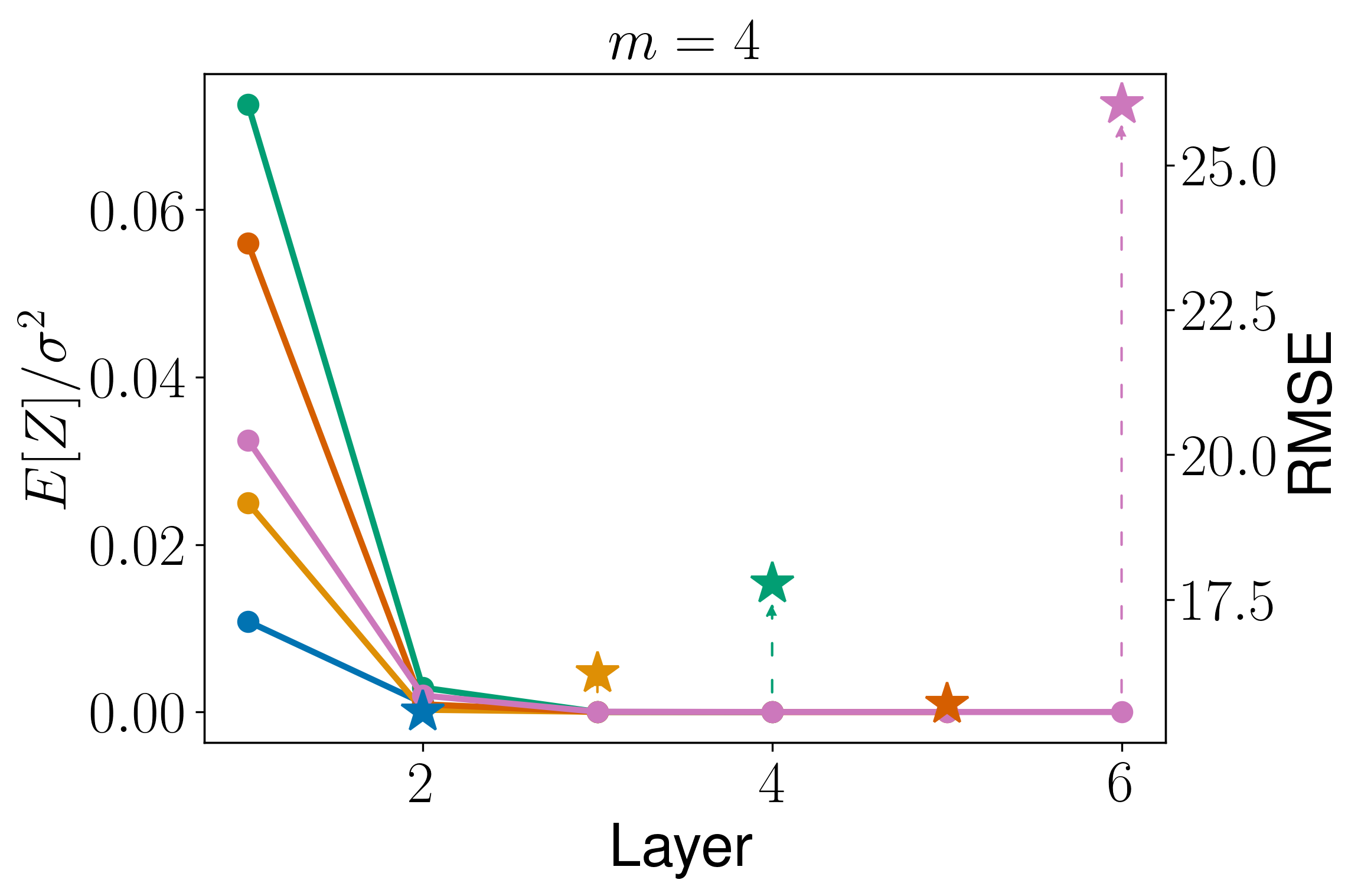}};
        
        \node[inner sep=0pt] at (8,0) {\includegraphics[width=0.22\textwidth]{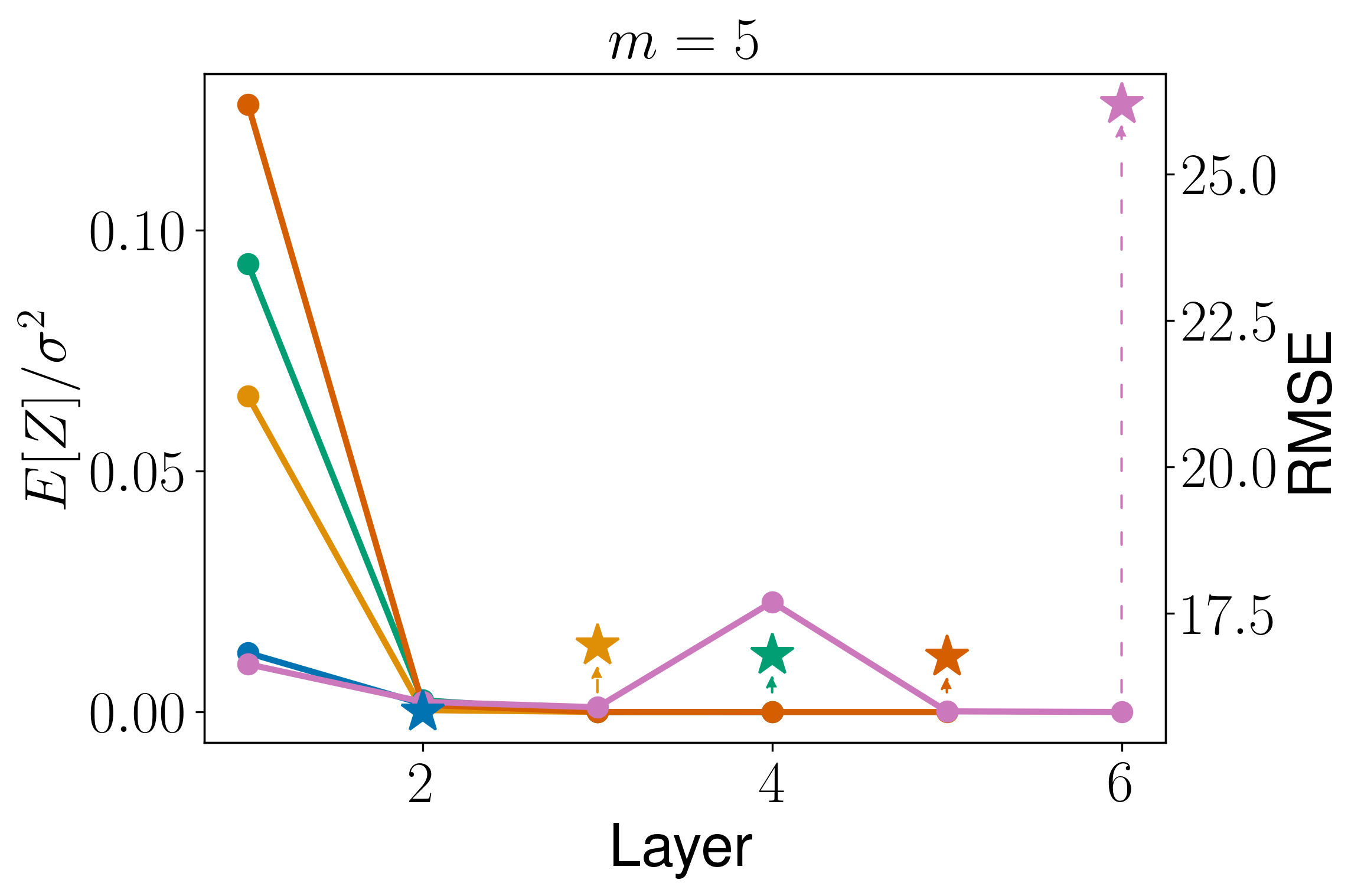}};
        
        \node[inner sep=0pt] at (-4,-2.8) {\includegraphics[width=0.22\textwidth]{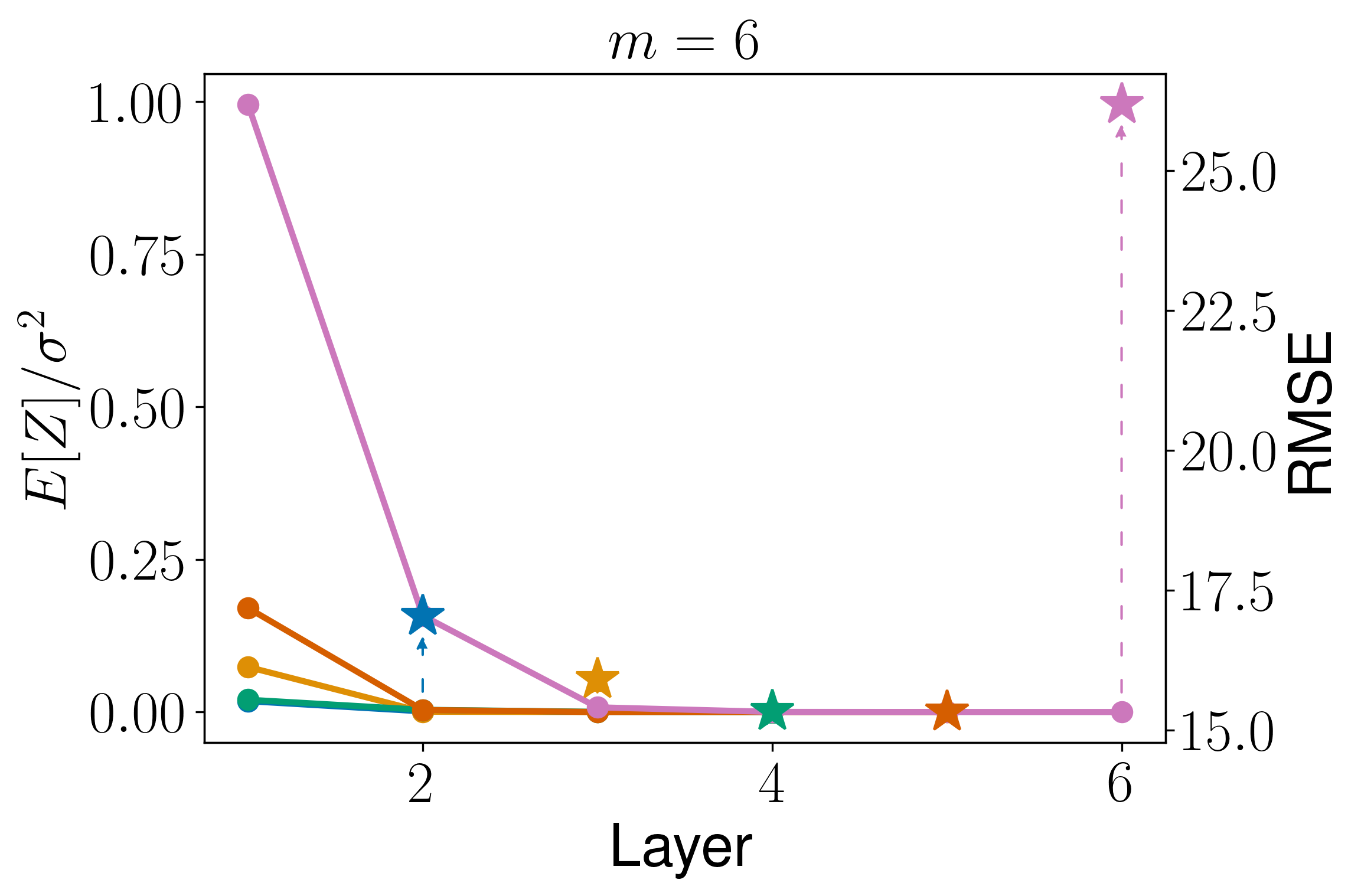}};
        
        \node[inner sep=0pt] at (0,-2.8) {\includegraphics[width=0.22\textwidth]{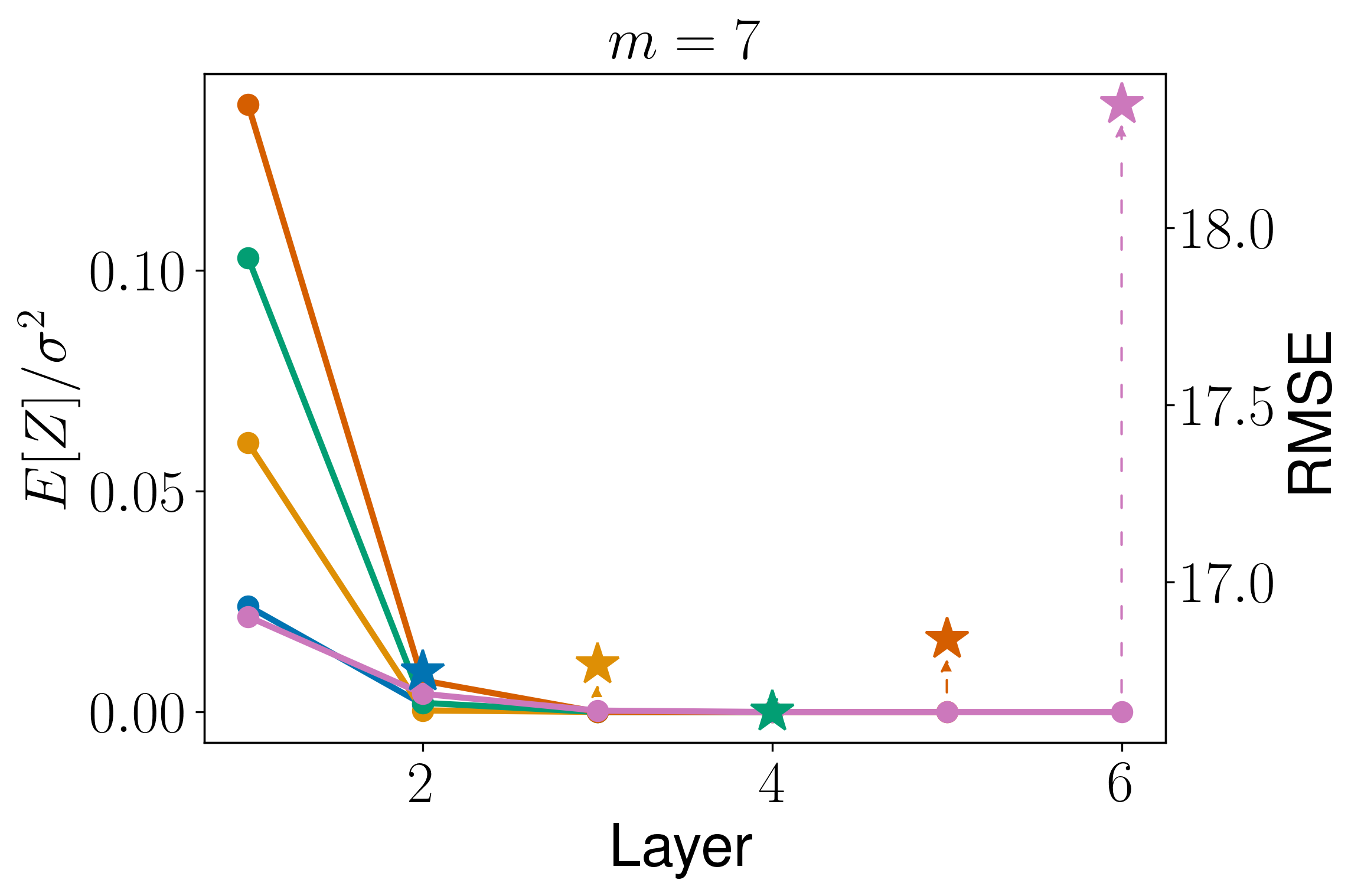}};
        
        \node[inner sep=0pt] at (4,-2.8) {\includegraphics[width=0.22\textwidth]{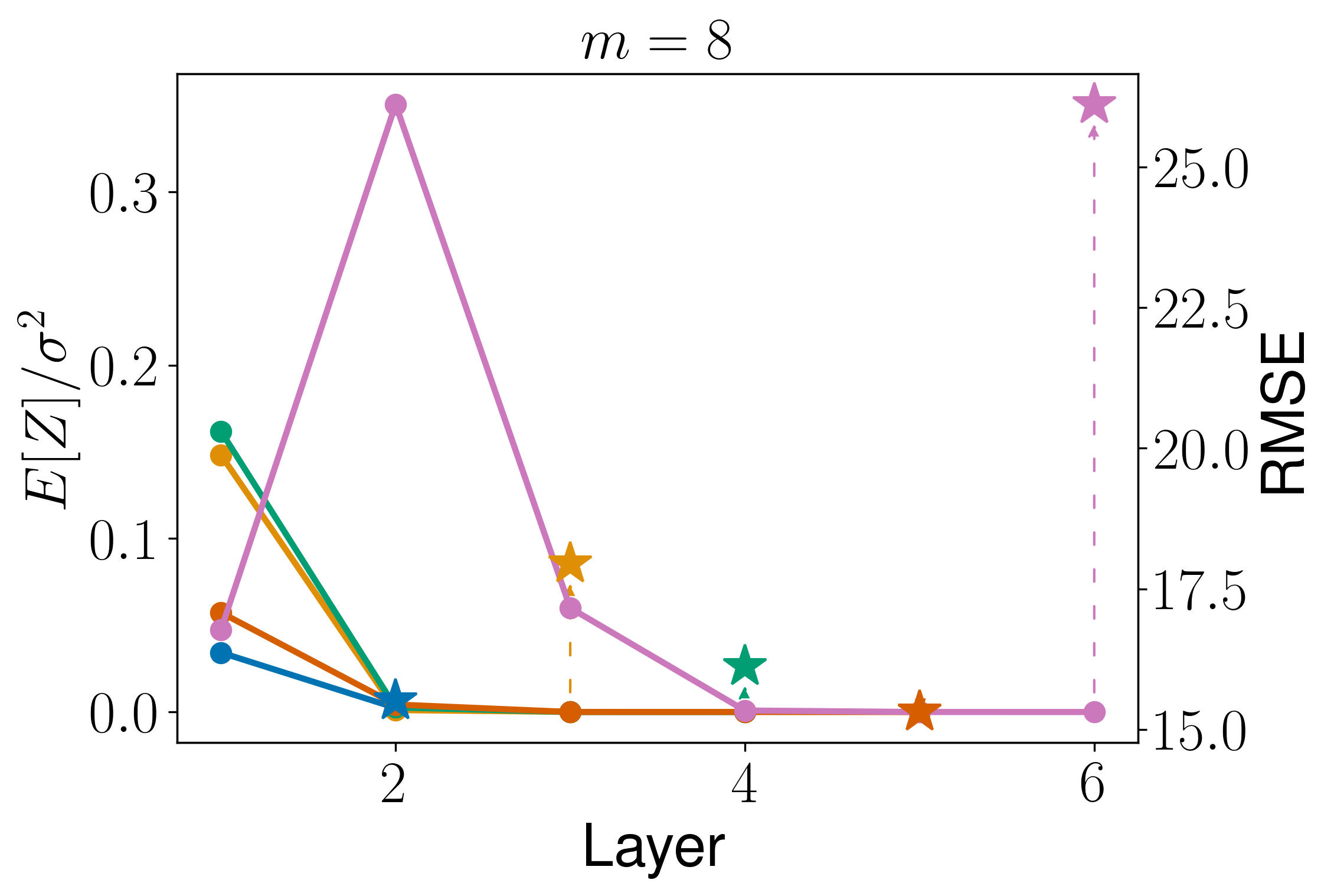}};
        
        \node[inner sep=0pt] at (8,-2.8) {\includegraphics[width=0.22\textwidth]{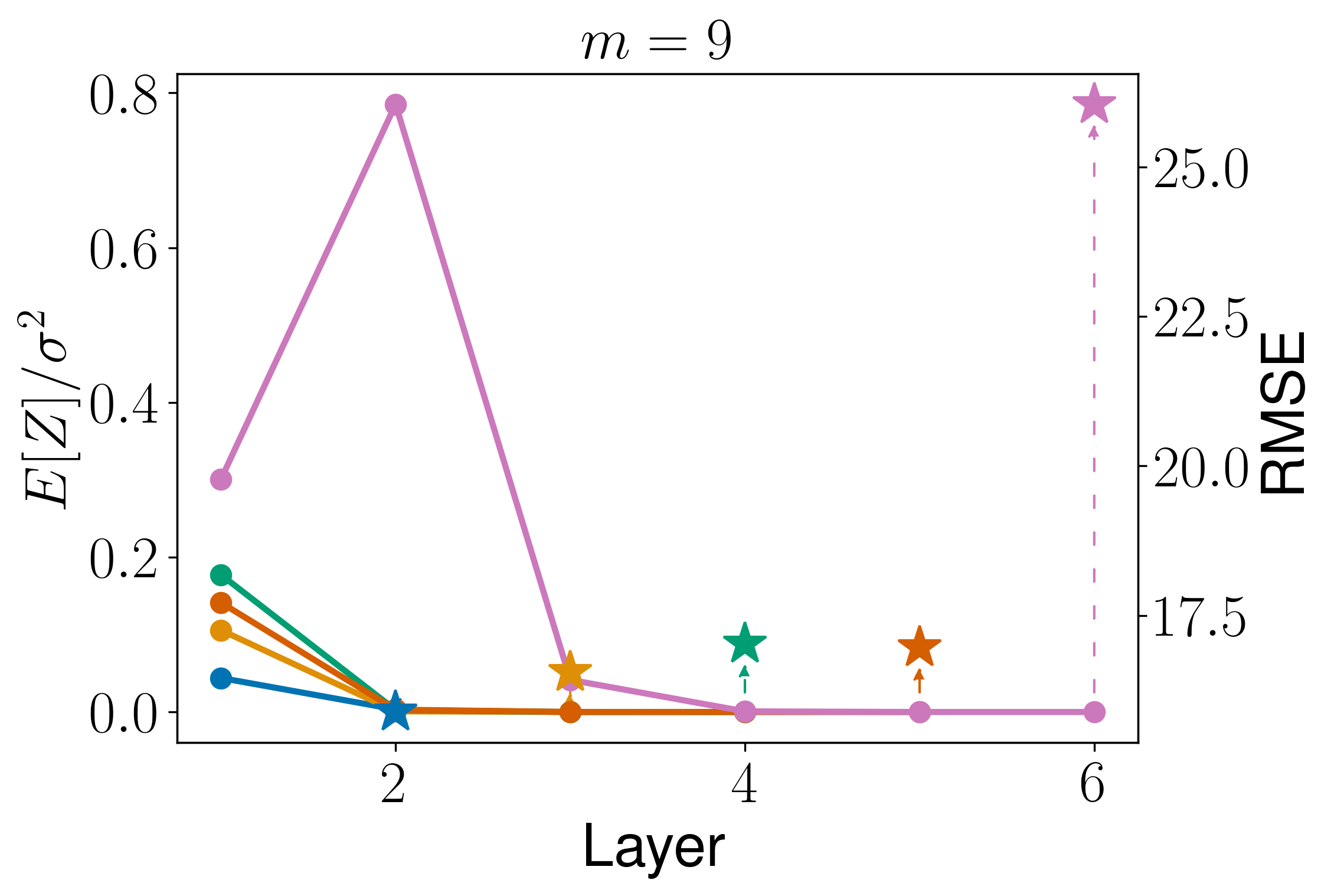}};

    \end{tikzpicture}
    \caption{\textbf{Constrained~\dgps{}}: Results of diabetes data set}
    \label{fig:plot_diabetes}
\end{figure}

Figure~\ref{fig:plot_diabetes} contains the results from the diabetes data set with \textbf{constrained} zero-mean~\dgps{} where $c_0 = 0.1$.

\end{document}